\documentclass[sigconf]{acmart}
\renewcommand\footnotetextcopyrightpermission[1]{} 
\AtBeginDocument{%
  \providecommand\BibTeX{{%
    \normalfont B\kern-0.5em{\scshape i\kern-0.25em b}\kern-0.8em\TeX}}}

\usepackage{tabularx}
\usepackage{array,multirow}
\graphicspath{ {./images/} }

\newcolumntype{Y}{>{\raggedright\arraybackslash}X}
\newcolumntype{M}{>{\centering\arraybackslash}m{1.8cm}}
\newcolumntype{N}{>{\centering\arraybackslash}m{2.5cm}}
\newcolumntype{K}{>{\arraybackslash}m{1.5cm}}
\usepackage{graphicx}
\usepackage{subfigure}
\usepackage{flexisym}
\usepackage{textcomp}
\usepackage{xcolor,colortbl}
\usepackage{geometry}
\usepackage{makecell}
\usepackage{color,soul}
\usepackage{textcomp}
\usepackage{color, colortbl}
\definecolor{Gray}{gray}{0.9}
\usepackage{wrapfig}
\usepackage{lscape}
\usepackage{rotating} 
\usepackage{tablefootnote}
\usepackage{amsmath}
\usepackage{caption}
\usepackage{tabularx}
\usepackage[utf8]{inputenc}
\usepackage{amsfonts}
\usepackage{array}
\usepackage[linesnumbered]{algorithm2e}

\newcommand{\argmin}[1]{\underset{#1}{\operatorname{argmin}}\;}
\newcommand\independent{\protect\mathpalette{\protect\independenT}{\perp}}
\def\independenT#1#2{\mathrel{\rlap{$#1#2$}\mkern2mu{#1#2}}}

\newcommand{\pr}{\mathbb{P}}
\newcommand{\ep}{\mathbb{E}}

\definecolor{Gray}{gray}{0.9}

\newcolumntype{a}{>{\columncolor{Gray}}p{0.3cm}}
\newcolumntype{e}{>{\columncolor{Gray}}p{0.75cm}}

\begin{document}

\title{Survey on Causal-based Machine Learning Fairness Notions}

\author{Karima Makhlouf}
\email{karima.makhlouf@lix.polytechnique.fr}
\orcid{0000-0001-6318-0713}
\affiliation{%
  \institution{INRIA, École Polytechnique, IPP}
  \city{Paris}
  \country{France}
}
\author{Sami Zhioua}
\email{sami.zhioua@lix.polytechnique.fr}
\orcid{0000-0001-7491-6271}
\affiliation{%
  \institution{INRIA, École Polytechnique, IPP}
  \city{Paris}
  \country{France}
}
\author{Catuscia Palamidessi}
\email{catuscia@lix.polytechnique.fr}
\orcid{0000-0003-4597-7002}
\affiliation{%
  \institution{Inria, École Polytechnique, IPP}
  \city{Paris}
  \country{France}
}
\renewcommand{\shortauthors}{Makhlouf, et al.}

\begin{abstract}
 Addressing the problem of fairness is crucial to safely use machine learning algorithms to support decisions with a critical impact on people's lives such as job hiring, child maltreatment, disease diagnosis, loan granting, etc. Several notions of fairness have been defined and examined in the past decade, such as statistical parity and equalized odds. The most recent fairness notions, however, are causal-based and reflect the now widely accepted idea that using causality is necessary to appropriately address the problem of fairness. This paper examines an exhaustive list of causal-based fairness notions and study their applicability in real-world scenarios. As the majority of causal-based fairness notions are defined in terms of non-observable quantities (e.g., interventions and counterfactuals), their deployment in practice requires to compute or estimate those quantities using observational data. This paper offers a comprehensive report of the different approaches to infer causal quantities from observational data including identifiability (Pearl's SCM framework) and estimation (potential outcome framework). The main contributions of this survey paper are (1) a guideline to help selecting a suitable fairness notion given a specific real-world scenario, and (2) a ranking of the fairness notions according to Pearl's causation ladder indicating how difficult it is to deploy each notion in practice.
\end{abstract}

\maketitle
\pagestyle{plain}

\section{Introduction}
\label{intro}
Machine learning algorithms are increasingly used to inform automated decisions with critical impact on people's lives including job hiring, loan granting, predicting recidivism during parole, etc. Correcting bias in the decision prediction requires first to measure it.  The most commonly used fairness notions are observational and rely on mere correlation between variables. For example, statistical parity~\cite{darlington1971} requires that the proportion of positive outcome (e.g. granting loans) is the same for all sub-populations (e.g. male and female groups).  Equal opportunity~\cite{hardt2016equality} requires that the true positive rate (TPR) is the same for all sub-populations. The main problem of correlation-based fairness notions is that they fail to detect discrimination in presence of statistical anomalies such as Simpson's paradox~\cite{simpson1951interpretation}. A famous example of the Simpson's paradox is the gender bias in 1973 Berkeley admission~\cite{berkeley75,loftus18}. In that year, 44\% of male applicants were admitted against only 34\% of female applicants. While this looks like a bias against female candidates, when the same data has been analyzed by department, acceptance rates were approximately the same. In other words, the statistical conclusions drawn from the sub-populations differ from that from the whole population. Considering the problem of fairness from the legal and philosophical point of view reveals another limitation of statistical fairness notions. In the disparate treatment liability framework~\cite{barocas2016big}, discrimination claims require plaintiffs to demonstrate a causal connection between the challenged decision (e.g., hiring, firing, admission) and the sensitive feature (e.g., gender, race). It is then necessary to investigate the causal relationship between the sensitive attribute and the decision rather than the associated relationship. Because of these two limitations, it is now widely accepted that causality is necessary to appropriately address the problem of fairness~\cite{loftus18}. 

Various causal-based fairness notions have been recently proposed to tackle the problem of fairness through causal inference lenses. These include total effect~\cite{pearl2009causality}, counterfactual fairness~\cite{kusner2017counterfactual}, counterfactual effects~\cite{zhang2018fairness}, interventional fairness~\cite{salimi2019interventional}, etc. These notions differ from statistical fairness approaches in that they are not totally based on data but consider additional knowledge about the structure of the world, in the form of a causal model. This additional knowledge helps to understand how data is generated in the first place and how changes in variables propagate in a system. Most of these fairness notions are defined in terms of non-observable quantities such as interventions (to simulate random experiments) and counterfactuals (which consider other hypothetical worlds, in addition to the actual world). Such quantities cannot be always uniquely computed from observational data which hinders significantly the applicability of causal-based notions in practical scenarios. Each one of the two main causal frameworks in the literature, namely, structural causal model (SCM) with causal graphs~\cite{pearl2009causality} and potential outcome~\cite{rubin2015book}, use a different approach to compute/estimate the causal quantities using observational data. The SCM framework relies mainly on the identifiability criterion~\cite{shpitser2006identification} to generate an expression for the causal quantity based only on observable probabilities. If the identifiability criterion is not satisfied, the causal quantity can not be computed using the available observable data. In such case, as an alternative, if the complete structure of the causal model is available, it is possible to estimate the distribution of the latent variables $U$ and consequently generate an estimation of the counterfactual outcomes~\cite{kusner2017counterfactual}. In the potential outcome framework, causal quantities are approximated using several estimation techniques (e.g., matching, re-weighting, etc.)~\cite{guo2020survey}. 

Nineteen causal-based fairness notions are examined in this paper. Given a real-world scenario, selecting which fairness notion to use is a challenging and error-prone task as using the wrong fairness notion may indicate unfairness in an otherwise fair scenario, or the opposite (failing to detect unfairness in an unfair scenario). This survey paper provides guidelines to help selecting a suitable fairness notion given a specific real-world scenario. The guidelines are summarized in a decision diagram that can be easily navigated using the characteristics of the real-world scenario at hand.  On the other hand, according to Pearl's SCM framework, computing causal quantities (interventions and counterfactuals) depends on their identifiability. Hence, even if a fairness notion is appropriate in some setup, it might not be applicable because of identifiability issues. Placing the various causal-based fairness notions in Pearl's causation ladder with the three corresponding rungs (observation, intervention, and counterfactual)~\cite{pearl2018book} allows to rank these notions and indicates how difficult to deploy each one of them in practice. 

This survey paper is a comprehensive report on assessing machine learning fairness with causality lenses. It starts by illustrating the need for causality through a hypothetical example of teacher firing (Section~\ref{sec:example}). Then, it provides essential background on causal inference in sufficient detail for our analysis (Section~\ref{sec:notation}). Section~\ref{sec:notions} examines a comprehensive list of causal-based fairness notions. Unlike other surveys in the literature, the subtleties of the fairness notions are illustrated using a very simple numerical job hiring example. A survey on the three approaches to compute causal quantities from observable data, namely, identifiability, estimation based on full causal model, and potential outcome estimation, is provided in Section~\ref{sec:computation}. The main contributions of the survey which are the suitability and applicability of causal-based fairness notions are described in Section~\ref{sec:applicability}. Finally, Section~\ref{sec:conclusion} concludes.

\section{The need for causality: an example}
\label{sec:example}
Consider the hypothetical example\footnote{Inspired by the prior convictions example in~\cite{shpister18}.} of an automated system for deciding whether to fire a teacher at the end of the academic year. Deployed teacher evaluation systems have been suspected of bias in the past. For example, IMPACT is a teacher evaluation system used in the city of Washington DC and have been found to be unfair against teachers from minority groups~\cite{impact,impactBias,weapons16}. Assume that the system takes as input two features, namely, the location of the school where the teacher is working ($C$) and the initial\footnote{At the beginning of the academic year.} average level of the students in her class ($A$). The outcome is whether to fire the teacher ($Y$). Assume also that all 3 variables are binary with the following values: if the school is located in a high-income neighborhood, $C=1$, otherwise (the school is located in a low-income neighborhood), $C=0$. If the initial average score for the students assigned to the teacher is high, $A=1$, otherwise (initial level is low), $A=0$. Firing a teacher corresponds to $Y=1$, while retaining her corresponds to $Y=0$. The level of students in a given class can be influenced by several variables, but in this example, assume that it is only influenced by the location of the school; students in high-income neighborhoods are more advantaged and typically perform better in school. 

Assume now that the automated decision system is suspected to be biased by the initial level of students assigned to the teacher. That is, it is claimed that the system will more likely fire teachers who have been assigned classes with low level students at the beginning of the academic year which is clearly unfair. The sensitive attribute in this case is the initial level of students assigned to the teacher ($A$). For concreteness, consider the prediction system that yields the following conditional probabilities:
\[\begin{array}{lcl}
		\pr(Y=1\;|\;A=1, C=0) = 0.02  & \quad & \pr(A=1\;|\;C=0) = 0.2 \\
		\pr(Y=1\;|\;A=1, C=1) = 0.0675 & \quad & \pr(A=1\;|\;C=1) = 0.8 \\
		\pr(Y=1\;|\;A=0, C=0) = 0.01 & \quad & \pr(A=0\;|\;C=0) = 0.8  \\
		\pr(Y=1\;|\;A=0, C=1) = 0.25 & \quad & \pr(A=0\;|\;C=1) = 0.2  \\
\end{array}\]
and that the dataset is collected from a population where schools are located with equal proportions in high-income and low-income neighborhood, that is, $\pr(C=1) = \pr(C=0) = 0.5$. Assume also that the proportion of classes with a low initial average level of students is the same as the one with high average initial level of students, that is, $\pr(A=1) = \pr(A=0) = 0.5$. To keep the scenario simple, assume that the level of students $A$ does not depend on any other feature except $C$ and that the firing decision $Y$ depends only on $A$ and $C$.

A simple approach to check the fairness of the firing decision $Y$ with respect to the sensitive attribute $A$ is to contrast the conditional probabilities: $\pr(Y=1\;|\;A=0)$ and $\pr(Y=1\;|\;A=1)$ which quantify, respectively, the likelihood of firing a teacher given that she is assigned students with an initial low level versus and the likelihood of firing a teacher given that she is assigned students with an initial high level class. Such probabilities can be computed as follows: 
\begin{align}
	\pr(Y=1\;|\;A=a) & = \sum_{c\in\{0,1\}} \pr(Y=1\;|\;A=a,C=c,) \nonumber \\
	& \qquad \qquad \times\pr(A=a\;|\;C=c) \label{eq:naiveExample1} 
\end{align}
Hence, 
\begin{align}
\pr(Y=1\;|\;A=1) &= 0.02 \times 0.2 + 0.0675 \times 0.8 = 0.058 \nonumber \\
\pr(Y=1\;|\;A=0) &= 0.01 \times 0.8 + 0.25 \times 0.2 = 0.058 \nonumber
\end{align}

As the values are equal, the rates of firing between teachers who were assigned low level students and high level students appear to be equal and hence no discrimination is detected\footnote{This corresponds to statistical parity.}. This conclusion is flawed because it doesn't consider the mechanism by which the observed data is generated. In particular, the location of the school in which the teacher is working influences both the initial level of students assigned to her as well as the decision to fire or retain her. The $\pr(A|C)$ distribution indicates that $80\%$ of classes in low income neighborhoods have students with low initial levels ($\pr(A=0\;|\;C=0) = 0.8$) while $80\%$ of classes in high income neighborhoods have students with high initial levels ($\pr(A=1\;|\;C=1) = 0.8$). The automated decision system is biased in this case because $\pr(Y=1\;|\;A=0,C=1)$, the probability of firing a teacher in high income neighborhoods which is assigned a class with low initial level, is exceptionally high ($0.25$). Using simple conditional probabilities (Eq.~\ref{eq:naiveExample1}) on this collected dataset fails to appropriately account for that bias because very few teachers in high income neighborhoods are assigned low level classes in this particular dataset ($\pr(A=0\;|\;C=1) = 0.2$). 
In general, any statistical fairness notion which relies solely on correlation between variables, will fail to detect such bias. 

\begin{figure}[!h]
	\centering
	{\includegraphics[scale=0.3]{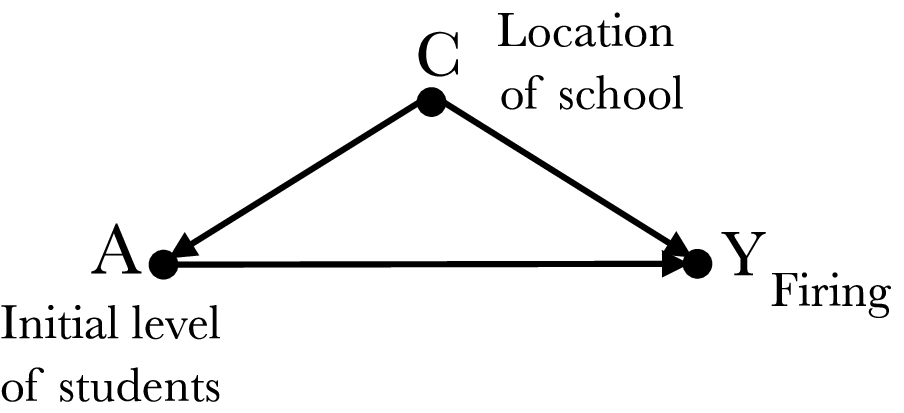}}
	\caption{Causal graph of the firing example.}
	\label{fig:firing_example}
\end{figure}

To avoid such misleading conclusions, the causal relationships between variables should be considered. Figure~\ref{fig:firing_example} illustrates the causal relations between the three variables of the above example where the location of the school $C$ is a confounder. Based on such causal graph, a firing decision system is fair if it is as likely to fire teachers in the following two hypothetical cases: (1) when \em{all teachers in the population} are assigned students of low level on average, and (2) when \em{all teachers in the population} are assigned students of high level on average. This is achieved using intervention ($do()$ operator)\footnote{Intervention and the $do()$ operator will be explained further in Section~\ref{identInterv}.} and allows to break the problematic dependence between $A$ and $C$. The probabilities of firing a teacher in these two hypothetical cases are expressed as $\pr(Y_{A=0}=1) = \pr(Y=1\;|\;do(A=0))$ and $\pr(Y_{A=1}=1) = \pr(Y=1\;|\;do(A=1))$ respectively. In this simple graph, and assuming no other variable is used in the prediction, these probabilities can be computed as follows:

\begin{align}
	\pr(Y_{A=a}=1) & = \sum_{c\in\{0,1\}} \pr(Y=1\;|\;A=a,C=c) \times \pr(C=c) \nonumber 
\end{align}
Hence, 
\begin{align}
\pr(Y_{A=1}=1) &= 0.02 \times 0.5 + 0.0675 \times 0.5 = 0.0437 \nonumber \\
\pr(Y_{A=0}=1) &= 0.01 \times 0.5 + 0.25 \times 0.5 = 0.13 \nonumber
\end{align}

The values confirm the existence of a bias against teachers which are assigned classes with initial low levels.

\section{Preliminaries and Notation}
\label{sec:notation}
Variables are denoted by capital letters. In particular, $A$ is used for the sensitive variable (e.g., gender, race, age) and $Y$ is used for the outcome of the automated decision system (e.g., health-care intervention, hiring, admission, releasing on parole). Small letters denote specific values of variables (e.g., $A=a'$, $W=w$). Bold capital and small letters denote a set of variables and a set of values, respectively.

There are two fundamental frameworks to mathematically represent and characterize causal relations between variables: structural causal model~\cite{pearl2009causality} and potential outcome~\cite{rubin2015book}. Formally, the two frameworks are equivalent~\cite{pearlBlog12,morgan2015book}. However, each of them is more equipped to address different problems in particular situations. For example, accounting for the many causal pathways that may exist in real-applications can be more straightforward using SCM. On the other hand, potential outcome framework is preferred when estimating individual-level causal effects. As most of causal-based fairness notions are defined using one of these frameworks, the rest of this section introduces the terminology and the notation for both frameworks.  

\subsection{Structural Causal Model (SCM) Framework}

A structural causal model~\cite{pearl2009causality} is a tuple $M = \langle \mathbf{U},\mathbf{V},\mathbf{F},\pr(\mathbf{U})\rangle$ where:
\begin{itemize}
	\item $\mathbf{U}$ is a set of exogenous variables which cannot be observed or experimented on but constitute the background knowledge behind the model.
	\item $\mathbf{V}$ is a set of observable variables which can be experimented on.
	\item $\mathbf{F}$ is a set of structural functions where each $f_i$ is mapping $U \cup V \rightarrow V\backslash\{V_i\}$ which represents the process by which variable $V_i$ changes in response to other variables in $U \cup V$.
	\item $\pr(\mathbf{u})$ is a probability distribution over the unobservable (latent) variables $\mathbf{U}$. 

\end{itemize}

\begin{figure}[!h]
\vspace{-2mm}
    \subfigure []  {%
    {\includegraphics [scale=0.2]{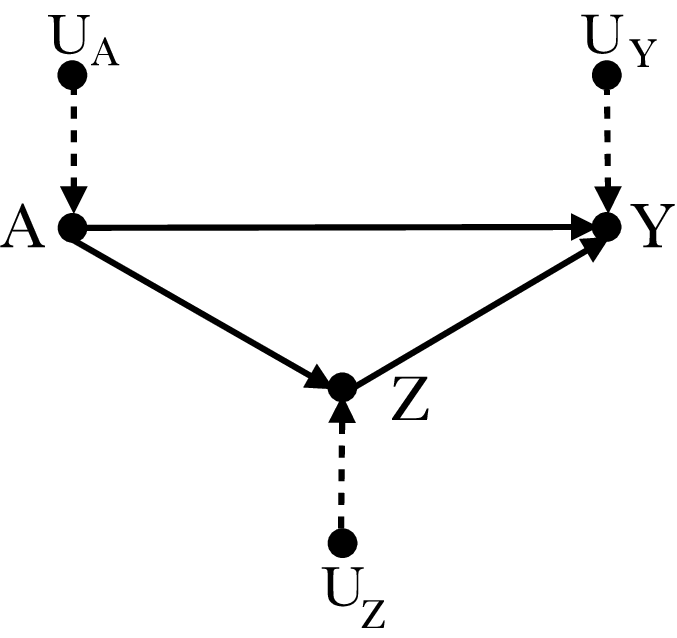} }
    \label{subfig:figMark_a}}
    \quad 
    \subfigure [] {%
    {\includegraphics[scale=0.2]{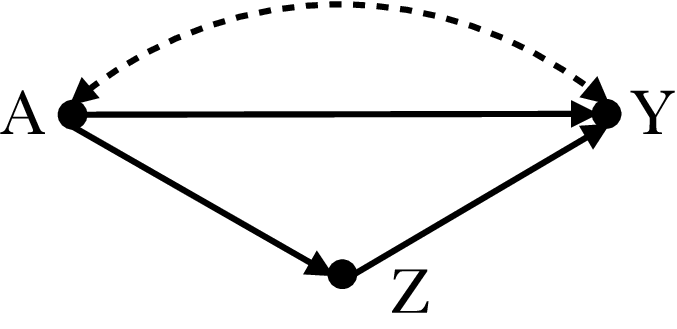} }
    \label{subfig:figMark_b}}
     \quad 
    \subfigure []  {%
    {\includegraphics [scale=0.2]{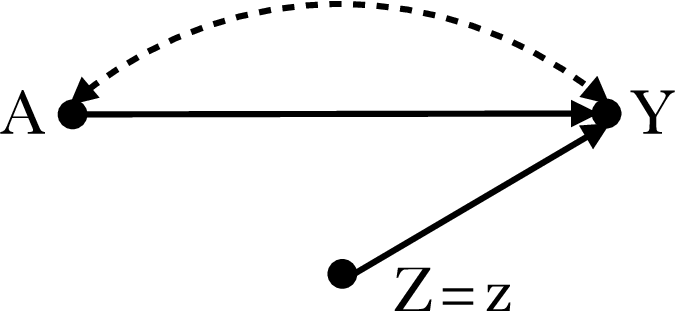} }
    \label{subfig:figMark_c}}
    \vspace{-2mm}
    \caption{Markovian and semi-Markovian causal models.}
    \label{fig:Mark_semiMark}
\end{figure}

Causal assumptions between variables are captured by a causal diagram $G$ which is a directed acyclic graph (DAG) where vertices represent variables and directed edges represent functional relationships between the variables. Directed edges can have two interpretations. A probabilistic interpretation where the edge represents a dependency among the variables such that the direction of the edge is irrelevant. A causal interpretation where the edge represents a causal influence between the corresponding variables such that the direction of the edge matters. In presence of a cause effect relation between two variables $A$ and $Y$, a confounder is a third variable $C$ which affect both the cause $A$ and the effect $Y$. For example, the location of school variable $C$ in Figure~\ref{fig:firing_example} is a confounder. Unobserved variables $\mathbf{U}$, which are typically not represented in the causal diagram, can be either mutually independent (Markovian model) or dependent from each others. In case the unobserved variables can be dependent and each $U_i \in \mathbf{U}$ is used in at most two functions in $F$, the model is called semi-Markovian. In causal diagrams of semi-Markovian models, dependent unobservable variables (unobserved confounders) are represented by a dotted bi-directed edge between observable variables. Figure~\ref{fig:Mark_semiMark} shows causal graphs of Markovian model (Figure~\ref{subfig:figMark_a}), semi-Markovian model (Figures~\ref{subfig:figMark_b})  and semi-Markovian model after intervening on $Z$ (Figure~\ref{subfig:figMark_c}). 

An intervention, noted $do(V=v)$, is a manipulation of the model that consists in fixing the value of a variable (or a set of variables) to a specific value regardless of the corresponding function $f_v$. Graphically, it consists in discarding all edges incident to the vertex corresponding to variable $V$. Figure~\ref{subfig:figMark_c} shows the causal diagram of the manipulated model after intervention $do(Z=z)$ denoted $M_{Z=z}$ or $M_z$ for short. The intervention $do(V=v)$ induces a different distribution on the other variables. For example, in Figure~\ref{subfig:figMark_c}, $do(Z=z)$ results in a different distribution on $Y$, namely, $\pr(Y|do(Z=z))$. Intuitively, while $\pr(Y|Z=z)$ reflects the population distribution of $Y$ among individuals whose $Z$ value is $z$, $\pr(Y|do(Z=z)$ reflects the population distribution of $Y$ if \textit{everyone in the population} had their $Z$ value fixed at $z$. The obtained distribution $\pr(Y|do(Z=z)$ can be considered as a \textit{counterfactual} distribution since the intervention forces $Z$ to take a value different from the one it would take in the actual world. Such counterfactual variable is noted $Y_{Z=z}$ or $Y_z$ for short\footnote{The notations $Y_{Z\leftarrow z}$ and $Y(z)$ are used in the literature as well. $ \pr(Y=y | do(Z=z)) = \pr(Y_{Z=z} = y) = \pr(Y_z = y) = \pr(y_z)$ is used to define the causal effect of $z$ on $Y$.}. The term counterfactual quantity is used for expressions that involve explicitly multiple worlds. In Figure~\ref{subfig:figMark_b}, consider the expression $\pr(y_{a'}|Y=y, A=a) = \pr(y_{a'}|y,a)$. Such expression involves two worlds: an observed world where $A=a$ and $Y=y$ and a counterfactual world where $Y=y$ and $A=a'$ and it reads ``the probability of $Y=y$ had $A$ been $a'$ given that we observed $Y=y$ and $A=a$''. In the common example of job hiring, if $A$ denotes race ($a:$white, $a'$:non-white) and $Y$ denotes the hiring decision ($y$:hired, $y'$:not hired), $\pr(y_{a'}|y,a)$ reads ``given that a white applicant has been hired, what is the probability that the same applicant is still being hired had he been non-white''. 
		Nesting counterfactuals can produce complex expressions. For example, in the relatively simple model of Figure~\ref{subfig:figMark_b}, $\pr(y_{a,z_{a'}}|y'_{a'})$ 
		reads the probability of $Y=y$ had (1) $A$ been $a'$ and (2) $Z$ been $z$ when $A$ is $a'$, given that an intervention $A=a'$ produced $y'$. This expression involves three worlds: a world where $A=a$, a world where $Z=z_{a'}$, and a world where $A=a'$. Such complex expressions are used to characterize direct, indirect, and path-specific effects.

		Causal-based discrimination discovery aims at telling if the outcome of an automated decision making is fair or discriminative. Several causal-based fairness notions are defined in the literature (Section~\ref{sec:notions}) and expressed in terms of joint, conditional, interventional, and counterfactual probabilities. The application of a fairness notion requires as input a dataset $D$ and a causal graph $G$. While joint probabilities (e.g., $\pr(X=x,Y=y,Z=z)$) and conditional probabilities (e.g., $\pr(Y=y|X=x)$) can be trivially estimated from the dataset $D$, probabilities involving interventions or counterfactuals cannot always be estimated from $D$ and $G$. When a probability can be estimated from observable data ($D$), it is said to be \textit{identifiable}. Otherwise it is \textit{unidentifiable}. More formally, let $M_1$ and $M_2$ be two causal models sharing the same causal graph (not including the unobservable variable $\mathbf{U}$) and the same set of probability distributions $\psi$, a quantity $Q$ (e.g., intervention or counterfactual) is identifiable using $\psi$ (noted $\psi$-identifiable), if the value of $Q$ is unique  and computable from $\psi$ in any models $M_1$ and $M_2$. In other words, if there exists two models $M_1$ and $M_2$ sharing the same graph structure and the same probability distributions, but yielding different $Q$ values, then $Q$ is unidentifiable. Typically, the identifiability of interventional and counterfactual quantities depends on the structure of the graph, in particular, the location of the unobserved confounding variables. Identifiability criteria are summarized in Section~\ref{sec:ident}.

\subsection{Potential Outcome}
\label{subsec:po}
Unlike the SCM framework, expressing causal relations in the potential outcome framework starts at the unit level.
A unit $i$ is the atomic research object. In fairness problems, it typically refers to an individual. For example, in the job hiring scenario, every candidate corresponds to a unit $i$. Using the same example, the sensitive attribute of the candidate (e.g., gender) corresponds to the treatment in the potential outcome terminology. Given an outcome random variable $Y$, applying a treatment $A=a$ on a unit $i$ yields a different random variable called the potential outcome $Y_i(A=a) = Y_i^{a}$. For example, if $A=0$ refers to male, $A=1$ refers to female, and $Y$ is the hiring decision, $Y_i^{1}$ is the potential hiring decision of unit $i$ when the gender (treatment) is female. Consequently, if the treatment variable $A$ is binary, there are two potential outcomes $Y_i^{0}$ and $Y_i^{1}$. In observational studies (by contrast to experimental studies), only one potential outcome can be observed which is the factual outcome. The other potential outcome is usually impossible to observe and is called the counterfactual outcome. For example, if a job candidate $i$ is female ($A=1$) and is not hired, the potential outcome $Y^{1}_{i}$ is observed and is equal to $0$. However, the potential outcome of that candidate $i$ had she be male $Y^{0}_i$ is impossible to observe because this requires going back in time (impossible) and changing the sex of that individual to male (not ethical in the cases where it is possible). 

Causal inference in the potential outcome framework relies typically on three assumptions, namely, SUTVA, ignorability, and positivity~\cite{rubin2015book}. SUTVA (Stable Unit Treatment Value Assumption) has two requirements. First, the absence of interference among units. In the job hiring example it means that the hiring decision for a candidate is independent from the hiring decisions for all other candidates. Second, there is only one version of the treatment. This is more relevant in medical scenarios when a treatment (medication) has different versions (e.g., different dosage). For fairness scenarios, this requirement is typically satisfied as the treatment corresponds generally to an intrinsic attribute of the individual (e.g., gender, race, etc.). Ignorability is satisfied when the sensitive attribute $A$ and the potential outcome variables $Y^{0}$ and $Y^{1}$ are independent given observable variables $X$. That is, $A \perp Y^{0},Y^{1} | X$\footnote{Strong ignorability is a stronger assumptions requiring independence between the potential outcomes and any covariate $X$ ($X \perp Y^{0},Y^{1}$).}. This corresponds to absence of hidden (unobservable) confounders. In the SCM framework, it is equivalent to the graphical Markovian model requirement. Positivity assumption requires that the sensitive attribute is not deterministic with respect to other observable variables. That is, $\pr(A=a | X=x) > 0, \quad \forall a$, and $x$. In the job hiring example, any candidate can have any values of variables regardless of the gender $A$. 

\subsection{SCM and graphical models vs potential outcome}

Although both causal frameworks are considered equivalent~\cite{pearlBlog12}, interesting differences exist between them. Depending on the task at hand, one framework might be more appropriate to use than the other. For example, reasoning about causal effects at the individual (unit) level is more straightforward with the potential outcome framework~\cite{morgan2015book} (Section 3.4). On the other hand, considering the different paths of causal effects (direct, indirect, and spurious) is much easier achieved using SCMs and causal graphs. More generally, potential outcome framework is more suitable for causal inference problems where the goal is to narrowly estimate the causal (treatment) effect of a cause variable $A$ on an outcome variable $Y$. There are two justifications for this point. First, developing estimators of causal effects and counterfactuals can be more straightforward using the potential outcome framework~\cite{yao2020survey}. Second, the potential outcome framework provides the possibility of decomposing the sources of inconsistency and bias into: unaccounted-for baseline differences between individuals and treatment effect bias~\cite{morgan2015book} (Section 3.4).
SCMs and causal graphs, however, are more suitable in causal discovery problems where the goal is to learn the causal relations among a set of variables~\cite{glymour2019review}. Potential outcome framework is not well equipped for such problems because the causal effect of variables other than the treatment (sensitive attribute) are not defined.

\section{Causality-based Fairness notions}
\label{sec:notions}
Without loss of generality, assume that the sensitive attribute  $A$ and the outcome $Y$ are binary variables where $A=a_0$ denotes the privileged group (e.g. male), typically considered as the reference in characterizing discrimination, and $A=a_1$ the disadvantaged group (e.g. female).

Whenever needed, the simple job hiring example will be used where $A$ is the sensitive attribute corresponding to the gender ($A=0$ for male and $A=1$ for female), $C$ is a covariate corresponding to the job type ($C=0$ for flexible schedule job and $C=1$ for non-flexible job schedule), and $Y$ is the outcome corresponding to the hiring decision ($Y=0$ for not-hired and $Y=1$ for hired). 
Table~\ref{tab:hiringExample1} is an example dataset corresponding to this scenario.

\begin{table}[!h] 
	\centering
	\caption{A job hiring example with 24 applications. $A$ is the gender (sensitive attribute) where $A=1$: female, $A=0$: male. $C$ is the job type where $C=0$: flexible time job, $C=1$: non-flexible time job. $Y$ is the hiring decision (outcome) where $Y=0$: not-hired, $Y=0$: hired.}
\label{tab:hiringExample1} 
\begin{tabular}{ccccccccc}
\multicolumn{4}{c}{Female applicants} & & \multicolumn{4}{c}{Male applicants} \\
\multicolumn{4}{c}{(Treatment group)} & & \multicolumn{4}{c}{(Control Group)} \\
$i$ & $A$ & $C$ & $Y$ & & $i$ & $A$ & $C$ & $Y$ \\ \cline{1-4} \cline{6-9}
  1: & 1 & 0 & 1 & & 13: & 0 & 0 & 1\\
  2: & 1 & 0 & 1 & &  14: & 0 & 0 & 0\\
  3: & 1 & 0 & 0 & & 15: & 0 & 0 & 0\\
  4: & 1 & 0 & 0 & & 16: & 0 & 0 & 0\\
  5: & 1 & 0 & 0 & & 17: & 0 & 1 & 1\\
  6: & 1 & 0 & 0 & &  18: & 0 & 1 & 1\\
  7: & 1 & 0 & 0 & & 19: & 0 & 1 & 1\\
  8: & 1 & 0 & 0 & & 20: & 0 & 1 & 1\\
  9: & 1 & 1 & 1 & & 21: & 0 & 1 & 0\\
  10: & 1 & 1 & 1 & & 22: & 0 & 1 & 0\\
  11: & 1 & 1 & 1 & & 23: & 0 & 1 & 0\\
  12: & 1 & 1 & 0 & &  24: & 0 & 1 & 0\\
 \cline{1-4} \cline{6-9}
\end{tabular}  
\end{table}

The most common non-causal fairness notion is total variation (TV), known as statistical parity, demographic parity, or risk difference. The total variation of $A=a_1$ on the outcome $Y=y$ with reference $A=a_0$ is defined using conditional probabilities as follows:
\begin{equation}
\label{eq:TV}
TV_{a_1,a_0} (y) = \pr(y \mid a_1) - \pr(y \mid a_0)
\end{equation}
Intuitively, $TV_{a_1,a_0} (y)$ measures the difference between the conditional distributions of $Y$ when we (passively) observe $A$ changing from $a_0$ to $a_1$. In the example of Table~\ref{tab:hiringExample1}: $$TV = \pr(Y=1 \mid A=0) - \pr(Y=1 \mid A=1) = \frac{5}{12} - \frac{5}{12} = 0.$$ So according to $TV$, the predicted hiring decision is fair. The main limitation of $TV$ is its purely statistical nature which makes it unable to reflect the causal relationship between $A$ and $Y$, that is, it is insensitive to the mechanism by which data is generated and collected.  
Total effect ($TE$)~\cite{pearl2009causality}\footnote{Known also as average causal effect ($ACE$).} is the causal version of $TV$ and is defined in terms of experimental probabilities as follows: 
\begin{equation}
\label{eq:TE}
TE_{a_1,a_0} (y) = \pr(y_{a_1}) - \pr(y_{a_0})
\end{equation}
$TE$ measures the effect of the change of $A$ from $a_1$ to $a_0$ on $Y=y$ along all the causal paths from $A$ to $Y$. 
Intuitively, while $TV$ reflects the difference in proportions of $Y=y$ in the current cohort, $TE$ reflects the difference in proportions of $Y=y$ in the entire population. For the binary outcome case, $TE$ is equivalent to the average treatment effect ($ATE$)~\cite{morgan2015book} in the potential outcome framework which is defined as follows:
\begin{align}
\label{eq:ATE}
ATE_{a_1,a_0} & =  \ep[Y^{a_1} - Y^{a_0}]  \\ 
& =  \frac{1}{n} \sum_{i=1}^{n}(Y_i^{a_1} - Y_i^{a_0})
\end{align}
where $n$ is the number of observed samples. $ATE$ corresponds exactly to $FACE$ in~\cite{khademi2019fairness}.

Computing exactly $ATE$ requires the knowledge of both potential outcomes: the observed and the counterfactual. As the later is almost impossible to observe, exact computation of ATE is typically not possible. However, for the sake of illustration, we assume the counterfactual outcome is available, and show how ATE is computed. Later sections will show how $ATE$ and counterfactual outcomes can be estimated from observable data. Table~\ref{tab:hiringExample2}, shows the same job hiring dataset, but with counterfactual outcomes.

\begin{table}[!h] 
	\centering
	\caption{The job hiring example with counterfactual outcomes. $A^{cf}$ denotes the gender of the candidate in the counterfactual world. $Y^{cf}$ denotes the counterfactual potential outcome.}
\label{tab:hiringExample2} 
\begin{tabular}{ccccaacccccaa}
\multicolumn{6}{c}{Female applicants} & & \multicolumn{6}{c}{Male applicants} \\
\multicolumn{6}{c}{(Treatment group)} & & \multicolumn{6}{c}{(Control Group)} \\
$i$ & $A$ & $C$ & $Y$ & $A^{cf}$ & $Y^{cf}$ & & $i$ & $A$ & $C$ & $Y$ & $A^{cf}$ & $Y^{cf}$ \\ \cline{1-6} \cline{8-13}
  1: & 1 & 0 & 1 & 0 & 1 & & 13: & 0 & 0 & 1 & 1 & 1\\
  2: & 1 & 0 & 1 & 0 & 0 & &  14: & 0 & 0 & 0 & 1 & 1\\
  3: & 1 & 0 & 0 & 0 & 1 & & 15: & 0 & 0 & 0 & 1 & 0\\
  4: & 1 & 0 & 0 & 0 & 0 & & 16: & 0 & 0 & 0 & 1 & 0\\
  5: & 1 & 0 & 0 & 0 & 0 & & 17: & 0 & 1 & 1 & 1 & 1\\
  6: & 1 & 0 & 0 & 0 & 0 & &  18: & 0 & 1 & 1 & 1 & 1\\
  7: & 1 & 0 & 0 & 0 & 0 & & 19: & 0 & 1 & 1 & 1 & 1\\
  8: & 1 & 0 & 0 & 0 & 0 & & 20: & 0 & 1 & 1 & 1 & 1\\
  9: & 1 & 1 & 1 & 0 & 1 & & 21: & 0 & 1 & 0 & 1 & 1\\
  10: & 1 & 1 & 1 & 0 & 1 & & 22: & 0 & 1 & 0 & 1 & 0\\
  11: & 1 & 1 & 1 & 0 & 0 & & 23: & 0 & 1 & 0 & 1 & 0\\
  12: & 1 & 1 & 0 & 0 & 0 & &  24: & 0 & 1 & 0 & 1 & 0\\
 \cline{1-6} \cline{8-13}
\end{tabular}  
\end{table}

ATE is computed by considering the average potential outcome if the gender is female $A=1$, that is, $\frac{1}{n} \sum_{i=1}^{n}(Y_i^{1})$ and the same if the gender is male $A=0$, $\frac{1}{n} \sum_{i=1}^{n}(Y_i^{0})$. The former ($\sum_{i=1}^{n}(Y_i^{1})$) corresponds to the average of the observed outcomes ($Y$) of samples $1$ to $12$ and counterfactual outcomes ($Y^{cf}$) of samples $13$ to $24$, which gives $\frac{12}{24} = \frac{1}{2}$. Similarly, the average potential outcome if gender is male corresponds to the counterfactual outcomes of samples $1$ to $12$ and the observed outcomes of samples $13$ to $24$ which gives $\frac{9}{24} = \frac{3}{8}$. Hence, $ATE = \frac{1}{2} - \frac{3}{8} = \frac{1}{8}$ which indicates a positive bias for female. 

Computing the causal effect based only on the observed treatment group samples (e.g. female applicants only) corresponds to a variant of $TE$ called effect of treatment on the treated ($ETT$)~\cite{pearl2009causality} and is defined as:
\begin{equation}
\label{eq:ETT}
ETT_{a_1,a_0} (y) = \pr(y_{a_1} \mid a_1) - \pr(y_{a_0} \mid a_1)
\end{equation}
In the binary outcome case, $ETT$ corresponds to the average treatment effect on the treated $ATT$~\cite{morgan2015book} in the potential outcome framework defined as:
\begin{align}
\label{eq:ATT}
ATT_{a_1,a_0} & = \ep[Y^{a_1} | A=a_1] - \ep[Y^{a_0} | A=a_1] \\
& = \frac{1}{n_1} \sum_{i:A=a_1}(Y_i^{a_1} - Y_i^{a_0})
\end{align}
where $n_1$ is the number of samples in the treatment group. $ATT$ is also called $FACT$ in~\cite{khademi2019fairness}. 
In the example of Table~\ref{tab:hiringExample2}, $ATT$ corresponds to the difference between the average observable outcome ($Y$) and the average counterfactual outcome ($Y^{cf})$ in samples $1$ to $12$, that is, $ATT = \frac{5}{12} - \frac{4}{12} = \frac{1}{12}$, which confirms the positive bias for female. 

Average treatment effect on the control group ($ATC$)~\cite{morgan2015book} is the same as $ATT$ but focusing instead on the control group:
\begin{align}
\label{eq:ATC}
ATC_{a_1,a_0} & = \ep[Y^{a_1} | A=a_0] - \ep[Y^{a_0} | A=a_0] \\
& = \frac{1}{n_2} \sum_{i:A=a_0}(Y_i^{a_1} - Y_i^{a_0})
\end{align}
where $n_2$ is the number of samples in the control group.
Using the example of Table~\ref{tab:hiringExample2}, $ATC = \frac{7}{12} - \frac{5}{12} = \frac{1}{6}$. Conditional average treatment effect ($CATE$)~\cite{morgan2015book} is defined in a similar way, but conditioning on some other covariate instead of the sensitive attribute $A$:
\begin{align}
\label{eq:CATE}
CATE_{a_1,a_0}(X=x) & = \ep[Y^{a_1} | X=x] - \ep[Y^{a_0} | X=x] \\
& = \frac{1}{n_x} \sum_{i:X=x}(Y_i^{a_1} - Y_i^{a_0})
\end{align}
where $n_x$ is the number of samples in the subgroup $X=x$. Using the covariate $C=0$ (flexible schedule jobs) in the hiring example of Tabel~\ref{tab:hiringExample2}, $CATE(C=0) = \frac{4}{12} - \frac{3}{12} = \frac{1}{12}$, which is again confirming hiring decisions in favor of female.

Unlike the SCM framework, in the potential framework, it is possible to define individual treatment effect $ITE$~\cite{morgan2015book} which is defined, for every unit $i$ as:
\begin{equation}
\label{eq:ITE}
ITE_{a_1,a_0}(i) = Y_i^{a_1} - Y_i^{a_0}
\end{equation}
For instance, in Table~\ref{tab:hiringExample2}, $ITE(i=3) = 0 - 1 = -1$ which indicates a discrimination against the female applicant $i=3$. $ATC$, $CATE$, and $ITE$ are defined and typically used in the potential outcome framework but have no equivalents in the SCM framework. However, although $ATC$ and $CATE$ can be easily represented in the SCM formalism, $ITE$ cannot be easily formalized in the SCM framework.

The job hiring example of Tables~\ref{tab:hiringExample1} and~\ref{tab:hiringExample2} is interesting because it illustrates a statistical anomaly where some statistical notions such as $TV$ fail to appropriately account for the bias between sub-populations (e.g. female vs male). Notice first that, according to the collected data, both female and male candidates are hired at the same rate $\frac{5}{12}$. Notice also that if the hiring rates are adjusted according to the job type, female candidates are hired at an equal or higher rate for both types of jobs: for flexible schedule jobs  ($C=0$), the hiring rates are the same $\frac{1}{4}$ and for non-flexible jobs ($C=1$), the hiring rates are $\frac{3}{4}$ for female and $\frac{4}{8} = \frac{1}{2}$ for male. The explanation for such counter-intuitive result is that most of female candidates ($8$ out of $12$) are applying for flexible schedule jobs (for family reasons) in which hiring is more difficult. On the other hand, few male candidates ($4$ out of $12$) are applying for flexible schedule jobs, and instead massively applying for the more accessible non-flexible jobs ($8$ out of $12$ applicants). To appropriately assess discrimination in this case, there is a need to adjust on the job type variable $C$, that is, assessing discrimination for each job type separately. This simple job hiring scenario is similar to the Berkeley sex discrimination in college admission~\cite{berkeley75} where data showed a bias for male applicants overall, but when results were analyzed separately for each department, data showed a slight bias in favor of female candidates. The Berkeley scenario is typically used as an example of Simpson's paradox~\cite{simpson1951interpretation}. In both scenarios, 
by considering the outcome of the observable samples in the counterfactual setup, the above causal-based fairness notions could appropriately assess gender ($A$) discrimination on the outcome ($Y$). 
The job hiring example illustrating the statistical anomaly can be easily modified to reflect a Simpson's paradox~\cite{simpson1951interpretation}. Table~\ref{tab:hiringExample3} shows the same example but with 30 observed samples. In such cohort, $TV = -\frac{1}{6}$ indicates a discrimination against female applicants. However, all causal notions ($TE$ = $ATE$ = $\frac{1}{3}$, $ATT = \frac{1}{3}$, and $ATC = \frac{1}{3}$, $CATE(C=0) = \frac{1}{10}$, and $CATE(C=1) = \frac{2}{10}$) are indicating a bias in favor of female.

\begin{table}[!h] 
	\centering
	\caption{The job hiring example with a Simpson's paradox.}
\label{tab:hiringExample3} 
\begin{tabular}{ccccaacccccaa}
\multicolumn{6}{c}{Female applicants} & & \multicolumn{6}{c}{Male applicants} \\
\multicolumn{6}{c}{(Treatment group)} & & \multicolumn{6}{c}{(Control Group)} \\
$i$ & $A$ & $C$ & $Y$ & $A^{cf}$ & $Y^{cf}$ & & $i$ & $A$ & $C$ & $Y$ & $A^{cf}$ & $Y^{cf}$ \\ \cline{1-6} \cline{8-13}
  1: & 1 & 0 & 1 & 0 & 1 & & 16: & 0 & 0 & 1 & 1 & 1\\
  2: & 1 & 0 & 1 & 0 & 1 & & 17: & 0 & 0 & 0 & 1 & 1\\
  3: & 1 & 0 & 1 & 0 & 0 & & 18: & 0 & 0 & 0 & 1 & 0\\
  4: & 1 & 0 & 0 & 0 & 0 & & 19: & 0 & 0 & 0 & 1 & 0\\
  5: & 1 & 0 & 0 & 0 & 0 & & 20: & 0 & 0 & 0 & 1 & 0\\
  6: & 1 & 0 & 0 & 0 & 0 & & 21: & 0 & 1 & 1 & 1 & 1\\
  7: & 1 & 0 & 0 & 0 & 0 & & 22: & 0 & 1 & 1 & 1 & 1\\
  8: & 1 & 0 & 0 & 0 & 0 & & 23: & 0 & 1 & 1 & 1 & 1\\
  9: & 1 & 0 & 0 & 0 & 0 & & 24: & 0 & 1 & 1 & 1 & 1\\
  10: & 1 & 0 & 0 & 0 & 0 & & 25: & 0 & 1 & 1 & 1 & 1\\
  11: & 1 & 1 & 1 & 0 & 1 & & 26: & 0 & 1 & 1 & 1 & 1\\
  12: & 1 & 1 & 1 & 0 & 1 & & 27: & 0 & 1 & 1 & 1 & 1\\
  13: & 1 & 1 & 1 & 0 & 1 & & 28: & 0 & 1 & 0 & 1 & 1\\
  14: & 1 & 1 & 1 & 0 & 0 & & 29: & 0 & 1 & 0 & 1 & 0\\
  15: & 1 & 1 & 0 & 0 & 0 & &  30: & 0 & 1 & 0 & 1 & 0\\
 \cline{1-6} \cline{8-13}
\end{tabular}  
\end{table}
All the above causal-based fairness notions fall into the framework of disparate impact~\cite{barocas2016big} which aims at ensuring the equality of outcomes among all groups (protected/treatment and unprotected/control). An alternative framework is the disparate treatment~\cite{barocas2016big} which seeks equality of treatment achievable through prohibiting the use of the sensitive attribute in the decision process. The main idea is to split the causal effect between the sensitive attribute $A$ and the outcome $Y$ into several causal pathways, each of which is either fair, unfair, or spurious.
Common fairness notions from the disparate treatment framework include direct effect, indirect effect, and path-specific effect~\cite{pearl01direct}. An effect can be deemed fair, unfair, or spurious by an expert of the scenario at hand. Unfair effect is called discrimination. Direct discrimination is assessed using causal effect along direct edge from $A$ to $Y$. Indirect discrimination is measured using the  causal effect along causal paths that pass through proxy attributes\footnote{A proxy is an attribute that cannot be objectively justified if used in the decision making process. It is Known also as redlining attribute.}. A fair or explainable discrimination is measured using causal pathways passing through explaining variables. Spurious effect corresponds to a pathway starting with an incident edge into the sensitive attribute $A$.
\begin{figure}[!h]
	\centering
	{\includegraphics[scale=0.3]{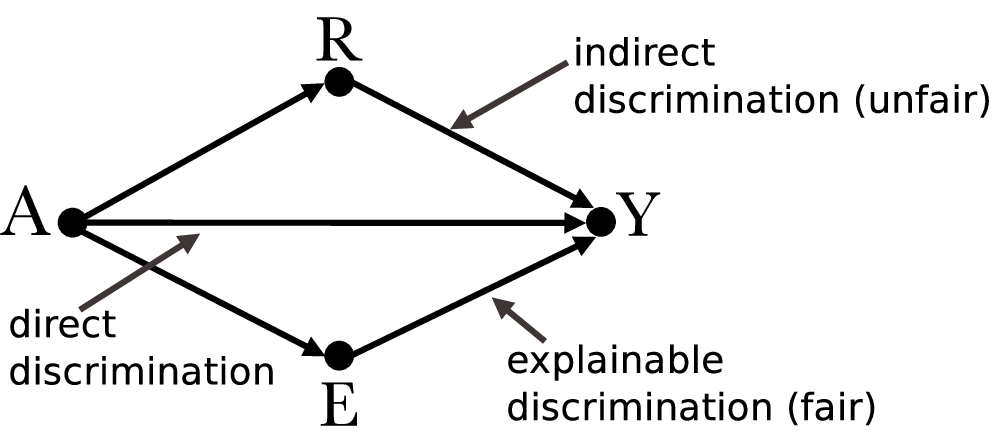}}
	\caption{Job hiring scenario where $A$ is gender, $Y$: hiring decision, $R$: hobby of a candidate ($R=1$ for mechanical hobby, $R=0$ for non-mechanical hobby), and $E$: education level of the candidate ($E=1$ for college degree, $E=0$ for no college degree).}
	\label{fig:hiringExampleGraph1}
\end{figure}

Figure~\ref{fig:hiringExampleGraph1} presents a causal graph of the job hiring scenario involving an explaining variable $E$ (e.g. education and academic degrees), and a proxy/redlining variable $R$ (e.g. the hobby of the candidate). Hiring discrimination due to education level is legitimate and considered fair, whereas a discrimination due to the hobby of the candidate is unfair as it is a proxy for the gender (the type of hobby indicates generally the gender of the candidate). Direct effect can be computed by simply ``blocking'' all indirect causal paths. An indirect causal path is a directed path from $A$ to $Y$ going through one or several mediator variables. For example, in Figure~\ref{fig:hiringExampleGraph1}, there are two indirect causal paths $A\;\longrightarrow\;R\;\longrightarrow\;Y$ and $A\;\longrightarrow\;E\;\longrightarrow\;Y$. To compute the direct causal effect ($A\;\longrightarrow\;Y$), both indirect causal paths need to be blocked by adjusting on variables $R$ and $E$. As there are no confounders, the direct effect can be simply computed as:
\begin{align}
DE_{a_1,a_0} (y) & = \pr(y\;|\;a_1,R,E) - \pr(y\;|\;a_0,R,E) \nonumber\\
                & = \sum_{r}\sum_{e}\left(\pr(y\;|\;a_1,r,e) - \pr(y\;|\;a_0,r,e)\right) \nonumber
\end{align}
In presence of confounders (between $A$ and $Y$, between $R$ and $Y$, etc.), natural direct effect ($NDE$)~\cite{pearl01direct} is a more general notion that measures the direct causal effect and is defined as: 
\begin{equation}
\label{eq:NDE}
NDE_{a_1,a_0} (y) = \pr(y_{{a_1},\mathbf{Z}_{a_0}}) - \pr(y_{a_0})
\end {equation}
Where $\mathbf{Z}$ is the set of mediator variables and $\pr(y_{{a_1},\mathbf{Z}_{a_0}})$ is the probability of $Y=y$ had $A$ been $a_1$ and had $\mathbf{Z}$ been the value it would naturally take if $A=a_0$. That is, $A$ is set to $a_1$ in the single direct path $A \rightarrow Y$ and is set to $a_0$ in all other indirect paths ($A\rightarrow R \rightarrow Y$ and $A \rightarrow E \rightarrow Y$). 
To see how $NDE$ is computed, consider the sample dataset in Table~\ref{tab:hiringExample4} corresponding to the causal graph in Figure~\ref{fig:hiringExampleGraph1}. Similarly to the previous examples, we assume the counterfactual values are available (grayed columns).
\begin{table}[!h] 
	\centering
	\caption{A job hiring scenario corresponding to the causal graph in Figure~\ref{fig:hiringExampleGraph1}.}
\label{tab:hiringExample4} 
\begin{tabular}{p{0.1cm}p{0.1cm}p{0.1cm}p{0.1cm}p{0.1cm}aaaeaaee}
$i$ & $A$ & $E$ & $R$ & $Y$ & $Y^{cf}$ & $E_{0}$ & $R_{0}$ & $Y_{1,E_{0},R_{0}}$ & $E_{1}$ & $R_{1}$ & $Y_{0,E_{1},R_{1}}$ & $Y_{1,E_{0},R_{1}}$ \\ \hline 
  1: & 1 & 1 & 0 & 1 & 1 & 1 & 1 & 1 & 1 & 0 & 1 & 1\\
  2: & 1 & 1 & 0 & 1 & 1 & 1 & 1 & 1 & 1 & 0 & 1 & 1\\
  3: & 1 & 1 & 1 & 0 & 1 & 0 & 1 & 1 & 1 & 1 & 1 & 0\\
  4: & 1 & 0 & 0 & 1 & 1 & 1 & 0 & 1 & 0 & 0 & 1 & 1\\
  5: & 1 & 0 & 0 & 0 & 1 & 0 & 1 & 0 & 0 & 0 & 1 & 1\\
  6: & 1 & 0 & 0 & 0 & 0 & 0 & 0 & 0 & 0 & 0 & 0 & 0\\
  7: & 0 & 1 & 1 & 1 & 1 & 1 & 1 & 1 & 1 & 0 & 1 & 1\\
  8: & 0 & 1 & 0 & 1 & 1 & 1 & 0 & 1 & 1 & 0 & 1 & 1\\
  9: & 0 & 1 & 1 & 1 & 0 & 1 & 1 & 0 & 1 & 1 & 0 & 0\\
  10: & 0 & 0 & 1 & 1 & 1 & 0 & 1 & 1 & 1 & 0 & 1 & 1\\
  11: & 0 & 0 & 1 & 0 & 1 & 0 & 1 & 1 & 0 & 0 & 1 & 0\\
  12: & 0 & 0 & 1 & 0 & 0 & 0 & 1 & 0 & 0 & 0 & 1 & 1\\
 \hline
\end{tabular}  
\end{table}
The cohort consists of $6$ female candidates and $6$ male candidates. $Y^{cf}$ is the counterfactual potential outcome (the gender is different from the observed sample). $E_0$ is the education level had the gender was male. $R_0$ is the hobby of the candidate had the gender was male. $Y_{1,E_{0},R_{0}}$ is the hiring decision had (1) the gender was female and (2) the education and hobby were set to the values if the candidate was male. According to Equation~\ref{eq:NDE}, $NDE_{1,0}(y=1) = \pr(y_{1,E_{0},R_{0}}) - \pr(y_0) = \frac{8}{12} - \frac{9}{12} = -\frac{1}{12}$ which indicates a direct discrimination against female candidates.
Notice the following:
\begin{itemize}
    \item For rows $7$ to $12$, the values of columns $E_0$, $R_0$, and $Y_{1,E_{0},R_{0}}$ are equal to the values in columns $E$, $R$, and $Y^{cf}$, respectively.
    \item $\pr(y_0)$ is computed based on the values in rows $1$ to $6$ of column $Y^{cf}$ and rows $7$ to $12$ of column $Y$.
\end{itemize}

Natural indirect effect ($NIE$)~\cite{pearl01direct} measures the indirect effect of $A$ on $Y$ and is defined as:
\begin{equation}
\label{eq:NIE}
NIE_{a_1,a_0} (y) = \pr(y_{{a_0},\mathbf{Z}_{a_1}}) - \pr(y_{a_0})
\end {equation}
In the example of Table~\ref{tab:hiringExample4}, 
$NIE_{1,0}(y=1) = \pr(y_{0,E_{1},R_{1}}) - \pr(y_0) = \frac{9}{12} - \frac{9}{12} = 0$

The problem with $NIE$ is that it does not distinguish between the fair (explainable) and unfair (indirect discrimination) effects. Path-specific effect~\cite{pearl2009causality,chiappa2019path,wu2019pc} is a more nuanced measure that characterizes the causal effect in terms of specific paths.

Given a path set $\pi$, the $\pi$-specific effect is defined as: 
\begin{equation}
\label{eq:PSE}
PSE^{\pi}_{a_1,a_0}(y) = \pr(y_{a_1 |_\pi, a_0 |_{\overline{\pi}}}) - \pr(y_{a_0})
\end{equation}
where $\pr(y_{a_1 |_\pi, a_0 |_{\overline{\pi}}})$ is the probability of $Y=y$ in the counterfactual situation where the effect of $A$ on $Y$ with the intervention ($a_1$) is transmitted along $\pi$, while the effect of $A$ on $Y$ without the intervention ($a_0$) is transmitted along paths not in $\pi$ (denoted by: $\overline{\pi}$). Using the job hiring example of Figure~\ref{fig:hiringExampleGraph1}, Eq.~\ref{eq:PSE} can be used to assess only unfair discrimination which is transmitted through the direct path $A\rightarrow Y$ and the indirect path $A\rightarrow R \rightarrow Y$. The third path $A \rightarrow E \rightarrow Y$ transmits explainable (fair) discrimination, and hence, should not be considered. Given $\pi = \{A\rightarrow Y, \;\;A\rightarrow R \rightarrow Y\}$, $PSE^{\pi}_{1,0} = \pr(Y_{1,E_{0},R_{1}}) - \pr(y_0) = \frac{8}{12} - \frac{9}{12} = - \frac{1}{12}$ which indicates a discrimination against female candidates.

\subsection{No unresolved discrimination}
\label{sec:nounresolved}
\begin{figure}[!h]
	\centering
	\subfigure{%
	{\includegraphics[scale=0.2]{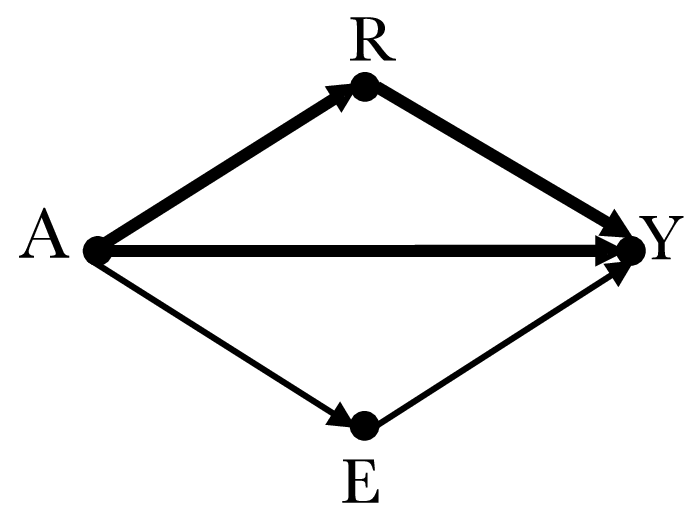}}
	\label{fig:unresolved1}}
	\qquad \qquad
	\subfigure{%
	{\includegraphics[scale=0.2]{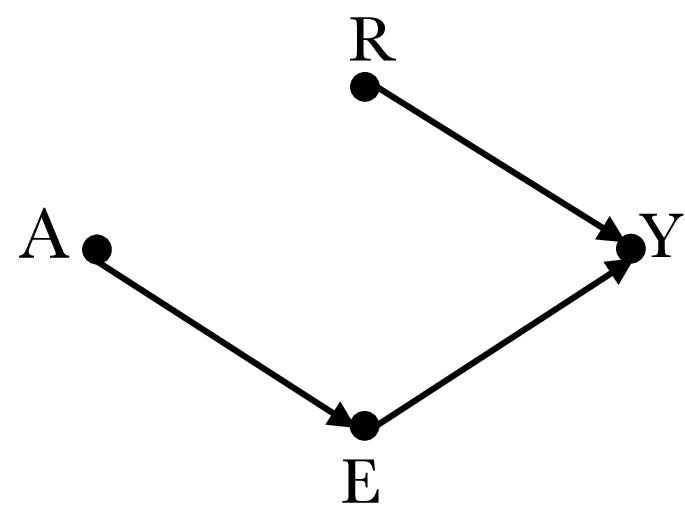}}
	\label{fig:unresolved2}}
	\caption{$Y$ exhibits unresolved discrimination in the left graph (along the heavy paths), but not the right one.}
	\label{fig:unresolved}
\end{figure}

No unresolved discrimination~\cite{kilbertus2017avoiding} is 
a fairness notion that falls into the disparate treatment framework and focuses on  the indirect causal effects from $A$ to $Y$. 
No unresolved discrimination is satisfied when no directed path from $A$ to $Y$ is allowed, except via a resolving (explaining) variable $E$. A resolving variable is any variable in a causal graph that is influenced by the sensitive attribute in a manner that is accepted as nondiscriminatory. Figure~\ref{fig:unresolved} presents two alternative causal graphs for the job hiring example. The graph at the left exhibits unresolved discrimination along the heavy paths: $A \rightarrow R \rightarrow Y$ and $A \rightarrow Y$. By contrast, the graph at the right does not exhibit any unresolved discrimination as the effect of $A$ on $Y$ is justified by the resolved variable $E$: $ A \rightarrow E \rightarrow Y$. 

The use of no unresolved discrimination in real scenarios is limited by the assumption of valid causal graph availability. \cite{kilbertus2017avoiding} provide a formal proof that even with prior knowledge of resolving variables, it is not always possible to tell, based on observational data only, if a predictor $Y$ satisfies no unresolved discrimination.   
	
\subsection{No proxy discrimination}
\label{sec:noproxy}
Similarly to no unresolved discrimination, no proxy discrimination~\cite{kilbertus2017avoiding} focuses on indirect discrimination. A causal graph exhibits potential proxy discrimination if there exists a path from the protected attribute $A$ to the outcome $Y$ that is blocked by a proxy/redlining variable $R$. It is called proxy because it is used to decide about the outcome $Y$ while it is a descendent of $A$ which is significantly correlated with it in such a way that using the proxy in the decision has almost the same impact as using $A$ directly. An outcome variable $Y$ exhibits no proxy discrimination if the equality: 
\begin{equation}
\label{eq:proxy}
\pr(Y\mid do(R=r)) = \pr(Y\mid do(R=r')) \quad \forall \; r, r'\in dom(R)
\end{equation}
holds for any potential proxy $R$.

\begin{figure}[!h]
		\vspace{-2mm}
    \centering
    \subfigure {%
{\includegraphics [scale=0.25]{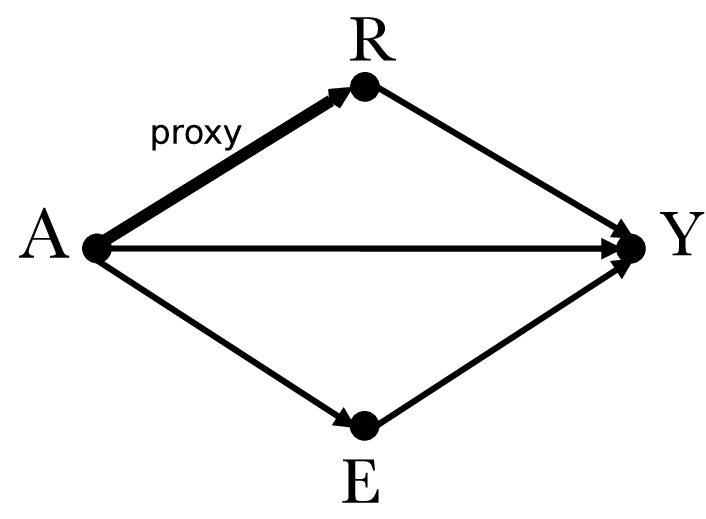} }
\label{fig:proxy1}}
    \qquad \qquad
    \subfigure {%
    {\includegraphics[scale=0.25]{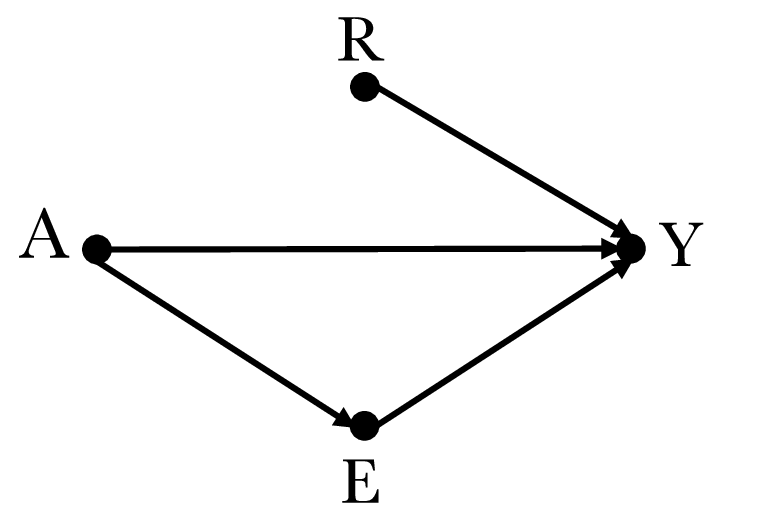} }
    \label{gig:proxy2}}
    \caption{The graph at the left exhibits a potential proxy discrimination (along the heavy edge between $A$ and $R$) but not in the right one.}
    \label{fig:proxy}
    		\vspace{-2mm}
\end{figure}

Figure~\ref{fig:proxy} shows two similar causal graphs for the same job hiring example. The causal graph at the left presents a potential proxy discrimination via the path: $A \rightarrow R \rightarrow Y$. However, the graph at the right is free of proxy discrimination as the edge between $A$ and its proxy $R$ has been removed due to the intervention on $R$ ($R=r$).

Similarly to no unresolved discrimination, no proxy discrimination requires a valid causal graph. Hence, both fairness notions depend on the correct output of the causal discovery task.

\subsection{Counterfactual fairness}
\label{sec:counterfactual}
		\vspace{-1mm}
Counterfactual fairness~\cite{kusner2017counterfactual} is a very strong fairness notion that requires equality between the observed outcome and the counterfactual outcome for every individual. That is, an outcome $Y$ is counterfactually fair if under any assignment of values $\mathbf{X}=\mathbf{x}$ and any individual in $U$,
\begin{equation}
\label{eq:counterfactual}
\pr(y_{a_1}(U) \mid \mathbf{X} =  \mathbf{x}, A =  a_0) = \pr(y_{a_0}(U) \mid \mathbf{X} = \mathbf{x}, A =  a_0) 
\end{equation}
where $\mathbf{X} = \mathbf{V}\backslash\{A,Y\}$ is the set of all remaining variables. As the latent variable $U$ appears in  Equation~\ref{eq:counterfactual}, counterfactual fairness is an individual fairness notion. It is satisfied if the probability distribution of the outcome $Y$ is the same in the actual and counterfactual worlds, for every possible individual. In practice, counterfactual fairness coincides typically with $ITE$ (Eq.~\ref{eq:ITE}).

\cite{kusner2017counterfactual} could test counterfactual fairness by making a very strong assumption. That is, they assumed the full structure of the causal model is available including the latent variables $U$. They could then estimate the distribution of $\pr(U)$ using Markov chain Monte Carlo methods and the observed data. Thanks to the estimated distribution of $\pr(U)$, they could compute counterfactuals using Pearl's three-step process: abduction, action, and prediction~\cite{pearl2009causality}. Hence, for every individual in the population, another sample with counterfactual sensitive value is generated. Counterfactual fairness is finally assessed by comparing the density functions of the actual and counterfactual samples. The process of testing counterfactual fairness is detailed in Section~\ref{subsec:ctf}.

\subsection{Counterfactual Effects}
\label{sec:ctf}
		\vspace{-1mm}
By conditioning on the sensitive attribute $A=a$, Zhang and Bareinboim~\cite{zhang2018fairness} defined two variants of $NDE$ (Eq.~\ref{eq:NDE}) and $NIE$ (Eq.~\ref{eq:NIE}) which focus on the direct and indirect effect for a specific group. In addition, they characterize a third type of effect, spurious, which considers the back-door paths between $A$ and $Y$, that is, paths with an arrow into $A$. 

The three effects are defined as follows:
\begin{align}
\label{eq:ctfde}
&DE_{a_1,a_0} (y|a) = \pr(y_{{a_1},\mathbf{Z}_{a_0}} |a) - \pr(y_{a_0} |a) \\
\label{eq:ctfie}
&IE_{a_1,a_0} (y|a) = \pr(y_{{a_0},\mathbf{Z}_{a_1}} |a) - \pr(y_{a_0} |a) \\
\label{eq:ctfse}
&SE_{a_1,a_0} (y) = \pr(y_{a_0} |a_1) - \pr(y|a_0) 
\end {align}
where in Eq.~\ref{eq:ctfde} and~\ref{eq:ctfie}, $a$ can be $a_0$ or $a_1$. Considering the simple job hiring example and focusing on the female group ($A=1$), $DE_{1,0}(y|1)$ measures the change in the probability of $Y$ (e.g. hiring) had $A$ been $1$ (female), while mediators $E$ and $R$ are kept at the level they would take had $A$ been $0$ (male).
Using the values in Table~\ref{tab:hiringExample4}, $DE_{1,0} = \pr(y_{1,E_{0},R_{0}} | 1) - \pr(y_0 | 1) = \frac{4}{6} - \frac{5}{6} = -\frac{1}{6}$ which indicates a a direct counterfactual discrimination against  female. Similarly, $IE_{1,0} = \pr(y_{0,E_{1},R_{1}} | 1) - \pr(y_0 | 1) = \frac{5}{6} - \frac{5}{6} = 0$ which indicates the absence of counterfactual indirect discrimination. $SE_{1,0}(y)$ reads the change in the probability of hiring $Y$ had $A$ been $0$ (male) for the female candidates with respect to the probability of hiring of male candidates. Using Table~\ref{tab:hiringExample4}, $SE_{1,0}(y) = \pr(y_0|1) - \pr(y|0) = \frac{5}{6} - \frac{4}{6} = \frac{1}{6}$ which indicates a spurious effect in favor of female. 
Compared to $NDE$ and $NIE$, counterfactual effects focus only on individuals of a specific group (e.g. only female candidates) and characterize the causal effect through spurious (back-door) paths. This spurious effect is what makes causal relations different from mere statistical correlation. However, counterfactual indirect effect $IE$ still does not distinguish between fair and unfair direct effects.

\subsection{Counterfactual Error Rates}
\label{sec:cer}
Equalized odds~\cite{hardt2016equality} is an important statistical fairness notion which requires equality of error rates ($TPR$ and $FPR$) across sub-populations, that is, 
\begin{equation}
	ER_{a_1,a_0}(\hat{y}|y) = \pr(\hat y \mid a_1,y) - \pr(\hat y \mid a_0,y) = 0 \label{eq:eqodds}
\end{equation}
where $\hat{y}$ denotes the prediction while $y$ denotes the true outcome.
The problem of this statistical notion is the difficulty to identify the causes behind the discrimination if any. \cite{zhang2018equality} decompose equalized odds (Eq.~\ref{eq:eqodds}) using three counterfactual measures corresponding to the direct, indirect and spurious effects of $A$ on $\hat{Y}$. The three measures are counterfactual direct error rate, counterfactual indirect error rate, and counterfactual spurious error rate. Let $\hat{y} = f(\hat{\mathbf{pa}})$ be a classifier where $\hat{\mathbf{PA}}$ is the set of input features (parent variables of $\hat{Y}$) for the classifier. The counterfactual error rates for a sub-population $a,y$ (with prediction $\hat{y}\neq y$) are defined as:
\begin{align}
\label{eq:cder}
&ER^d_{a_1,a_0}(\hat  y \mid a,y) = \pr(\hat y_{a_1,y,(\hat{\textbf{PA}}\setminus A)_{{a_0},y}}\mid a,y) - \pr(\hat y_{{a_0},y}\mid a,y))\\
\label{eq:cier}
&ER^i_{a_1,a_0}(\hat  y \mid a,y) = \pr(\hat y_{a_0,y,(\hat{\textbf{PA}}\setminus A)_{a_1,y}}\mid a,y) - \pr(\hat y_{a_0,y}\mid a,y))\\
\label{eq:cser}
&ER^s_{a_1,a_0}(\hat  y \mid y) = \pr(\hat y_{{a_0},y} \mid a_1,y) - \pr(\hat y_{{a_0},y}\mid a_0,y)
\end{align}

For example, the counterfactual direct error rate (Eq.~\ref{eq:cder}) measures the error rate (disparity between the true and the predicted outcome) in terms of the direct effects of the sensitive attribute $A$ on the prediction $\hat{Y}$. In the job hiring example, considering the female sub-population that \textit{should} be hired ($A=1$ and $Y=1$), it reads: for a female candidate that should be hired, how would the prediction $\hat Y$ change had the candidate been a female ($A$ been $1$), while keeping all the other features $\hat{\textbf{PA}}\setminus A$ at the level that they would attain had ``she was male'', compared to the prediction $\hat Y$ she would receive had ``she was male'' and should have been hired?

\begin{table}[!h] 
	\centering
	\caption{A job hiring scenario for counterfactual direct error rate $ER^{d}$ computation. $E_{0,1}$ is a short version of $E_{A=0,Y=1}$. $R_{0,1}$ means $R_{A=0,Y=1}$. $\hat{Y}_{1,1,E_{0,1},R_{0,1}}$ means $\hat{Y}_{A=1,Y=1,E_{0,1},R_{0,1}}$. $Y_{0,1}$ means $Y_{A=0,Y=1}$.}
\label{tab:hiringExample5} 
\begin{tabular}{ccccccaaaa}
$i$ & $A$ & $E$ & $R$ & $\hat{Y}$ & $Y$ & $E_{0,1}$ & $R_{0,1}$ & $\hat{Y}_{1,1,E_{0,1},R_{0,1}}$ & $\hat{Y}_{0,1}$ \\ \hline 
  1: & 1 & 1 & 0 & 1 & 1 & 1 & 0 & 1 & 1 \\
  2: & 1 & 1 & 0 & 1 & 1 & 1 & 0 & 1 & 1 \\
  3: & 1 & 1 & 1 & 0 & 1 & 0 & 1 & 1 & 1 \\
  4: & 1 & 0 & 0 & 1 & 1 & 1 & 0 & 1 & 0 \\
  5: & 1 & 0 & 0 & 0 & 1 & 0 & 0 & 0 & 0 \\
  6: & 1 & 0 & 0 & 0 & 0 & 0 & 0 & 0 & 0 \\
  7: & 0 & 1 & 1 & 1 & 1 & 1 & 1 & 1 & 1 \\
  8: & 0 & 1 & 0 & 1 & 1 & 1 & 0 & 1 & 1 \\
  9: & 0 & 1 & 1 & 1 & 0 & 0 & 1 & 0 & 1 \\
  10: & 0 & 0 & 1 & 1 & 1 & 0 & 1 & 0 & 0  \\
  11: & 0 & 0 & 1 & 0 & 1 & 0 & 1 & 1 & 0 \\
  12: & 0 & 0 & 1 & 0 & 0 & 0 & 1 & 0 & 0 \\
 \hline
\end{tabular}  
\end{table}

Table~\ref{tab:hiringExample5} shows the values (observed and counterfactual) needed to compute counterfactual direct error rate $ER^{d}$ for the female candidates that should be hired ($A=1$ and $Y=1$). 
\begin{align}
ER^{d}(\hat{Y}=1|A=1,Y=1)  & =  \pr(\hat{Y}_{A=1,Y=1,E_{0,1},R_{0,1}}|A=1,Y=1) \nonumber \\ & - \pr(\hat{Y}_{A=0,Y=1}|A=1,Y=1) \nonumber
\end{align}
where $E_{0,1}$ is a short version of $E_{A=0,Y=1}$ which refers to the education level of the candidate had ``she'' been male and hired. $R_{0,1}$ means $R_{A=0,Y=1}$ and indicates the hobby of the candidate had ``she'' been male and hired.  $\hat{Y}_{A=1,Y=1,E_{0,1},R_{0,1}}$ reads the hiring decision had the candidate was female, hired, with education $E_{0,1}$, and hobby $R_{0,1}$. $Y_{A=0,Y=1}$ reads the hiring decision had the candidate was male and hired. Using the values in Table~\ref{tab:hiringExample5} (rows $1$ to $5$ in the last two columns), $ER^{d}(\hat{Y}=1|A=1,Y=1) = \frac{4}{5} - \frac{3}{5} = \frac{1}{5}$ which indicates a higher direct error rate for the female group.  

Interestingly, the statistical equalized odd error rate (Eq.~\ref{eq:eqodds}) can be decomposed in terms of the three above causal-based error rates:
\begin{equation}
\label{eq:expFormula_ER}
ER_{a_1,a_0} (\hat y \mid y) = ER^d_{a_1,a_0} (\hat y \mid a_0,y) - ER^i_{a_0,a_1} (\hat y \mid a_0,y) - ER^s_{a_0,a_1} (\hat y \mid y)
\end{equation}

\subsection{Individual direct discrimination}
\label{sec:idd}
Individual direct discrimination~\cite{zhang2016situation} aims to discover the direct discrimination at the individual level. It is based on situation testing~\cite{bendick2007situation}, a legally grounded technique for analyzing the discrimination at an individual level. It consists in comparing the individual with similar individuals from both groups (protected and unprotected). That is, for an individual $i$ in question, find the $k$ other individuals which are the most similar to $i$ in the group $A=a_0$ and $k$ similar individuals from the group $A=a_1$. The first set is denoted as $\mathbf{S}^+$ while the second as $\textbf{S}^-$. 
The target individual is considered as discriminated if the difference observed between the rate of positive decisions in $\textbf{S}^-$ and $\textbf{S}^+$ is higher than a predefined threshold $\tau$ (typically $5\%$).

Causal inference is used to define the distance function $d(i,i')$ required to select the elements of $\textbf{S}^-$ and $\textbf{S}^+$. First, only attributes that are direct causes of the outcome should considered in the computation of the distance. That is, based on the causal graph, $\mathbf{Q}=Pa(Y)\backslash\{A\}$ denotes the set of variables that should be used in the distance function. Second, the causal effect of each of the selected attributes ($Q_k \in \mathbf{Q}$) on the the outcome should be considered in the function definition. In particular, for each variable $Q_k$, $CE(q_k, q'_k)$ measures the causal effect on the outcome when the value of $Q_k$ changes from $q_k$ to $q'_k$ and is defined as:

\begin{equation}
	CE(q_k, q'_k) = \pr(y_{ \mathbf{q}}) - \pr(y_{q'_k, \mathbf{q} \setminus \{q_k\}}) \label{eq:iddd}
\end{equation}
where $(\pr(y_{\mathbf{q}}))$ is the effect of the intervention that forces the set $\mathbf{Q}$ to take the set of values $\mathbf{q}$, and  $(\pr(y_{q'_k, \mathbf{q} \setminus \{q_k\}}))$ is the effect of the intervention that forces $Q_k$ to take value $q'_k$ and other attributes in $\mathbf{Q}$ to take the same values as $\mathbf{q}$.

The two individual fairness notions mentioned above, namely, $ITE$ (Equation~\ref{eq:ITE}) and counterfactual fairness (Section~\ref{sec:counterfactual}) rely on the counterfactual outcome to assess fairness for every individual. Individual direct discrimination drops this requirement and use instead the sets $\textbf{S}^-$ and $\textbf{S}^+$ composed of similar individuals in both groups. Hence, it can be considered as an estimation technique to circumvent the need for counterfactuals.
However, the distance function between two individuals $d(i,i')$ is unnecessarily complex; it is defined in terms of the causal effects of every covariate $X$ on the outcome $Y$. These causal effects are re-computed each time the distance between two individuals is needed. Matching techniques in the potential outcome framework use much simpler distance metrics. Matching techniques are discussed in Section~\ref{subsec:matching}.

\subsection{Non-Discrimination Criterion}
\label{sec:ndc}

Non-discrimination criterion~\cite{zhang2017achieving} is a group fairness notion that aims to discover and to quantify direct discrimination through the direct causal effect of $A$ on $Y$. Recall that, given a causal graph $G$, a direct effect of $A$ on $Y$ is the causal effect through the edge $A \rightarrow Y$. The idea is to consider a modified graph $G'$ where the edge in question ($A \rightarrow Y$) is discarded. A \textit{block set} $\mathbf{Q}$ is a set of variables which blocks all causal effects from $A$ to $Y$ in the modified graph $G'$. Hence, $A$ and $Y$ are independent conditioning on $Q$ in $G'$, that is, $(A \perp Y | \mathbf{Q})_{G'}$. Hence, conditioning on the same variables $\mathbf{Q}$, any dependence between $A$ and $Y$ in $G$ is due to the direct effect of $A$ on $Y$ which indicates a direct discrimination. This discrimination can be assessed using simply the risk difference~\cite{romei2011multidisciplinary}:
\begin{equation}
\label{eq:ndc}
\mid\;\Delta P\!\!\mid_{\mathbf{q}}\; \mid = \mid \pr(y\mid a_1, \mathbf{q}) - \pr(y\mid a_0, \mathbf{q}) \mid
\end{equation}
where $\mathbf{q}$ is a value assignment for the block set $\mathbf{Q}$ and the absolute value to consider both positive and negative discriminations. No direct discrimination can be concluded if the risk difference is less than a threshold $\tau$ for all combinations of values of all block sets, that is, Eq.~\ref{eq:ndc} holds for each value assignment $q$ of each block set $\mathbf{Q}$.

$NDE$ (Equation~\ref{eq:NDE}) and counterfactual direct effect $DE$ (Section~\ref{sec:ctf}) focus also on assessing the direct discrimination, but they both rely on nested counterfactual quantities which are not observable from data. Non-discrimination criterion circumvents this difficulty by using block sets and considering all combinations of values of these block sets. Similarly to individual direct discrimination, it can be considered as an estimation technique to avoid dealing with counterfactual quantities. This approach, however, does not work in semi-markovian models as $A$ and $Y$ will never be independent in $G'$ ($(A \perp Y | \mathbf{Q})_{G'}$) because of hidden counfounders.

\subsection{Equality of Effort}
\label{sec:ftee}
Equality of effort~\cite{huan2020fairness} fairness notion identifies discrimination by assessing how much effort is needed by the disadvantaged individual/group to reach a certain level of outcome. 
A treatment variable $T$ is selected and used to address the question: ``to what extent the treatment variable $T$ should change to make the individual (or a group of individuals) achieve a certain outcome level?''. Hence, this notion focuses on whether the effort to reach a certain outcome level is the same for the protected and unprotected groups. Considering the simple job hiring example, the education level $E$ is a good choice for the treatment variable.
Two equality of effort notions are defined based on the potential outcome framework, individual $\gamma$-Equal effort and system $\gamma$-Equal effort. Let $Y^{(t)}_i$ be the potential outcome for individual $i$ had $T$ been $t$ and $\ep[Y_i^{(t)}]$ be the expected outcome for individual $i$. Situation testing~\cite{bendick2007situation} is used to estimate the counterfactual potential outcome in a similar way as individual direct discrimination (Section~\ref{sec:idd}). Let $\mathbf{S^+}$ and $\mathbf{S^-}$ be the two sets of similar individuals with $A=a_0$ and $A=a_1$, respectively, and  $\mathbf{E} [Y_{\mathbf{S}^+}^{(t)}]$ be the expected outcome under treatment $t$ for the subgroup of individuals $\mathbf{S^+}$. The minimal effort needed to achieve $\gamma$-level of outcome variable within the subgroup $\mathbf{S^+}$ is defined as:
\begin{equation}
	\quad \Psi _{\mathbf{S}^+} (\gamma) =\argmin{t \in T} \{ \ep[Y_{\mathbf{S}^+}^{(t)}] \geq \gamma \}
\end{equation}
Individual $\gamma$-Equal effort is satisfied for individual $i$ if:
\begin{equation}
\label{eq:effil}
\Psi _{\mathbf{S^+}} (\gamma) = \Psi _{\mathbf{S^-}} (\gamma)
\end{equation}
System $\gamma$-Equal effort is satisfied for a sub-population (e.g. $A=a_1$) if:
 \begin{equation}
\label{eq:effil2}
\Psi _{\mathbf{D^+}} (\gamma) = \Psi _{\mathbf{D^-}} (\gamma)
\end{equation}
where $\mathbf{D}^+$ and $\mathbf{D}^-$ are the subsets of the entire dataset with sensitive attributes $a_0$ and $a_1$, respectively. Both criteria can be used to measure the effort discrepancy between protected and unprotected groups by considering the difference $\Psi_{X^{+}}(\gamma) - \Psi_{X^{-}}(\gamma)$. Unlike most of causal-based fairness notions who intervene ($do$ operator) on the sensitive attribute $A$ ($y_a$, $Y^{a}_i$, etc.), equality of effort intervenes instead on a treatment variable $T$ ($Y^{(t)}_i$). The main limitation of equality of effort notion is that, typically, a single treatment variable does not appropriately reflect the discrepancy between protected and unprotected groups. 

\subsection{Interventional and justifiable fairness}
\label{sec:justif}

Interventional fairness~\cite{salimi2019interventional} is a group-level fairness that can be seen as a strong version of total effect (Eq.~\ref{eq:TE}). Instead of intervening only on the sensitive attribute $A$, interventional fairness intervenes on all remaining variables. Let $\mathbf K$ be a subset of $\mathbf V$ excluding $A$ and $Y$, that is, ${\mathbf K} \subseteq {\mathbf V} - \{A,Y\}$. A predicting algorithm is $\mathbf K$-fair if for any assignment of values ${\mathbf K} = {\mathbf k}$ and outcome $Y=y$:
\begin{equation}
    \label{eq:intervFair}
    \pr(y_{a_1,{\mathbf k}}) = \pr(y_{a_0,{\mathbf k}}) 
\end{equation}

A predicting algorithm is interventionally fair if it is $\mathbf K$-fair for every set of variables $\mathbf K$.
Using the job hiring example of Figure~\ref{fig:hiringExampleGraph1}, interventional fairness holds between male and female groups if $\pr(y_{1,E_u,R_v} = \pr(y_{0,E_u,R_v}), \; \forall u,v\in{0,1}$. Interventional fairness formula (Eq.~\ref{eq:intervFair}) is similar to non-discrimination criterion formula (Eq.~\ref{eq:ndc}). However, while Eq.~\ref{eq:ndc} uses simple conditioning on $A$ and covariates, Eq.~\ref{eq:intervFair} makes an intervention on $A$ and all other covariates and hence works on markovian as well as semi-markovian models.

Justifiable fairness is a relaxation of interventional fairness achieved by classifying the variables as admissible (denoted as $\mathbf E$) or inadmissible (denoted as $\mathbf R$) which correspond, respectively, to explainable and proxy/redlining variables as defined previously.
A predicting algorithm is justifiably fair if it is $\mathbf K$-fair with respect to only supersets of $\mathbf E$, that is, $\mathbf{K} \supseteq \textbf{E}$. Hence, instead of intervening on all variables, it is enough to intervene on only admissible variables (or any superset of them). Graphically, if all directed paths from the sensitive attribute $A$ to the outcome $Y$ go through an admissible attribute in $\mathbf{E}$, then the algorithm is justifiably
fair, which typically coincide with no-unresolved discrimination (Section~\ref{sec:nounresolved}). Using the job hiring example (Figure~\ref{fig:hiringExampleGraph1}), justifiable fairness holds if $\pr(y_{1,E_u}) = \pr(y_{0,E_u}), \; \forall u\in{0,1}$.
Notice that in case $\mathbf{E} = \emptyset$, justifiable fairness coincides with interventional fairness. Interestingly, being based solely on interventions, interventional and justifiable notions of fairness do not require the presence of the underlying causal model. The only assumption is the ability to distinguish admissible and inadmissible variables. 

\subsection{Individual equalized counterfactual odds}
\label{sec:IndECOD}
Individual equalized counterfactual odds~\cite{pfohl2019counterfactual} is a stronger version of counterfactual fairness (Section~\ref{sec:counterfactual}) requiring, in addition, that the factual-counterfactual pair share the same value of the outcome $Y$. The aim is to have a counterfactual version of equalized odds~\cite{hardt2016equality}. This is achieved by conditioning both sides of Eq.~\ref{eq:counterfactual} on the same outcome $Y=y$. 
A predictor satisfies individual equalized counterfactual odds if:
\begin{equation}
    \label{eq:IndECOD}
\pr(\hat{y}_{a_1} \mid \mathbf{X} =  \mathbf{x}, y_{a_1} , A =  a_0) = \pr(\hat{y}_{a_0} \mid \mathbf{X} = \mathbf{x}, y_{a_0} , A =  a_0) 
\end{equation}
The only difference with Eq.~\ref{eq:counterfactual} is the additional conditioning $Y = y_{a_1}$ in the LHS and $Y = y_{a_0}$ in the RHS. The only other causal-based fairness notions considering the outcome $Y$ are counterfactual error rates (Section~\ref{sec:ctf}). However, unlike counterfactual error rates, individual equalized counterfactual odds requires intervention on $Y$. This is the only fairness criterion that requires intervention on the prediction $\hat{Y}$ and on the actual outcome $Y$.

\section{Computing causal quantities from observable data}
\label{sec:computation}
Using causal-based fairness notions is challenging for two reasons. First, among the two possible outcomes, only the factual outcome can be observed. The counterfactual outcome is usually impossible to observe (e.g., if the gender of a candidate is female (factual), it is impossible to observe the counterfactual outcome when the same candidate would have been a male). Second, sensitive attribute (e.g., male and female) is typically not assigned in random in observational data. Hence, the main difficulty to apply causal-based fairness notions is to compute and/or estimate the causal quantities (counterfactual outcomes, causal effects, counterfactual effects, etc.) using observational data. This includes all grayed columns in the simple toy datasets used in Section~\ref{sec:notions} as well as all fairness notions such as $ATE$, $ETT$, counterfactual fairness, etc.
Each causal framework, namely, SCM with causal graphs and potential outcome, uses a different approach to compute/estimate the causal quantities using observational data. The SCM framework relies mainly on the identifiability criterion to generate an expression for the causal quantity based only on observable probabilities. If the identifiability criterion is not satisfied, the causal quantity can not be computed using the available observable data. In such case, as an alternative, if the complete structure of the causal model is available, it is possible to estimate the distribution of the latent variables $U$ and consequently generate an estimation of the counterfactual outcomes. In the potential outcome framework, causal quantities are approximated using several estimation techniques (e.g., matching, re-weighting, etc.). The following subsections illustrate the above three approaches, namely, identifiability, estimation based on full causal model, and potential outcome estimation techniques.

\subsection{Identifiability}
\label{sec:ident}

The identifiability of causal quantities has been extensively studied in the literature: causal effect (intervention) identifiability~\cite{galles1995testing,tian02,tian03,tian04,shpitser2006identification,huang06,shpitser08,pearl2009causality}, counterfactual identifiability~\cite{shpitser07-counterfactuals,shpitser08,shpitser13,wu19}, direct/indirect effects~\cite{pearl01direct} and path-specific effect identifiability~\cite{avin2005identifiability,shpitser13,zhang2017causal,zhang2017anti,malinsky19}.
This section summarizes the main identifiability conditions as they relate to the specific problem of discrimination discovery.
\subsubsection{Identifiability of causal effect (intervention)}
\label{identInterv}
The causal effect of a cause variable $X$ on an effect variable $Y$ is computed using $\pr(Y_x) = \pr (Y|do(X=x))$, the distribution of $Y$ after the intervention $X=x$. In discrimination setup, the cause is typically the sensitive attribute $A$. 
A basic case where identifiability can be avoided altogether is when it is possible to perform experiments by intervening on the sensitive attribute $A$. When this is possible, randomized controlled trial (RCT)~\cite{fisher92} can be used to estimate the causal effect. RCT consists in randomly assigning subjects (e.g., individuals) to treatments (e.g., gender), then comparing the outcome $Y$ of all treatment groups. 
However, in the context of machine learning fairness, RCT is often not an option as experiments can be too costly to implement, physically impossible, or ethically not acceptable (e.g., changing the gender of a job applicant).

\begin{figure}[!h]
    \subfigure [Simple Markovian model with a collider $(W)$ and a mediator $(Z)$.]  {%
    {\includegraphics [scale=0.2]{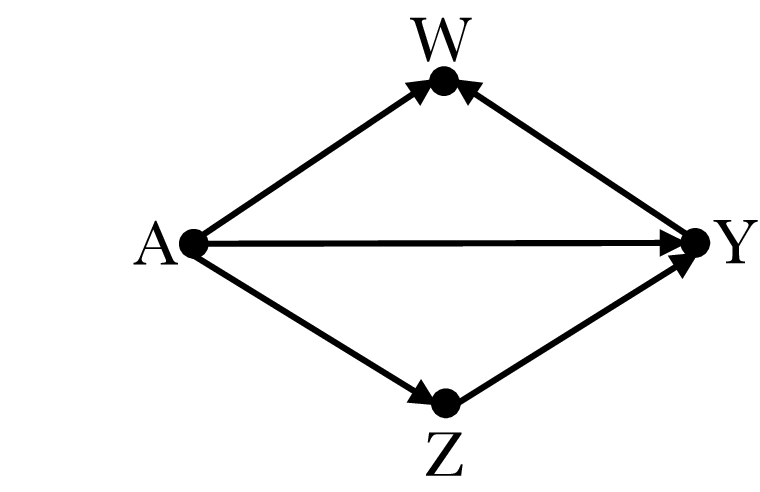} }
    \label{subfig:fig33_a}}
    \quad 
    \subfigure [Simple Markovian model with a confounder $(C)$.] {%
    {\includegraphics[scale=0.2]{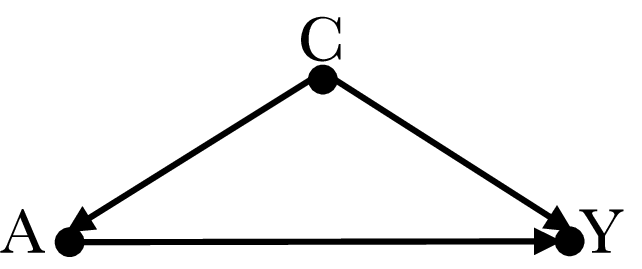} }
    \label{subfig:fig33_b}}
     \quad 
    \subfigure [Simple Markovian model with a confounder $(C)$ and a mediator $(Z)$.]  {%
    {\includegraphics [scale=0.2]{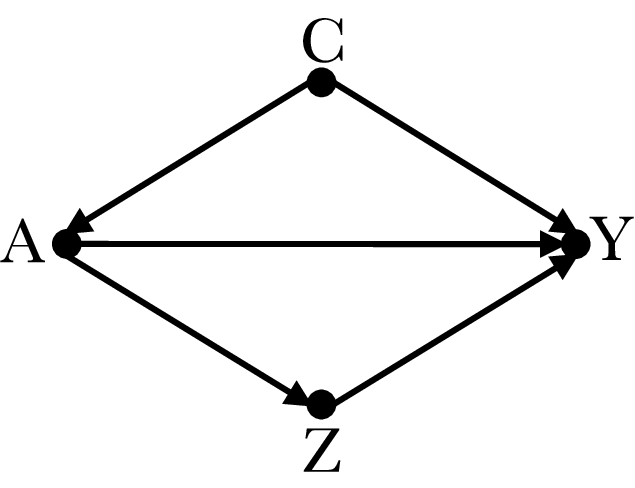} }
    \label{subfig:fig33_c}}
    \caption{Simple causal graphs}
    \label{fig:fig33}
\end{figure}

In Markovian models (no unobserved confounding), the causal effect is always identifiable (Corollary 3.2.6 in~\cite{pearl2009causality}). The simplest case is when there is no confounding between $A$ and $Y$ (Figure~\ref{subfig:fig33_a}). In that case, the causal effect matches the conditional probability regardless of any mediator:
\begin{equation}
	\pr(y_a) = \pr(y|do(a)) = \pr(y|a) \label{eq:int1}
\end{equation}

In presence of an observable confounder (Figure~\ref{subfig:fig33_b}), $\pr(y_a)$ is identifiable by adjusting on the confounder:
\begin{equation}
	\pr(y_a) = \sum_{C}\pr(y|a,c)\;\pr(c)  \label{eq:int2}
\end{equation}
where the summation is on values $c$ in the domain (sample space) of $C$ denoted as $dom(C)$. Eq.~\ref{eq:int2} is called the back-door formula\footnote{Called also adjustment formula or stratification.}. 
The backdoor adjusting formula is different from the joint probability \[ \pr(y,a,c) = \pr(y|a,c)\;\pr(a|c)\;\pr(c)\] and the conditional probability \[ \pr(y|a) = \sum_{C}\pr(y|a,c)\;\pr(c|a)\].

For semi-Markovian models, identifiability of $\pr(y_a)$ is not guaranteed. In case it is identifiable, Pearl~\cite{pearl2009causality} proposes a $do$-calculus composed of three rules allowing to express interventional probabilities in terms of observational ones:
\begin{enumerate}
	\item $\pr(y_a|\mathbf{z},w) = \pr(y_a|\mathbf{z})$ provided that the set of variables $\mathbf{Z}$ blocks all backdoor paths from $W$ to $Y$ after all arrows leading to $A$ have been deleted.
	\item $\pr(y_a|\mathbf{z}) = \pr(y|a,\mathbf{z})$ provided that the set of variables $\mathbf{Z}$ blocks all backdoor paths from $A$ to $Y$. 
    \item $\pr(y_a) = \pr(y)$ provided that there are no causal paths between $A$ and $Y$.
\end{enumerate}

\begin{figure}[!h]
    \subfigure []  {%
    {\includegraphics [scale=0.22]{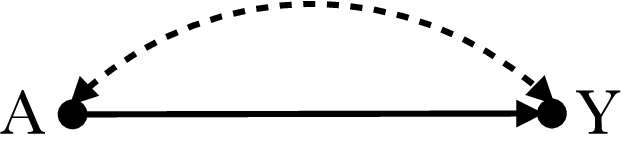} }
    \label{subfig:ident_a}}
    \qquad 
    \subfigure [] {%
    {\includegraphics[scale=0.25]{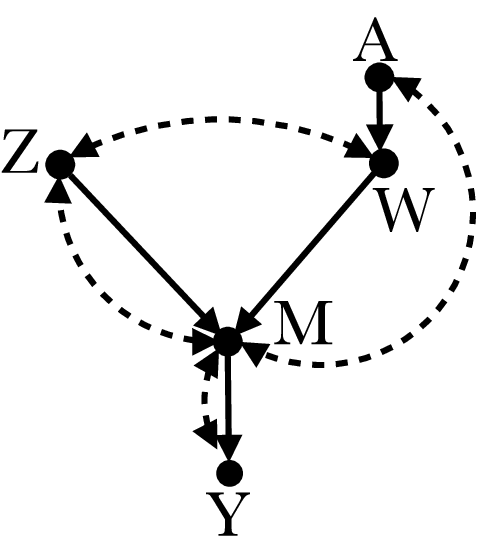} }
    \label{subfig:ident_b}}
     \qquad 
    \subfigure []  {%
    {\includegraphics [scale=0.25]{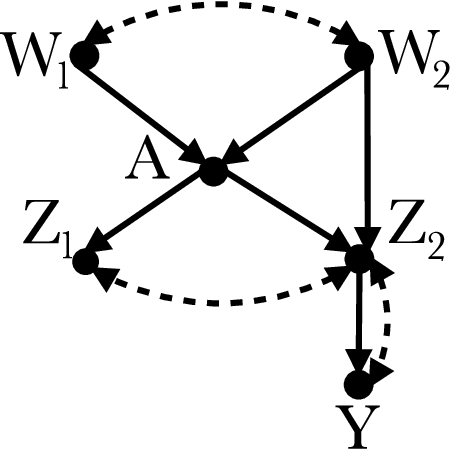} }
    \label{subfig:ident_c}}
    \subfigure []  {%
    {\includegraphics [scale=0.25]{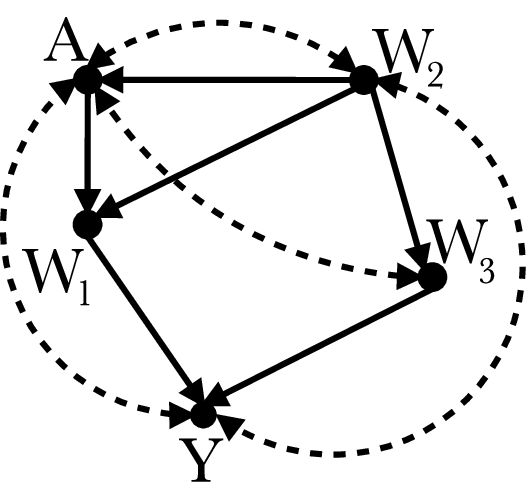} }
    \label{subfig:ident_d}}
    \qquad \quad
    \subfigure []  {%
    {\includegraphics [scale=0.25]{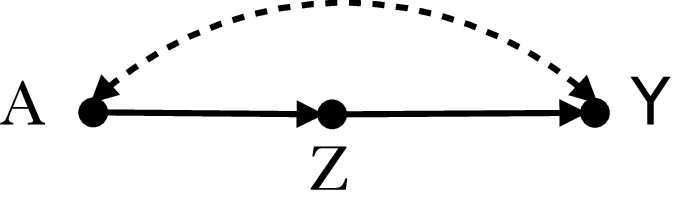} }
    \label{subfig:ident_e}}
    \vspace{-2mm}
    \caption{Figure~\ref{subfig:ident_a} presents the``bow" graph, Figure~\ref{subfig:ident_b} illustrates the structure of a c-tree, Figure~\ref{subfig:ident_c} shows a semi-Markovian model where $\pr P(y_a)$ is observable, Figure~\ref{subfig:ident_d} presents a semi-Markovian model where $\pr P(y_a)$ is identifiable and Figure~\ref{subfig:ident_e} illustrates a simple example of the front-door criterion.}
    \label{fig:ident}
\end{figure}

$do$-calculus has been proven to be sound and complete in the identification of interventional distributions~\cite{huang06}. 
For example, $\pr(y_a)$ is identifiable in Figure~\ref{subfig:ident_d}. By applying the chain rule following the topological order: $W_2<A<W_1<W_3<Y$, we get:
\begin{align}
     \label{eq:do1}
     \pr(y_a)  & = \sum_{w_1 w_2 w_3} \pr(y|do(a),w_1,w_2,w_3)\;\pr(w_1|do(a),w_2)\;\pr(w_2) \nonumber\\
             & \qquad \qquad \times \pr(w_3|w_2,w_1,do(a)) \\
     \label{eq:do2}
             & = \sum_{w_1 w_2} \pr(y|do(a),w_1,w_2)\;\pr(w_1|do(a),w_2)\;\pr(w_2)\\
     \label{eq:do3}
             & = \sum_{w_1 w_2} \pr(y|do(a),w_2)\;\pr(w_1|a,w_2)\;\pr(w_2)\\
     \label{eq:do4}
             & = \sum_{w_1 w_2} \sum_{a'} \pr(y|a',w_2,do(w_1))\;\pr(a'|do(w_1),w_2) \nonumber\\
             & \qquad \qquad \times \pr(w_1|a,w_2)\;\pr(w_2)\\
     \label{eq:do5}
             &= \sum_{w_1'}\; \sum_{w_2'}\; \sum_{a'}\; \pr(y|w_1',w_2',a')\;\pr(a'|w_2') \; \pr(w_1'|w_2',a)\;\pr(w_2')
\end{align}
Note that $w_3$ is omitted from (\ref{eq:do2}) since it is considered latent~\cite{tikka2017enhancing}. Applying Rule 2 followed by Rule 3 to the first term in~(\ref{eq:do2}) yields to $\pr(y|do(a),w_2)$ (\ref{eq:do3}). Likewise, applying Rule 2 to the second term in ~(\ref{eq:do2}) leads to $\pr P(w_1|a,w_2)$. Thus, the original problem reduces to identifying the term $\pr(y|do(a),w_2)$ in~(\ref{eq:do3}). Here we cannot apply Rule 2 to exchange $do(a)$ with $a$ because $G_{\underline{A}}$ (graph obtained by removing all emanating arrows from $A$) contains a backdoor path from $A$ to $Y$. Thus, to block that path, we need to condition and to sum over all values of $A$ as shown in Eq. (\ref{eq:do4}) ($\sum_{a'}\;\pr(y|a',w_2,do(w_1))\\ \pr(a'|do(w_1),w_2$)). Now, applying Rule 2 to $\pr(y|a',w_2,do(w_1))$ and Rule 3 to $\pr(a'|do(w_1),w_2)$ and adding the other terms results in the final expression in (\ref{eq:do5}). The problem of $do$-calculus is the difficulty to determine the correct order of application of the rules. Using the wrong order may hinder the identifiability or produce a very complex expression~\cite{tikka2017simplifying}. As an alternative to using the do-calculus, several contributions in the identifiability literature focused on defining graphical patterns and mapping them to simple and concise intervention-free expressions~\cite{tian02,tian03,tian04,shpitser2006identification}.

All graphical criteria can be generalized to the case where the sensitive attribute is not connected to any of its children through a confounding path. In such case, c-component factorization can be used. A c-component is a set of vertices in the graph such that every pair of vertices are connected by a confounding edge. The idea of c-component factorization is to decompose the identification problem into smaller sub-problems, that is, a disjoint set of c-components in order to calculate $\pr(y_a)$. For example, in Figure~\ref{subfig:ident_c}, there are three c-components: $\{\{W1,W2\},\{A\}, \{Z1,Z2,Y\}\}$. Hence, as long as there is no confounding path connecting $A$ to any of its direct children, $\pr(y_a)$ is identifiable. C-component factorization is used in the ID algorithm~\cite{shpitser08} which is proven to be complete for causal effect identification. 

In case there is an unobservable confounding between the sensitive attribute $A$ and the outcome $Y$, all the above criteria will fail. However, $\pr(y_a)$ can still be identifiable using the front-door criterion. This criterion is satisfied in Figure~\ref{subfig:ident_e} and consists in having a mediator variable $Z$ such that:
\begin{itemize}
\item there are no backdoor paths from $A$ to $Z$
\item all backdoor paths from $Z$ to $Y$ are blocked by $A$.
\end{itemize}
A backdoor path from $A$ to $Z$ is any path starting at $A$ with a backward edge $\leftarrow$ into $A$ (e.g., $A \leftarrow \ldots Z$). If such criterion is satisfied, $\pr(y_a)$ can be computed as follows:
\begin{align}
	\pr(y_a)  & = \sum_{Z} \pr(y|do(z))\;\pr(z|do(a)) \nonumber \\
		& = \sum_{Z} \pr(y|z,a)\;\pr(a)\;\pr(z|a) \label{eq:int6}
\end{align}

Shpitser and Pearl proved that all the unidentifiable cases of the causal effect $\pr(y_a)$ boil down to a general graphical structure called the hedge criterion. Based on this criterion, they designed a complete identifiability algorithm called $ID$ which outputs the expression of $\pr(y_a)$ if it is identifiable, or the reason of the unidentifiability, otherwise.

The simplest graph in which the causal effect between $A$ and $Y$ is not identifiable is the``bow" graph (Figure~\ref{subfig:ident_a}). This simple unidentifiability criterion can be generalized to a more complex graphs called c-tree. A c-tree is a graph that is at the same time a tree\footnote{Notice that, in this paper, the direction of the arrows between nodes is reversed compared to the usual tree structure.} and a c-component. Figure~\ref{subfig:ident_b} shows an example of a c-tree. If the causal graph is a c-tree rooted in the outcome variable $Y$, $\pr(y_a)$ is unidentifiable~\cite{shpitser08}.

\subsubsection{Identifiability of counterfactuals}
\label{sec:ident_ctf}
Most of causal-based fairness notions in the disparate treatment framework ($NDE$~(Eq.~\ref{eq:NDE}), path-specific effect~(Eq.~\ref{eq:PSE}), counterfactual effects~(Section~\ref{sec:ctf}), etc.) are defined in terms of counterfactual quantities. Hence, the applicability of those notions depends heavily on the identifiability of the counterfactuals composing them. 
In Markovian, as well as semi-Markovian models, if all parameters of the causal model are known (including $\pr(\mathbf{u})$), any counterfactual is identifiable and can be computed using the three steps abduction, action, and prediction (Theorem 7.1.7 in~\cite{pearl2009causality}).   

Let $P_{*}= \{P_{\mathbf{x}}|\mathbf{X} \subseteq \mathbf{V},\;\mathbf{x}\;\text{\em a value assignment of }\; \mathbf{X}\}$ be the set of all interventional distributions in a given causal model. While the identifiability of interventional proabilities $\pr(y_a)$ is characterized based on observational probabilities $\pr(\mathbf{v})$, in this section, the identifiability of counterfactuals is characterized in terms of interventional probabilities $P_{*}$. Then, combining these results with the criteria of the previous section, a counterfactual can, in turn, be identified using observational probabilities $\pr(\mathbf{v})$. 

\begin{figure}[!h]
    \subfigure [] {%
    {\includegraphics[scale=0.25]{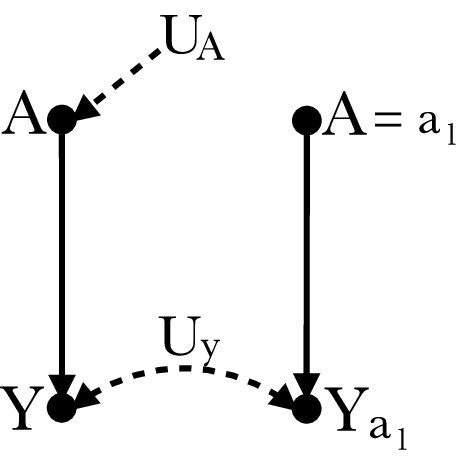} }
    \label{subfig:ident_ctf_b}}
     \qquad 
    \subfigure []  {%
    {\includegraphics [scale=0.25]{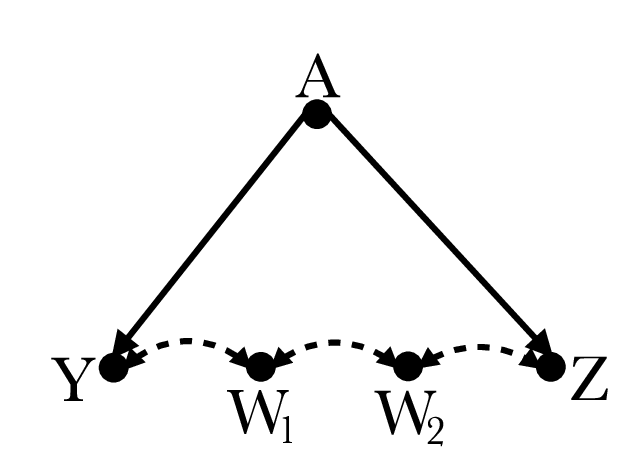} }
    \label{subfig:ident_ctf_c}}
        \\
    \subfigure []  {%
    {\includegraphics [scale=0.25]{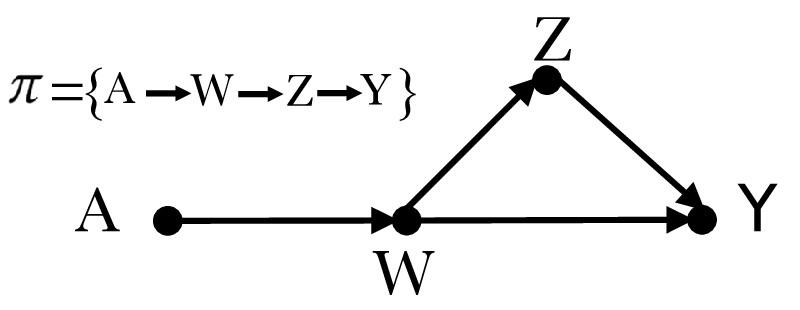} }
    \label{subfig:ident_ctf_d}}
        \qquad 
    \subfigure []  {%
    {\includegraphics [scale=0.22]{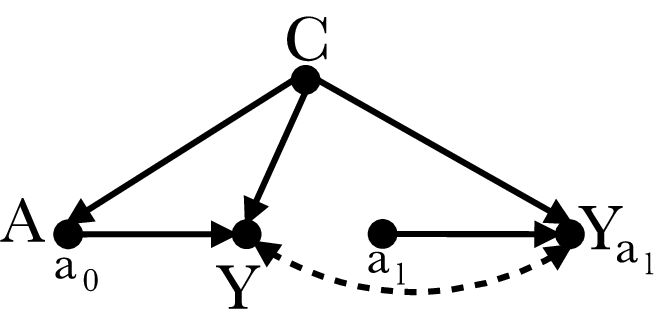} }
    \label{subfig:ident_ctf_e}}
    \caption{Causal graphs.}
    \label{fig:ident_ctf}
    \vspace{-3mm}
\end{figure}

Given a causal graph $G$ of a Markovian model and a counterfactual expression $\gamma = v_{x}|e$ with $e$ some arbitrary set of evidence, identifying and computing $\pr(\gamma)$ requires to construct a counterfactual graph which combines parallel worlds. Every world is represented by a model $M_{x}$ corresponding to each subscript in the counterfactual expression. For example, given the causal graph in Figure~\ref{fig:firing_example} and the counterfactual expression $y_{a_1}|a_0$, the resulting counterfactual graph is shown in Figure~\ref{subfig:ident_ctf_e}. 
The counterfactual graph should be ``reduced'' by merging together vertices that share the same causal mechanism (\textbf{make-cg} algorithm in~\cite{shpitser08} automates this procedure). The resulting counterfactual graph can be considered as a typical causal graph for a larger causal model. Consequently, all the graphical criteria listed in Section~\ref{identInterv} for the identifiability of causal effects apply on the counterfactual graph to identify counterfactual quantities, in particular, the c-component factorization of the counterfactual graph~\cite{shpitser07-counterfactuals}. $ID^*$ and $IDC^*$ algorithms~\cite{shpitser08} automate the identifiability and computation of counterfactuals based on all the above criteria. Note that $ID^*$ and $IDC^*$ output expressions in terms of interventional probabilities $P_*$. Then, $ID$ algorithm is used to express those interventional probabilities in terms of observational probabilities. 

The simplest unidentifiable counterfactual quantity is $\pr(y_{a'},y'_a)$ which is called the probability of necessity and sufficiency. The corresponding counterfactual graph is the W-graph that has the same structure as to Figure~\ref{subfig:ident_ctf_b}. This simple criterion can be generalized to the zig-zag graph (Figure~\ref{subfig:ident_ctf_c}) where the counterfactual $\pr(y_a,w_1,w_2,z')$ is not identifiable.

Pearl~\cite{pearl2009causality} proves two results about the identifiability of counterfactuals. First, for linear causal models (i.e., the functions $\mathbf{F}$ are linear), any counterfactual is experimentally (using $P_{*}$) identifiable whenever the model parameters are identified. Second, in linear causal models, if some of the model parameters are unknown, any counterfactual of the form  $\ep(Y_a|e)$ where $e$ is some arbitrary set of evidence, is identifiable provided that  $\ep(y_a)$ is identifiable. Finally, there is no single necessary and sufficient criterion for the identifiability of counterfactuals in semi-Markovian models~\cite{avin2005identifiability}. 

To illustrate the computation of a counterfactual probability, consider the teacher firing example of Figure~\ref{fig:firing_example} and the counterfactual probability $\pr(y_{a_1}|a_0)$ which reads the probability of firing a teacher who is assigned a class with a high initial level of students ($a_0$) had she been assigned a class with a low initial level of students ($a_1$). Applying \textbf{make-cg} algorithm based on this counterfactual quantity produces the counterfactual graph in Figure~\ref{subfig:ident_ctf_e} which combines two worlds: the actual world where the teacher has normally $A=a_0$ and the counterfactual world where \textit{the same} teacher is assigned $A=a_1$. Both variables $C$ are reduced to a single variable and $Y$ and $Y_{a_1}$ are connected by an unobservable confounder. The counterfactual graph is composed of three c-components $\{C\},\{A\},\{Y,Y_{a_1}\}$. 
Applying algorithm $IDC^*$~\cite{shpitser08} results in:

\begin{align}
	\pr(y_{a_1}|a_0) & = \frac{\sum_{y,c}\;Q(c)\;Q(a_0)\;Q(y,y_{a_1})}{\pr(a_0)}
\end{align}
where $Q(\mathbf{v}) = \pr(\mathbf{v}|pa(\mathbf{V}))$ in the counterfactual graph.
Hence,
\begin{align}
	\pr(y_{a_1}|a_0) & = \frac{\sum_{y,c}\;\pr(c)\;\pr(a_0|c)\;\pr(y,y_{a_1}|c)}{\pr(a_0)} \nonumber\\
	& = \frac{\sum_{c}\;\pr(c)\;\pr(a_0|c)\;\pr(y_{a_1}|c)}{\pr(a_0)} \label{eq:cs2} \\
	& = \frac{\sum_{c}\;\pr(c)\;\pr(a_0|c)\;\pr(y|a_1,c)}{\pr(a_0)}\label{eq:cs3} \\
			& = \frac{0.5\times0.8\times0.25 + 0.5\times0.2\times0.01}{0.5} \nonumber \\
			& = 0.202 \nonumber
\end{align}

$y$ in Eq.~\ref{eq:cs2} is cancelled by summation while $\pr(y_{a_1}|c)$ in the same equation is transformed into $\pr(y|a_1,c)$ in Eq.~\ref{eq:cs3} using Rule 2 of the $do$-calculus.

\subsubsection{Identifiability of direct and indirect effects}
In Markovian models, the average natural direct effect $NDE$ and the average natural indirect effect $NIE$ are always identifiable (from observational data) and can be computed as follows~\cite{pearl01direct}:
\begin{equation}
	NDE_{a_1,a_0}(Y) = \sum_{\mathbf{s}} \sum_{\mathbf{z}} \bigg(\ep[Y|a_1,\mathbf{z}]-\ep[Y|a_0,\mathbf{z}] \bigg)\pr(\mathbf{z}|a_0,\mathbf{s})\pr(\mathbf{s}) \label{eq:nde}
\end{equation}
\begin{equation}
NIE_{a_1,a_0}(Y) = \sum_{\mathbf{s}} \sum_{\mathbf{z}} \ep[Y|a_0,\mathbf{z}]\;\bigg (\pr(\mathbf{z}|a_1,\mathbf{s})-\pr(\mathbf{z}|a_0,\mathbf{s})\bigg )\;\pr(\mathbf{s}) \label{eq:nie}
\end{equation}
where $\mathbf{Z}$ is a set of mediator variables and $\mathbf{S}$ is any set of variables satisfying the back-door criterion between the sensitive variable $A$ and the mediator variables $\mathbf{Z}$, that is, \textit{(i)} no variable in $\mathbf{S}$ is a descendant of $A$ and \textit{(ii)} $\mathbf{S}$ blocks all back-door paths between $A$ and $\mathbf{Z}$.
A simpler formulation can be used in case there is no confounding between $A$ and $\mathbf{Z}$, where the need for $\mathbf{S}$ is dropped altogether:
\begin{equation}
	NDE_{a_1,a_0}(Y) = \sum_{\mathbf{z}} \bigg(\ep[Y|a_1,\mathbf{z}]-\ep[Y|a_0,\mathbf{z}] \bigg)\;\pr(\mathbf{z}|a_0) \label{eq:nde2}
\end{equation}
\begin{equation}
NIE_{a_1,a_0}(Y) = \sum_{\mathbf{z}} \ep[Y|a_0,\mathbf{z}]\;\bigg (\pr(\mathbf{z}|a_1)-\pr(\mathbf{z}|a_0)\bigg ) \label{eq:nie2}
\end{equation}

In semi-Markovian models, $NDE$ and $NIE$ are not generally identifiable, even if we have the luxury to perform any experiment using $RCT$, because of the nested counterfactuals $\pr(Y_{a_1},\mathbf{Z}_{a_0})$ and  $\pr(Y_{a_0},\mathbf{Z}_{a_1})$ in Eq.~\ref{eq:NDE} and Eq.~\ref{eq:NIE}, respectively.  Nevertheless, these quantities are identifiable \textit{from experimental data} provided that there is a set of variables $\mathbf{W}$ which are parents of the outcome variable $Y$ but non-descendants of $A$ and $Z$ such that $Y_{a_0,z} \independent Z_{a_0} | \mathbf{W}$ (reads: $Y_{a_0,z}$ and $Z_{a_0}$ are independent conditional on $\mathbf{W}$). This condition can be easily checked from the causal graph as follows: $\mathbf{W}$ d-separates $Y$ and $\mathbf{Z}$ in the graph formed by deleting all arrows emanating from $A$ and $\mathbf{Z}$, denoted simply as $(Y \independent \mathbf{Z} | \mathbf{W})_{G_{\underline{AZ}}}$.

If such graphical condition is satisfied, $NDE$ and $NIE$ can be computed from experimental quantities as follows:
\begin{equation}
NDE_{a_1,a_0}(Y) = \sum_{\mathbf{z,w}}\bigg (\ep[Y_{a_1,\mathbf{z}}|\mathbf{w}] - \ep[Y_{a_0,\mathbf{z}}|\mathbf{w}]\bigg )\;\pr(\mathbf{Z}_{a_0}=\mathbf{z}|\mathbf{w})\;\pr(\mathbf{w}) \label{eq:nde3} 
	\end{equation}
\begin{equation}
NIE_{a_1,a_0}(Y) = \sum_{\mathbf{z,w}} \ep[Y_{a_0,\mathbf{z}}|\mathbf{w}]\bigg (\pr(\mathbf{Z}_{a_1}=\mathbf{z}|\mathbf{w})-\pr(\mathbf{Z}_{a_0}=\mathbf{z}|\mathbf{w}) \bigg )\pr(\mathbf{w}) \label{eq:nie3} 
	\end{equation}
\subsubsection{Identifiability of path-specific effects}
\label{identPS}

The identifiability of $PSE_{\pi}(a_1,a_0)$ in Markovian models depends on whether\\ $\pr(y|do(a_1|_{\pi},a_0|_{\bar{\pi}}))$ is identifiable. Avin et al.~\cite{avin2005identifiability} gave a single necessary and sufficient criterion for the identifiability of $\pr(y|do(a_1|_{\pi},a_0|_{\bar{\pi}}))$ in Markovian models called recanting witness criterion. This criterion holds when there is a vertex $W$ along the causal path $\pi$ that is connected to $Y$ through another causal path not in $\pi$. For instance, Figure~\ref{subfig:ident_ctf_d} satisfies the recanting witness criterion when $\pi = A\rightarrow W \rightarrow Z \rightarrow Y$ with $W$ as witness. The corresponding graph structure is called ``kite'' graph. When this criterion is satisfied,  $\pr(y|do(a_1|_{\pi},a_0|_{\bar{\pi}}))$ is not identifiable, and consequently,  $PSE_{\pi}(a_1,a_0)$ is not identifiable. 
Shpitser~\cite{shpitser13} generalizes this criterion to semi-Markovian models known as recanting district criterion.

\subsection{Estimation based on full knowledge of the causal model parameters}
\label{subsec:ctf}
The main reason behind the unidentifiability of causal quantities (causal effect, counterfactuals, etc.) is the presence of unobservable variables, namely, hidden latent variables.
Some causal-based fairness notions, such as counterfactual fairness~\cite{kusner2017counterfactual}, can be assessed in presence of such unobservable latent variables. The only requirement, however, is the knowledge of the causal model structure (skeleton). Based on the causal model, the latent/background variables are estimated using observable data. Then, the predictor is trained using both observable (non-descendants of the sensitive attributes) as well as the estimated latent variables. Such predictor tends to be more fair than typical predictors (trained using only observable variables) since it takes into consideration hidden bias captured by latent variables. Given the full causal model, counterfactual fairness can be assessed by generating, for every observable data sample, a counterfactual data sample by simply changing the sensitive attribute value (e.g., turn male into female) then using the three-steps process (abduction, action, prediction) to compute the outcome. The predictor is considered fair if the predicted outcomes distributions of both groups (protected and unprotected) are similar. 

In this survey, and for clarity of presentation, the counterfactually fair learning and estimation approach described by Kusner et al.\cite{kusner2017counterfactual} is illustrated in Algorithm~\ref{alg:one}. 
\RestyleAlgo{ruled}
\SetKwComment{Comment}{/* }{ */}
\begin{algorithm}
\caption{\small Counterfactual learning and assessment}\label{alg:one}
\SetKwInOut{Input}{input}\SetKwInOut{Output}{output}
\Input{\small Labelled dataset $\mathcal{D}\equiv\{(A^{(i)},X^{(i)},Y^{(i)})\}$ $i=1\ldots n$, \\Causal model $\mathcal M$ with causal graph $\mathcal G$} 
\Output{\small Predictor $\hat{Y}$ w/ counterfactual fairness constraints, \\Estimation of counterfactual bias}
Fit the causal model $\mathcal M$ based on labelled dataset $\mathcal D$\label{alg1-line1}\;
Estimate the posterior distribution of the latent variable(s) $U$: $\pr_{\mathcal M}(U|X,A)$\label{alg1-line2}\;
For every sample $(x^{(i)},a^{(i)})$ of $\mathcal D$, generate a counterfactual sample for every possible combination of sensitive attribute values $a^{\prime(i)}$\label{alg1-line3}\;
For every counterfactual sample generated in Step~\ref{alg1-line3}, use the three inference steps: \textbf{Abduction}, \textbf{Action}, and \textbf{Prediction} to compute the values of the remaining variables $x^{\prime(i)}$ as well as the label $y^{(i')}$ \label{alg1-line4}\;
Train a predictor $\hat{Y}$ using only variables non-descendants of $A$ and the estimated latent variables $U$, that is, $\hat{Y} \equiv h_{\theta}(U,X_{\nsucc A})$ where $\theta$ represents the predictor parameters and $X_{\nsucc A} \subseteq X$ are non-descendants of $A$\label{alg1-line5}\;
Use the trained predictor $\hat{Y}$ to compute $\hat{y}^{(i)}$ for every observed as well as generated counterfactual sample in the test set\label{alg1-line6}\;
Using predicted outcomes $\hat{y}^{(i)}$ of observed and counterfactual samples, estimate the counterfactual bias\label{alg1-line7}\;
\end{algorithm}
The approach takes as input the labelled dataset $\mathcal D$ and assumes the structure of the causal model $\mathcal M$ is available including the relationships between all variables (observed and latent) along with the causal graph. For example, Kusner et al.\cite{kusner2017counterfactual} assumed a single latent variable $U$ and a combination of Normal and Poisson distributions for their illustrated example. If the parameter values are not known, they can be estimated using the observed data (Step~\ref{alg1-line1}). 
The approach has two objectives $(1)$ learning a predictor $\hat{Y}$ while taking into consideration counterfactual fairness constraints and $(2)$ assessing whether the prediction is counterfactually fair. Given the causal model $\mathcal M$, Step~\ref{alg1-line1} fits the model parameters to the observed data $\mathcal D$. This allows to estimate the posterior distribution of the latent variable(s) $U$ (Step~\ref{alg1-line2}). Step~\ref{alg1-line3} is straightforward and consists in generating a counterfactual sample for every observed sample by assigning a different value to the sensitive attribute. For example, if the observed sample is $(X=x_1\;,\;A=male)$, the counterfactual sample would be $(X=x_1\;,\;A=female)$. Step~\ref{alg1-line4} is the hard part of generating the counterfactual sample, which is, computing the ``would-be'' values of the remaining variables of these ``made-up'' samples. It tries to answer the core question of causal approaches: what would the output be, had the sensitive attribute value was different? This is achieved by the three steps process (Theorem~7.1.7 in~\cite{pearl2009causality}) to compute counterfactuals: 
\begin{enumerate}
    \item \textbf{Abduction}: compute the posterior distribution of $U$ given the evidence, that is, the observed data $(X=x^{(i)},A=a^{(i)})$
    \item \textbf{Action}: Substitute the observed sensitive attribute equation with the counterfactual value (e.g., $A=female$) in the causal model $\mathcal M$
    \item \textbf{Prediction}: Compute the remaining variables values (including $Y$) using the new equations.
\end{enumerate}
Enforcing counterfactual fairness while training the predictor (Step~\ref{alg1-line5}) is achieved through the use of two principles. First, by using only variables non-descendants of the sensitive attribute $A$ which is direct implication of counterfactual fairness definition (Lemma~1 in~\cite{kusner2017counterfactual}). Second, unlike typical predictors, by considering latent variables $U$ in the training. This allows to involve the background sources of the social bias encoded in the latent variables. The predictor is then used to generate predictions for observed and counterfactual samples in the test set\footnote{$80\%$ of the data is used for training while $20\%$ of the data is used for testing.}(Step~\ref{alg1-line6}). Finally, counterfactual bias can be estimated by comparing the output of observed samples and counterfactual samples (Step~\ref{alg1-line7}). In Kusner et al.~\cite{kusner2017counterfactual} and in this survey, density distributions are used to estimate the counterfactual bias.   

Algorithm~\ref{alg:one} illustrates the steps for the (counterfactually) fair learning of a predictor and the estimation of counterfactual bias. However, typical real-world scenarios come with an already trained predictor and hence need only the second goal, that is, counterfactual fairness assessment of that predictor. Algorithm~\ref{alg:two} is a modified version of Algorithm~\ref{alg:one} which, given an already trained predictor, tries to tell how counterfactually-fair the predictor is.

\begin{algorithm}
\caption{\small Counterfactual fairness assessment}\label{alg:two}
\SetKwInOut{Input}{input}\SetKwInOut{Output}{output}
\Input{\small Labelled dataset $\mathcal{D}\equiv\{(A^{(i)},X^{(i)},Y^{(i)})\}$ $i=1\ldots n$, \\
Predictor $\hat{Y}$,\\Causal model $\mathcal M$ with causal graph $\mathcal G$} 
\Output{\small Estimation of counterfactual bias in $\hat{Y}$}
Fit the causal model $\mathcal M$ based on labelled dataset $\mathcal D$\label{alg2-line1}\;
Estimate the posterior distribution of the latent variable(s) $U$: $\pr_{\mathcal M}(U|X,A)$\label{alg2-line2}\;
For every sample $(x^{(i)},a^{(i)})$ of $\mathcal D$, generate a counterfactual sample for every possible combination of sensitive attribute values $a^{\prime(i)}$\label{alg2-line3}\;
For every counterfactual sample generated in Step~\ref{alg2-line3}, use the three inference steps: \textbf{Abduction}, \textbf{Action}, and \textbf{Prediction} to compute the values of the remaining variables $x^{\prime(i)}$ as well as the label $y^{(i')}$ labels \label{alg2-line4}\;
Use predictor $\hat{Y}$ to compute $\hat{y}^{(i')}$ for every generated counterfactual sample in the test set\label{alg2-line5}\;
Using observed and counterfactual samples, estimate the counterfactual bias\label{alg2-line6}\;
\end{algorithm}

It is important to mention that assessing counterfactual bias of a given predictor requires to have access to that predictor. This is needed to generate predictions for the counterfactual samples (Step~\ref{alg2-line5}).

\begin{table*}[th]
  \centering
  \begin{tabular}{lll}
  \hline
    \multicolumn{3}{c}{Law school}\\
    \hline
    \begin{minipage}{.2\textwidth}
      \includegraphics[width=\textwidth,height=30mm]{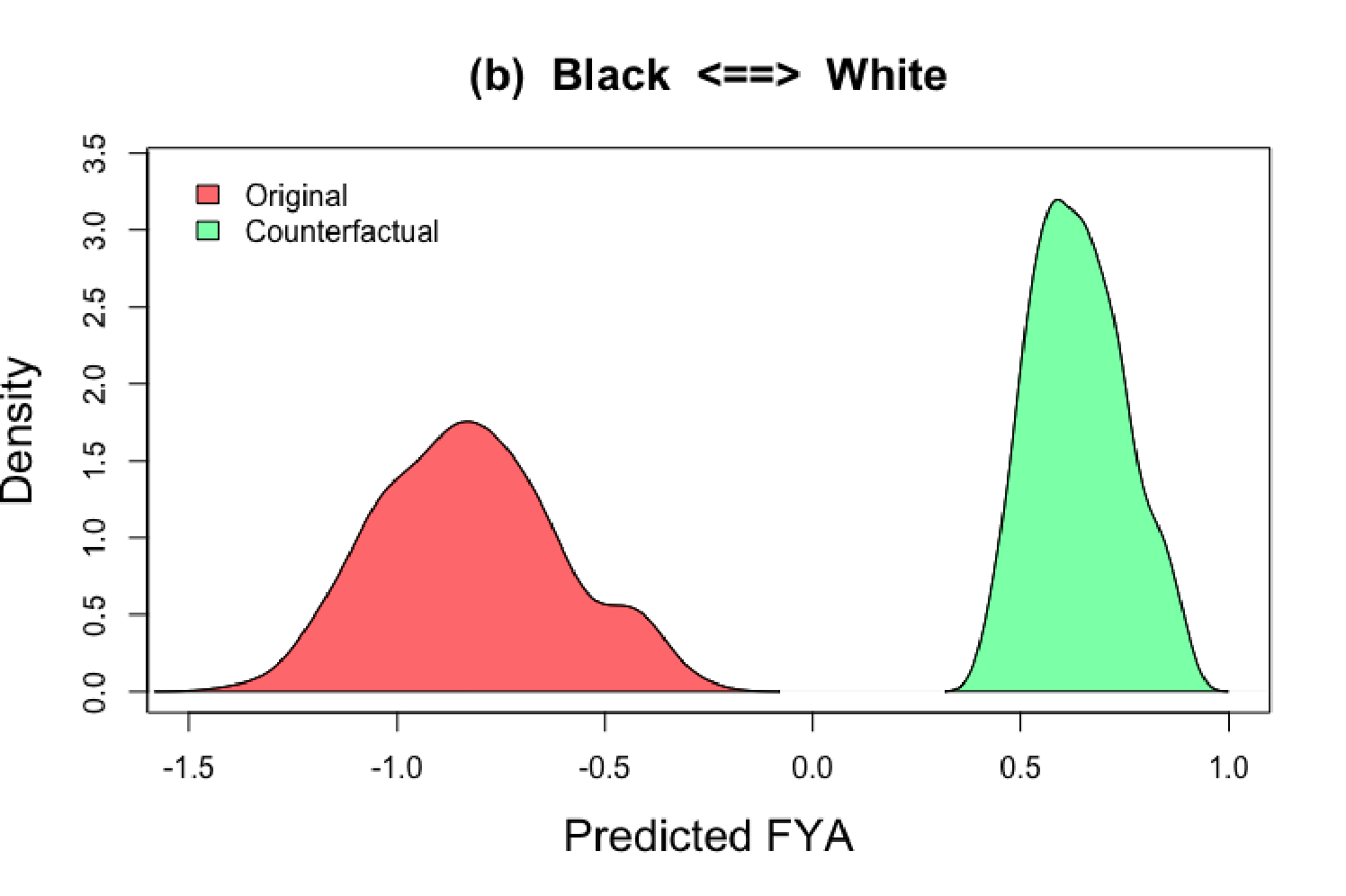}
    \end{minipage}& \begin{minipage}{.2\textwidth}
      \includegraphics[width=\textwidth, height=30mm]{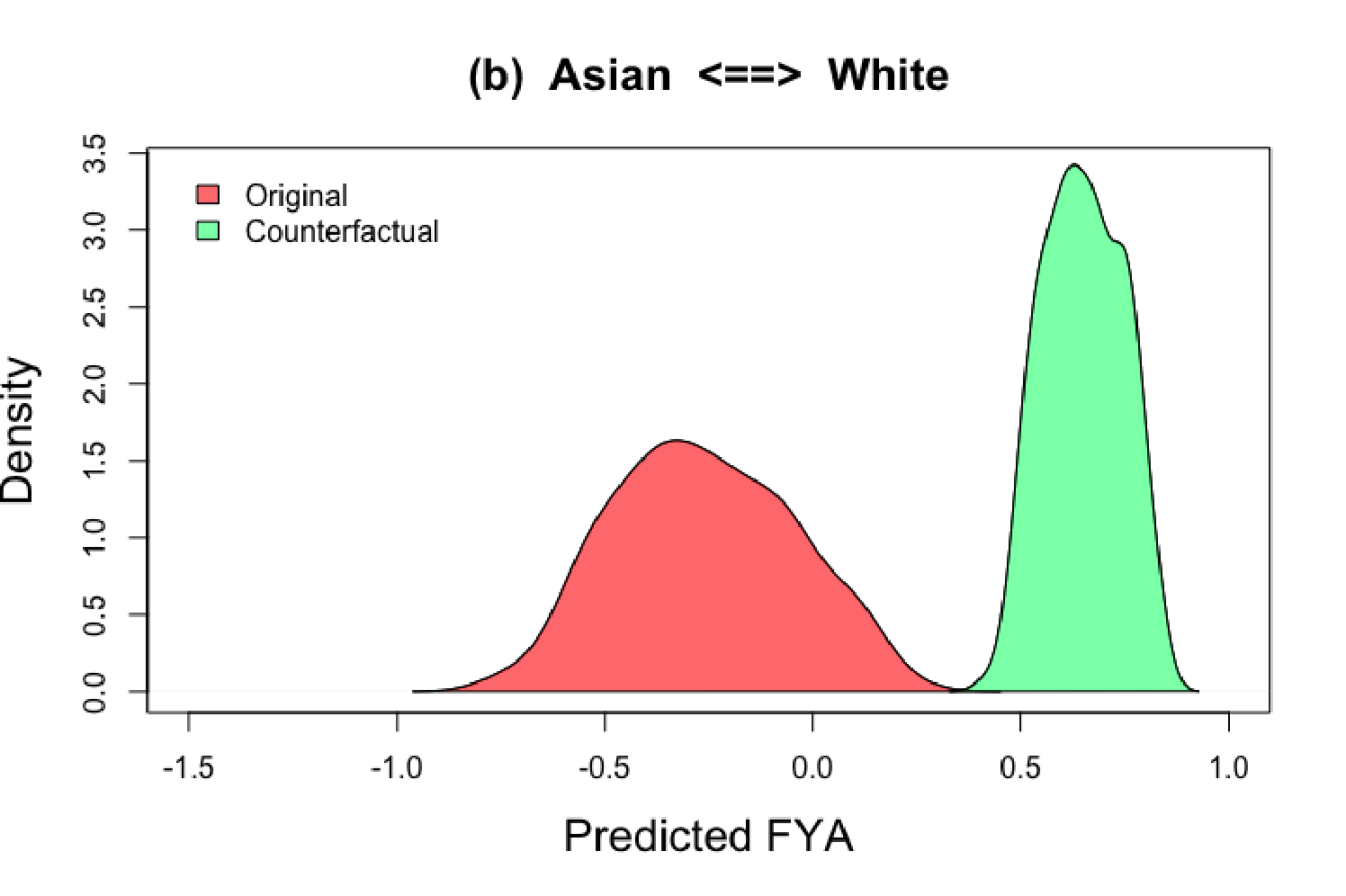}
    \end{minipage}& \begin{minipage}{.2\textwidth}
      \includegraphics[width=\textwidth, height=30mm]{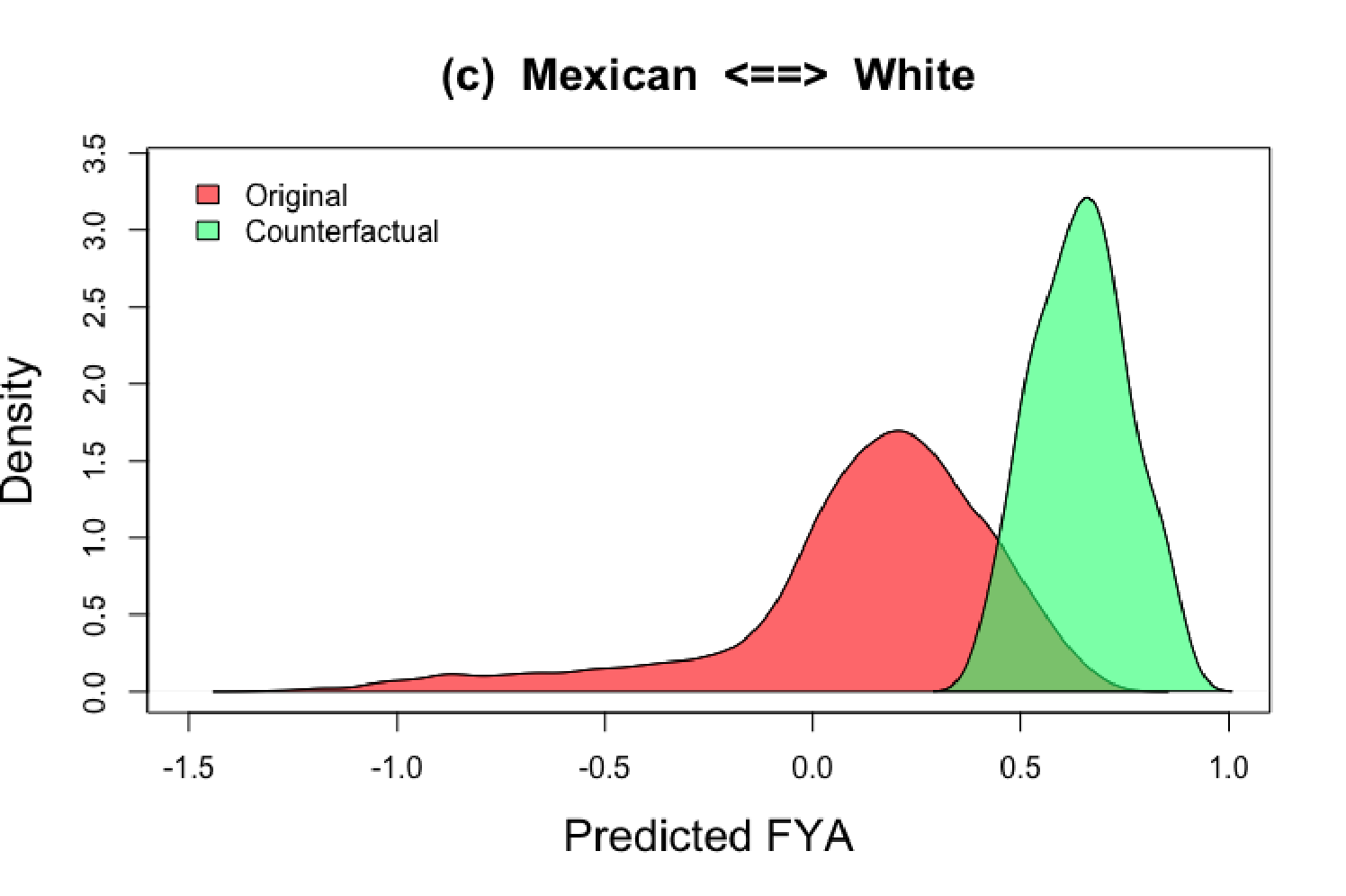}
    \end{minipage}\\
    \hline
    \multicolumn{3}{c}{Communities and crime}\\
    \hline
    \begin{minipage}{.2\textwidth}
      \includegraphics[width=\textwidth, height=30mm]{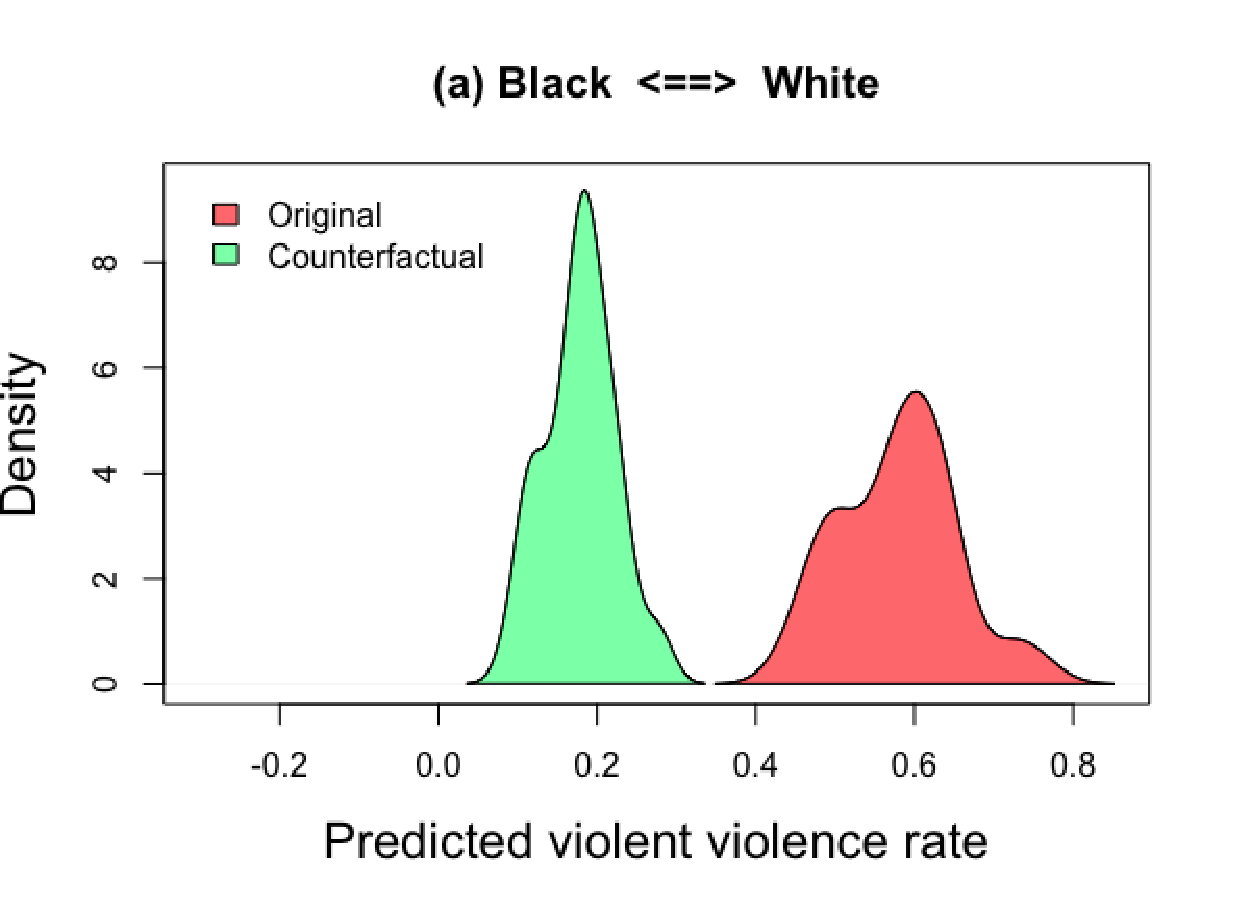}
    \end{minipage} & \begin{minipage}{.2\textwidth}
      \includegraphics[width=\textwidth,height=30mm]{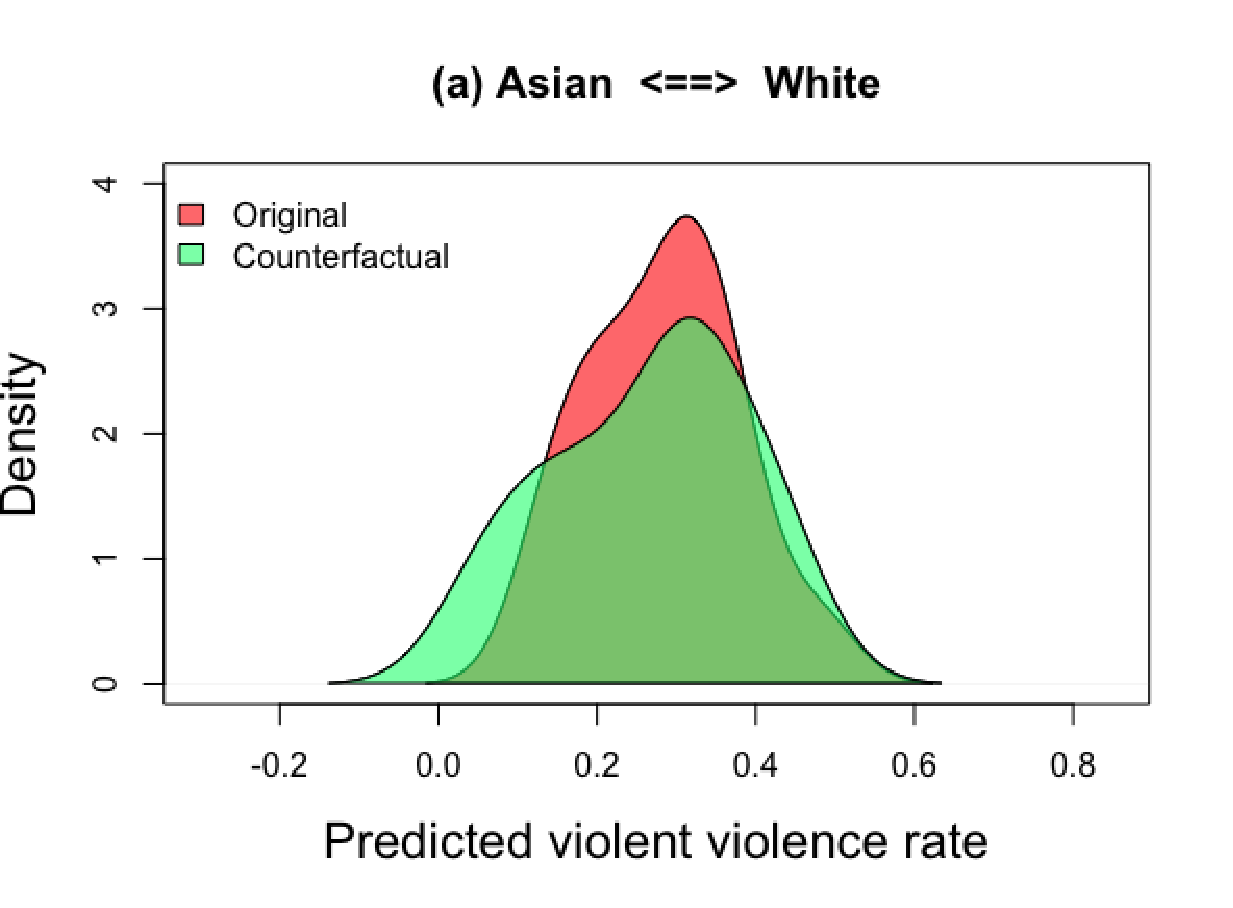}
    \end{minipage}& \begin{minipage}{.2\textwidth}
      \includegraphics[width=\textwidth, height=30mm]{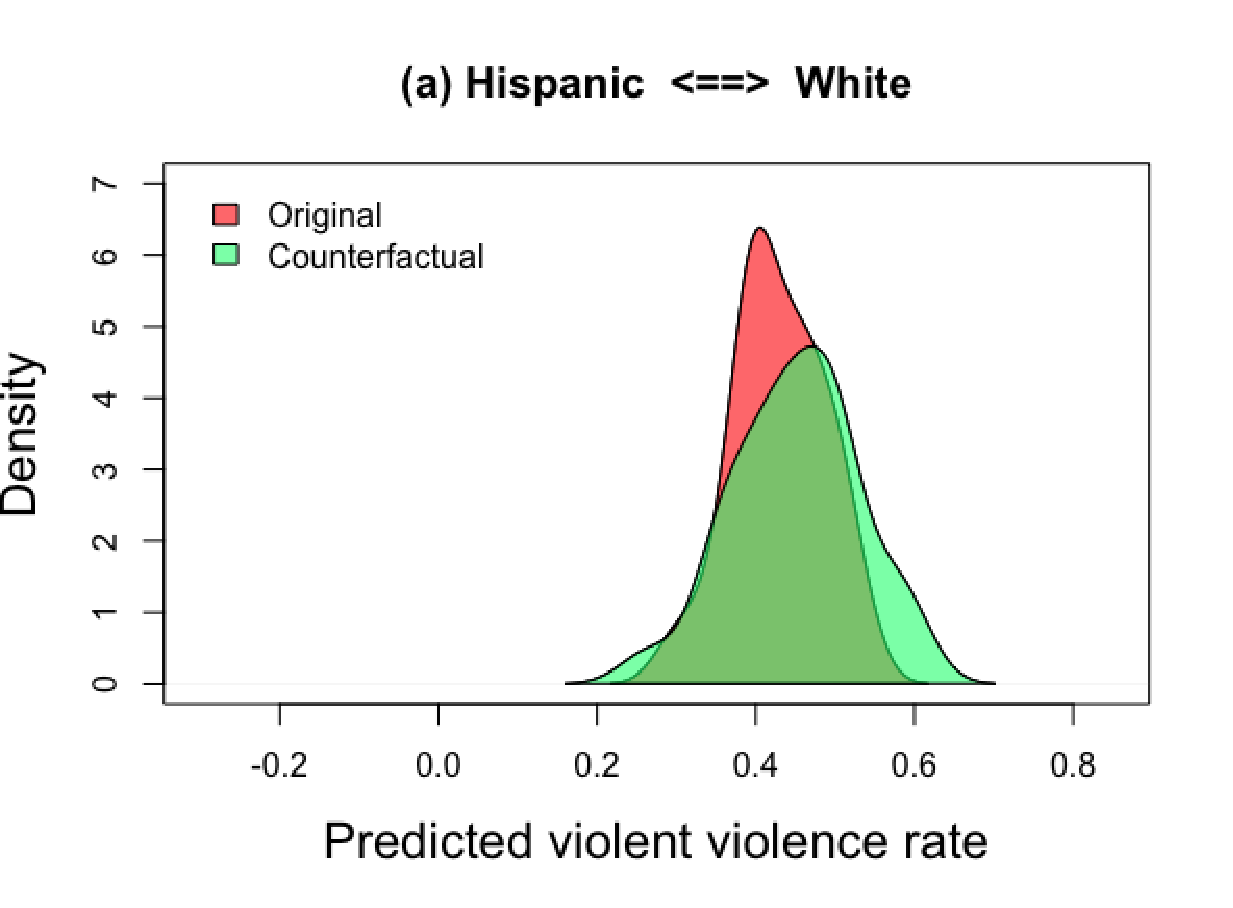}
    \end{minipage}\\
    \hline
  \end{tabular}
  \caption{Counterfactual learning and assessment of the law school and communities and crime datasets.}
  \label{tab:exp}
\end{table*}

\begin{figure}[!h]
    \subfigure [Law school.]  {%
    {\includegraphics [scale=0.25]{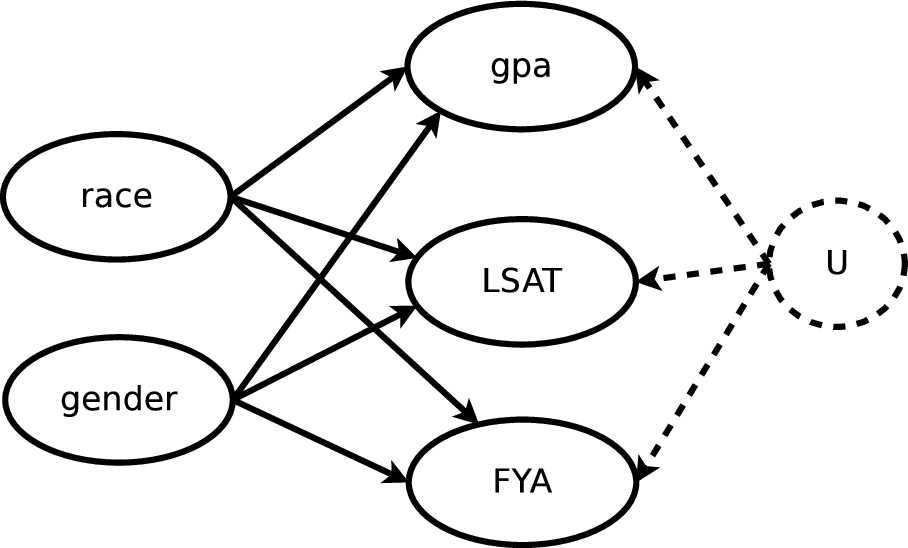} }
    \label{subfig:CGLaw}}
    \subfigure [Communities and crime.] {%
    {\includegraphics[scale=0.25]{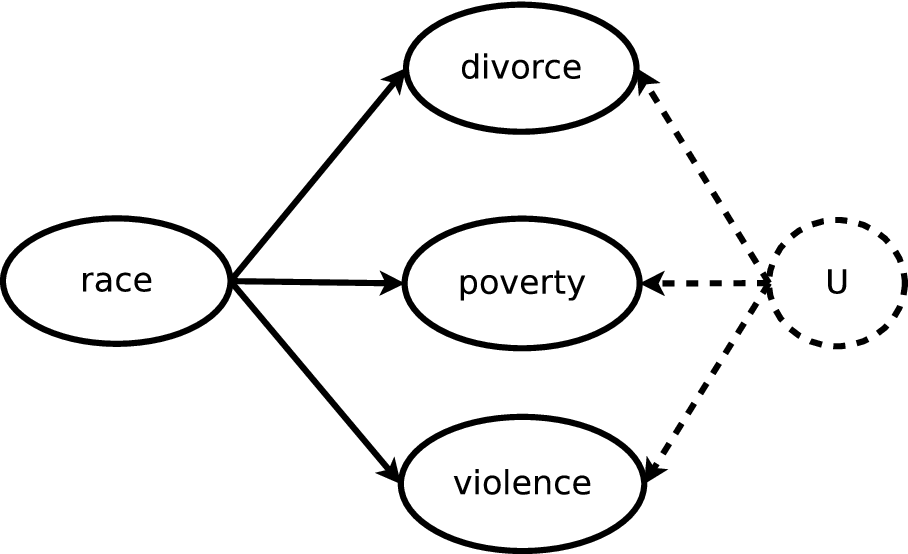} }
    \label{subfig:CGComm}}
    \caption{Causal graphs of the law school and the communities and crime datasets.}
    \label{fig:CG}
\end{figure}

Similarly to Kusner et al.~\cite{kusner2017counterfactual}, Stan programming language~\cite{team2016rstan} is used for counterfactual learning and assessment. However, in addition to the \textit{law school}~\cite{wightman1998lsac} used in Kusner et al., a second dataset is used in the empirical analysis, namely, \textit{communities and crime}\\ \cite{redmond2002data}. The \textit{law school} dataset~\cite{wightman1998lsac} includes information on $21,790$ law students such as their LSAT, their GPA collected prior to law school, and their first year average grade (FYA). Given this data, a school wishes to predict whether an applicant will have a high FYA. Race is considered as a sensitive attribute. \textit{Communities and crime} dataset~\cite{redmond2002data} contains information relevant to per capita violent crime rates in different communities in the United States (e.g., percentage of people under the poverty level, median family income, percentage of population who are divorced, etc,.). The goal is to predict crime rate. Race is considered as sensitive attribute.

For both datasets, the experiment consists in training a baseline model using logistic regression\footnote{The predictor uses all the variables in the dataset, including the sensitive variables to make the predictions.}, then applying  Algorithm~\ref{alg:two} to assess counterfactual fairness to both the original and counterfactual sampled data and plot how the distribution of predicted FYA changes for that baseline model.
Since sensitive attributes are not binary, the experiment is repeated for every counterfactual change (e.g., white vs black, white vs asian, etc.). If both distributions are superposed, it means the prediction is counterfactually fair. Far apart distributions indicate counterfactual bias.

Table~\ref{tab:exp} shows the density plots of $\hat{Y}$ for both datasets. We assume that the true model of the world is given by the causal graphs presented in Figure~\ref{fig:CG}. The causal graphs use a single latent/background variable $U$. For the \textit{law school}, $U$ can represent the knowledge of the student, while for the \textit{communities and crime}, it refers to the criminality of the individual which can capture other aspects of the individual that might have been used by the police. The red plots refer to the filtered original samples while the green plots refer to the corresponding counterfactual samples. For example, in plot (a), the red plot represents the predicted first year average ($\hat{FYA}$) for observed black students whereas the red plot represents $\hat{FYA}$ for the corresponding counterfactual samples. Density plots for the law school corroborate the findings of Kusner et al.~\cite{kusner2017counterfactual}, that is, racial differences exhibit counterfactual unfairness in favor of whites. The unfairness is peaked when predicting the FYA for white versus black students. The race-based discrimination is present in the \textit{communities and crime} result in a similar way: the counterfactual bias is peaked for white versus black communities. However, racial comparison can be considered counterfactually fair when comparing hispanic/white and asian/white communities. 
\subsection{Potential outcome estimation techniques}
\label{subsec:potEstimation}

Causal inference in the potential outcome framework focuses on estimating the causal effect of a treatment variable $A$ (e.g., the sensitive attribute) on an effect variable $Y$ (e.g., the decision outcome). As mentioned in Section~\ref{subsec:po}, there are three assumptions that are typically made for causal effect estimation, SUTVA, ignorability, and positivity. Inline with the potential outcome framework literature, this survey focuses on causal inference approaches that rely on the three assumptions~\cite{yao2020survey,guo2020survey}, namely, re-weighting~\cite{rubin2015book}, matching~\cite{morgan2015book}, and stratification~\cite{rubin2015book}.  

\subsubsection{Re-weighting}
\label{subsec:re-weighting}

\begin{table}[!h] 
	\centering
	\caption{Estimation of $ATE$ using inverse propensity weighting (IPW) on the job hiring example with propensity score $e(c_i)$ and balancing score $b(c_i)$.}
\label{tab:hiringExampleIPW} 
\begin{tabular}{cp{0.1cm}p{0.1cm}p{0.1cm}ccccp{0.1cm}p{0.1cm}p{0.1cm}cc}
\multicolumn{6}{c}{Female applicants} & & \multicolumn{6}{c}{Male applicants} \\
\multicolumn{6}{c}{(Treatment group)} & & \multicolumn{6}{c}{(Control Group)} \\
$i$ & $A$ & $C$ & $Y$ & $e(c_i)$ & $b(c_i)$ & & $i$ & $A$ & $C$ & $Y$ & $e(c_i)$ & $b(c_i)$ \\ \cline{1-6} \cline{8-13}
  1: & 1 & 0 & 1 & $\scriptstyle 2/3$ & $\scriptstyle 3/2$ & & 13: & 0 & 0 & 1 & $\scriptstyle 2/3$ & 3\\
  2: & 1 & 0 & 1 & $\scriptstyle 2/3$ & $\scriptstyle 3/2$ & &  14: & 0 & 0 & 0 & $\scriptstyle 2/3$ & 3\\
  3: & 1 & 0 & 0 & $\scriptstyle 2/3$ & $\scriptstyle 3/2$ & & 15: & 0 & 0 & 0 & $\scriptstyle 2/3$ & 3\\
  4: & 1 & 0 & 0 & $\scriptstyle 2/3$ & $\scriptstyle 3/2$ & & 16: & 0 & 0 & 0 & $\scriptstyle 2/3$ & 3\\
  5: & 1 & 0 & 0 & $\scriptstyle 2/3$ & $\scriptstyle 3/2$ & & 17: & 0 & 1 & 1 & $\scriptstyle 1/3$ & $\scriptstyle 3/2$\\
  6: & 1 & 0 & 0 & $\scriptstyle 2/3$ & $\scriptstyle 3/2$ & &  18: & 0 & 1 & 1 & $\scriptstyle 1/3$ & $\scriptstyle 3/2$\\
  7: & 1 & 0 & 0 & $\scriptstyle 2/3$ & $\scriptstyle 3/2$ & & 19: & 0 & 1 & 1 & $\scriptstyle 1/3$ & $\scriptstyle 3/2$\\
  8: & 1 & 0 & 0 & $\scriptstyle 2/3$ & $\scriptstyle 3/2$ & & 20: & 0 & 1 & 1 & $\scriptstyle 1/3$ & $\scriptstyle 3/2$\\
  9: & 1 & 1 & 1 & $\scriptstyle 1/3$ & 3 & & 21: & 0 & 1 & 0 & $\scriptstyle 1/3$ & $\scriptstyle 3/2$\\
  10: & 1 & 1 & 1 & $\scriptstyle 1/3$ & 3 & & 22: & 0 & 1 & 0 & $\scriptstyle 1/3$ & $\scriptstyle 3/2$\\
  11: & 1 & 1 & 1 & $\scriptstyle 1/3$ & 3 & & 23: & 0 & 1 & 0 & $\scriptstyle 1/3$ & $\scriptstyle 3/2$\\
  12: & 1 & 1 & 0 & $\scriptstyle 1/3$ & 3 & &  24: & 0 & 1 & 0 & $\scriptstyle 1/3$ & $\scriptstyle 3/2$\\
 \cline{1-6} \cline{8-13}
\end{tabular}  
\end{table}

One of the main challenges in causal inference is that the sensitive attribute is not assigned in random in observational data. That is, the distribution in the observed dataset does not reflect the true distribution. Sample re-weighting methods try to overcome this discrepancy by assigning appropriate weights to sample units in the observational data. The aim is to generate a pseudo-population on which the distributions of the protected (e.g., female) and unprotected (e.g., male) groups are the same as in the original total population. This is achieved by defining a balancing score $b(x)$ satisfying $A \perp x\;|\;b(x)$. The most common approach to balancing score is based on the propensity score~\cite{rubinPropensity83} which is defined as the conditional probability of sensitive attribute given background variables:
\begin{equation}
    e(x) = \pr(A=1\;|\;X=x) \label{eq:prop}
\end{equation}
Propensity scores can be used to equate groups based on covariates $X$. In inverse propensity weighting (IPW), the balancing score $b(x)$ for each sample is defined as:
\begin{equation}
    b(x) = \frac{A}{e(x)} + \frac{1-A}{1-e(x)} \label{eq:balScore}
\end{equation}
where $A=1$ corresponds to the protected group and $A=0$ corresponds to the unprotected group. The IPW estimator of $ATE$ (Eq.~\ref{eq:ATE}) is defined as:
\begin{equation}
    \hat{ATE}_{IPW} = \frac{1}{n}\sum_{i=1}^{n}\frac{A_i Y_i}{\hat{e}(x_i)} - \frac{1}{n}\sum_{i=1}^{n}\frac{(1-A_i)\; Y_i}{1-\hat{e}(x_i)} \label{eq:ipw}
\end{equation}
Notice that the estimation of $ATE$ is based only on the observable outcome (no counterfactual outcomes) and on the estimation of $e(x_i)$, that is, $\hat{e}(x_i)$.

Table~\ref{tab:hiringExampleIPW} shows the values of propensity ($e(c_i)$) as well as balance ($b(c_i)$) scores for each unit $i$ in the simple job hiring example. Using Eq.~\ref{eq:ipw}, the $\hat{ATE}_{IPW}$ estimation of $ATE$ is $0.25$ indicating a discrimination in favor of the female group. 

When the propensity score is estimated, the normalized version of $\hat{ATE}_{IPW}$ is preferred:
\begin{equation}
    \hat{ATE}_{IPW}^{norm} = \left[ \sum_{i=1}^{n}\frac{A_i Y_i}{\hat{e}(x_i)} \middle/\sum_{i=1}^{n}\frac{A_i}{\hat{e}(x_i)} -  \sum_{i=1}^{n}\frac{(1-A_i)\; Y_i}{1-\hat{e}(x_i)} \middle/\sum_{i=1}^{n}\frac{(1-A_i)}{1-\hat{e}(x_i)} \label{eq:ipw} \right]
\end{equation}

$\hat{ATE}_{IPW}^{norm}$ for the same example equals $0.125$ which is a perfect estimation of $ATE$ in this case as both values coincide ($ATE = 0.125$). 

The correctness of the IPW estimation relies heavily on the quality of the propensity score estimation ($\hat{e}(X)$). A slight misspecification of propensity scores may lead to significant discrepancy in the $ATE$ estimation. In such cases, doubly robust ($DR$) estimation is recommended~\cite{doublyRobust2011}. $DR$ combines IPW estimation with outcome regression so that the estimation remains valid even if one of the approaches is incorrect (but not both). Another limitation of IPW can be observed if the propensity score $e(X) = \pr(A\;|\;X)$ for some value of $X$ is small. In such case, the estimation may suffer instability. To address this issue, trimming~\cite{trimming2011} is typically used. Trimming consists in removing the samples with a propensity score less than a certain threshold.

\subsubsection{Matching}
\label{subsec:matching}

Matching techniques~\cite{morgan2015book} focus on estimating the counterfactual outcome of units. The idea is to estimate the counterfactual outcomes $Y^1_{i}|A=0$ and $Y^0_{i}|A=1$ based on the matched neighbours of unit $i$ in the opposite group. For example, given an observed female candidate $f_k$, estimating the counterfactual outcome (hiring decision) had she been a male is based on the units in the male group that are the most comparable to $f_k$. Hence, the first and main issue is to define a similarity metric between two given units (e.g. $x_i$ and $x_j$). The most common approach is to rely on the propensity scores of units:
\begin{equation}
    D(i,j) = |e(x_i) - e(x_j)| \label{eq:matchingDist1}
\end{equation}
and its logit version:
\begin{equation}
    D(i,j) = |logit(e(x_i)) - logit(e(x_j))| \label{eq:matchingDist1}
\end{equation}
which is preferred as it has been proven to reduce the bias~\cite{matching2010stuart}.

The second issue is the matching algorithm, that is, how many neighbours to consider and how these neighbours are weighted to obtain the estimation? Matching algorithms include~\cite{morgan2015book}:
\begin{itemize}
    \item Exact matching: uses only identical matches. Typically infeasible since many units will remain unmatched.
    \item Nearest neighbour matching (NNM): constructs the counterfactual using the closest neighbours according to a similarity metric. It can run with or without replacement, that is, by returning a matched unit to the pool or not. The no replacement variant makes the estimation dependent on the order in which the units are matched.
    \item Caliper matching: a version of NNM that restricts matching to a chosen maximum distance. Its main problem is that some units may not receive matches because no neighbours fall within their caliper. A hybrid variant consists in using caliper, and in case of no neighbours, select a matching neighbour outside the caliper.
    \item Radius matching: the same as caliper, but matching is done with replacement.
    \item Kernel matching: an extension to all the above algorithms where to match a unit $i$, all units in the opposite group are used. Each unit is weighted according to the distance to the unit $i$. 
\end{itemize}

\begin{table*}[]
\centering
\caption{Characteristics of the real-world datasets.}
\label{tab:datasetsInfo}
\setlength{\tabcolsep}{6pt}
\begin{tabular}{l|l|l|l|l|l}
\hline
\multirow{2}{*}{Dataset} & Sample & Sensitive  & \multirow{2}{*}{Covariate(s)}     &  \multirow{2}{*}{Mediator(s)} & \multirow{2}{*}{Outcome} \\

            & size   & feature(s) &    &     &    \\
\hline
\hline
Communities & \multirow{2}{*}{$1994$} & \multirow{2}{*}{race}   & age & poverty rate &  \multirow{2}{*}{crime rate} \\
  and crimes &  &  & unemp. rate & divorce rate &  \\
\hline
\multirow{2}{*}{Compas} & \multirow{2}{*}{$5915$} & \multirow{2}{*}{race}   & age& \multirow{2}{*}{priors}  & \multirow{2}{*}{recidivism}\\
 &    &   & gender &  &  \\
\hline
\multirow{2}{*}{German credit} & \multirow{2}{*}{$1000$}   & gender & \multirow{2}{*}{age}& \multirow{2}{*}{emp. length} & \multirow{2}{*}{default}\\
&  & marital status &  &  &   \\
\hline
Berkeley & $4526$ & gender   & department& department & admission\\
\hline
\end{tabular}
\end{table*}

\subsubsection{Stratification}
\label{subsec:strat}

Stratification~\cite{rubin2015book} uses the same principle underlying identifiability approach (Section~\ref{sec:ident}), that is, adjusting on confounders. The aim is to split the entire observed data into consistent groups such that the units in the same group can be considered as sampled from data under RCT. The two ingredients of stratification are the splitting of groups and then the combination of the created groups. The stratification estimator of $ATE$ can be defined generically as:
\begin{equation}
    \hat{ATE}^{strat} = \sum_{k=1}^{K}m(k)[\overline{Y}_1(k) - \overline{Y}_0(k)] \label{eq:atestrat}
\end{equation}
where $K$ is the number of stratification groups, $m(k)$ is the portion of units in group $k$ to the total number of units $N$, $\overline{Y}_1(k)$ and $\overline{Y}_(k)$ are the $CATE$ (Eq.~\ref{eq:CATE}) for groups $A=1$ and $A=0$, respectively. $\hat{ATE}^{strat}$ expression has the same structure as the back-door formula (Eq.~\ref{eq:int2}). 

If all variables needed for the stratification are observed and the available data is infinitely large, $ATE^{strat}$ can lead to a consistent and unbiased estimator of $ATE$. However, in typical datasets, stratification may result in strata with few or no units. Consequently, some $CATE$ estimates cannot be calculated with the available data. Propensity score can be used to address this data sparseness problem. The main idea is the following: ``strata with identical propensity scores can be combined into more coarse strata''~\cite{morgan2015book}. In other words, propensity score can be considered as a single stratifying variable that will usually result in larger strata. The same idea is used in the SCM framework to address the sparseness of data when computing identifiable expressions. 

Other estimation methods in the potential outcome framework include tree-based methods~\cite{treemethods16imbens}, representation learning methods~\cite{representation2007}, and meta-learning methods~\cite{metalearners2019}.

\subsection{Estimating causal effects on benchmark datasets}
\label{subsec:empirical}
We conduct experiments on four real-world datasets which are commonly used in discrimination-discovery literature, namely: \textit{communities and crime}~\cite{redmond2002data}, \textit{Compas}~\cite{angwin2016machine}, \textit{German credit}~\cite{verma2018fairness} and \textit{Berkely admission}~\cite{freedman1998statistics}.

\textit{Communities and crime} dataset~\cite{redmond2002data} contains information relevant to per capita violent crime rates in $1994$ different communities in the United States and the goal is to predict that crime rate. Race is kept as sensitive attribute. However, in these experiments, we transformed the label into a binary feature by thresholding\footnote{The median value of the violent crime rate in the dataset is used as threshold.} where $1$ corresponds to high violent crime rate and $0$ corresponds to low violent crime rate. \textit{Compas} dataset~\cite{angwin2016machine} includes information of $5915$ individuals from Broward County, Florida, initially compiled by ProPublica~\cite{angwin2016machine} and the goal is to predict the two-year violent recidivism. That is, whether a convicted individual would commit a violent crime in the following two years ($1$) or not ($0$). We consider race as sensitive feature. \textit{German credit} dataset comprises data of $1000$ individuals applying for loans. The goal is to predict whether an individual will default on the loan $(1)$ or not $(0)$. The gender and the personal status of an individual are considered sensitive features. Finally, \textit{Berkeley admission}~\cite{freedman1998statistics} dataset consists of $4526$ applicants to UC-Berkeley's graduate programs in Fall $1973$. The goal is to predict whether a student is admitted $(1)$ or not $(0)$. Gender is treated as sensitive feature. \textit{Berkeley admission} is commonly used to illustrate the Simpson's Paradox~\cite{simpson1951interpretation}. 

Table~\ref{tab:datasetsInfo} summarizes the main characteristics of the four benchmark datasets. In the potential outcome framework, estimation techniques rely on the selection of a set of covariates and mediators. Column four lists the covariate(s)  while column five lists the mediator variables for each dataset. A covariate can be a coufounder or any other variable that might distort the causal effect estimation. A mediator is a covariate which is at the same time influenced by the sensitive feature and influences the outcome\footnote{Graphically, a mediator is a variable in the causal path between the sensitive feature and the outcome i.e. an explaining or redlining (proxy) variable.}. 
In addition to total variation $TV$ (statistical parity) serving as baseline measure and also to illustrate the need for causality, we use estimation techniques for seven causal-based fairness notions defined in the potential outcome framework, namely, $ATE\_{IPW}$, $ATE\_{match}$, $ATE\_{DR}$, $ATE\_{strat}$, $TE\_{imp}, DE\_{imp}$, and $IE\_{imp}$. The four first notions are estimations of $ATE$ using four different approaches. $ATE\_{IPW}$  uses inverse propensity weighting as explained in Section~\ref{subsec:re-weighting}. The variables used to estimate the propensity score are the covariates listed in column $4$ of Table~\ref{tab:datasetsInfo}. $ATE\_match$ is matching estimation of $ATE$ (Section~\ref{subsec:matching}). Recall that this approach matches each unit (individual) with one or more ``similar'' units in the other group. Propensity score matching is used.
$ATE\_DR$ is a double robust estimation of $ATE$ which requires at least propensity score estimation or outcome regression to be correct as explained in Section~\ref{subsec:re-weighting}. $ATE\_strat$ is a startification-based estimation of $ATE$ (Section~\ref{subsec:strat}). For all these $ATE$ estimations, we used the \textit{causallib} package implementations~\cite{shimoni2019evaluation}. The three remaining measures, namely, $TE\_imp$, $DE\_imp$ and $IE\_imp$ are imputation-based estimations of total, direct, and indirect effects. The main idea behind imputation approach is to provide $K+1$ outcome models that characterize $\ep[Y | \textbf{X},A]$, $\ep[Y | \textbf{X},A,M_1],\; \ldots\; \ep[Y | \textbf{X},A,M_k]$, where $\textbf{X}$ is the set of the covariates (column $4$ in Table~\ref{tab:datasetsInfo}) and $M_1,\ldots, M_k$ are the mediators (column $5$ in Table~\ref{tab:datasetsInfo}). The imputation approach involves modeling the conditional means of the outcome variable $Y$, given the sensitive feature $A$, pretreatment confounders $X$, and varying sets of mediators $M$. The \textit{paths} package implementation~\cite{zhou2020tracing} is used to compute these estimations.

\begin{figure*}[!h]
    \includegraphics[scale=0.45]{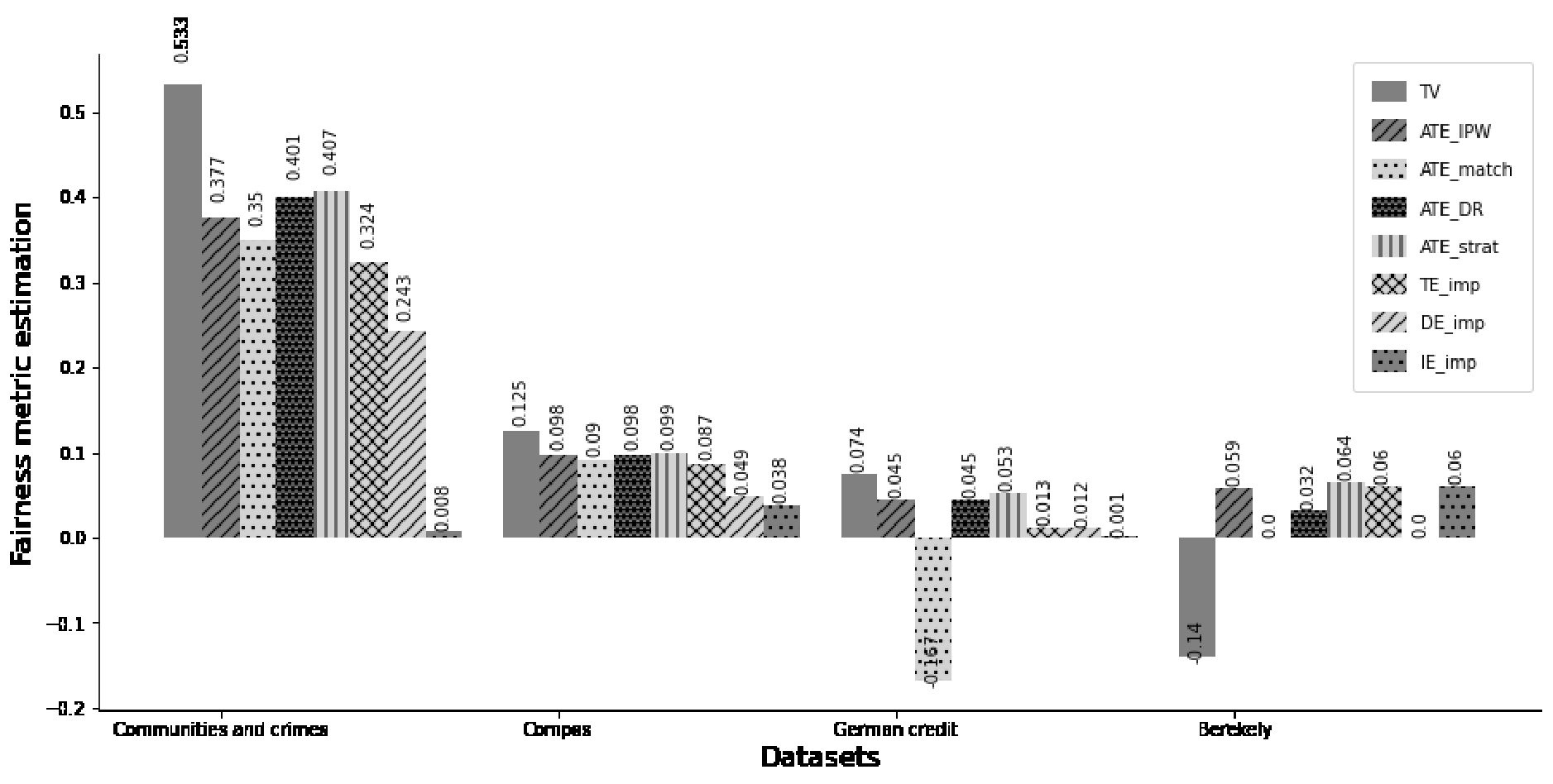} 
    \caption{Estimation of causal effects on real-world datasets.}
    \label{fig:EstimAll}
\end{figure*}

Figure~\ref{fig:EstimAll} shows the causal effects (of the sensitive attribute $A$ on the outcome $Y$ values according to the seven estimators on the four benchmark datasets. The values are between $-1$ (full discrimination against one group) and $1$ (full discrimination against the second group). A value $0$ indicates fairness of the outcome with respect to both groups.  

For the three first datasets (\textit{Communities and crimes, Compas,} and \textit{German credit}), $TV$ overestimates the discrimination as it results in higher values compared to all causal-based notions (except $ATE\_match$ in \textit{German credit} where the results are reverted. In other words, the discrimination is now against male applicants: $-0.167$). For \textit{Berekley}, $TV$ concludes a significant discrimination against women ($-0.14$) while causal-based notions show either the absence of discrimination ($0$ for $ATE\_match$ and $DE\_imp$) or a discrimination in favor of women which confirms the Simpson's paradox. For the three last estimations ($TE\_{imp}, DE\_{imp}$, and $IE\_{imp}$), notice that $TE$ is approximately the total of $DE$ and $IE$ which is expected as the total effect is composed of direct and indirect effects. 

\begin{figure*}[!h]
    \subfigure [Communities and crimes] {%
    {\includegraphics[scale=0.25]{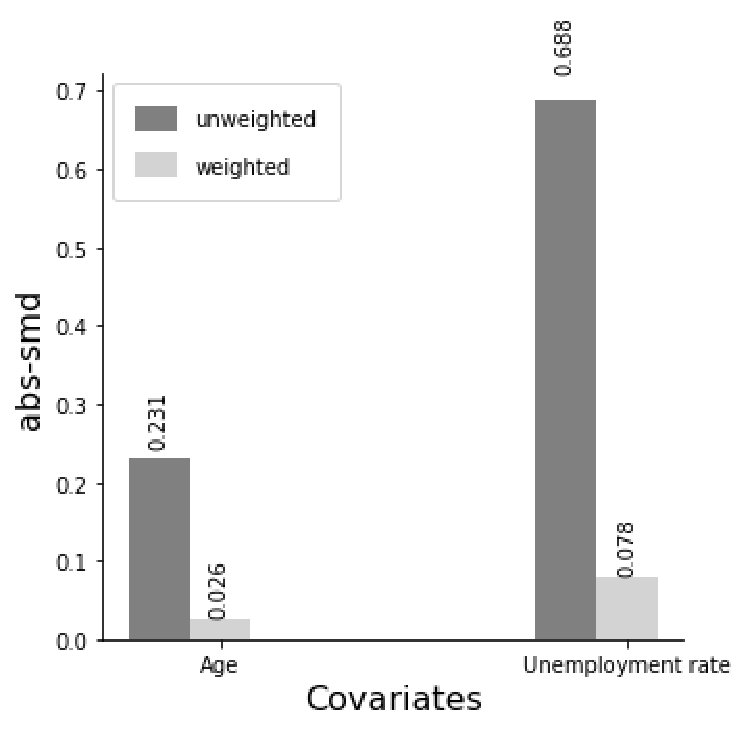} }
    \label{subfig:balance1}}
    \quad
    \subfigure [Compas]  {%
    {\includegraphics [scale=0.25]{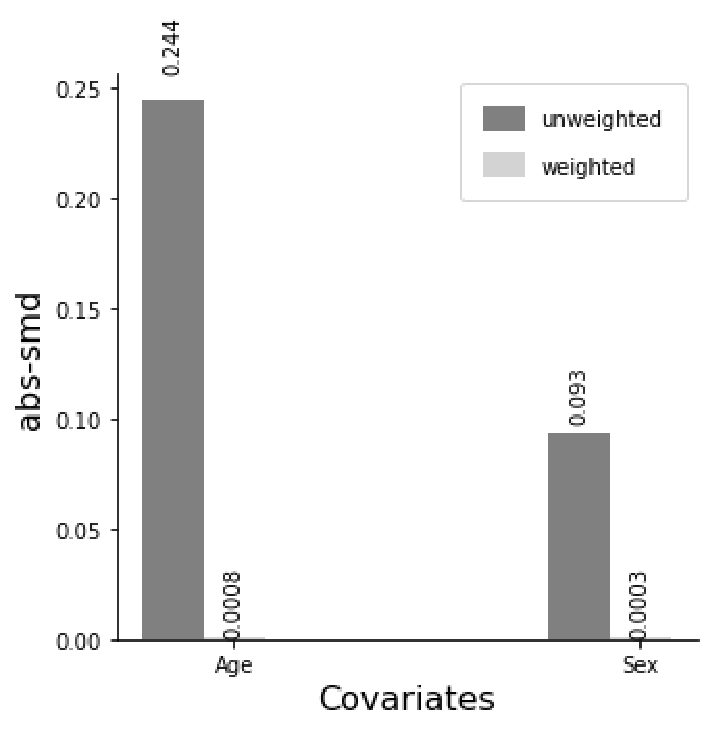} }
    \label{subfig:balance2}}
    \quad
    \subfigure [German credit]  {%
    {\includegraphics [scale=0.25]{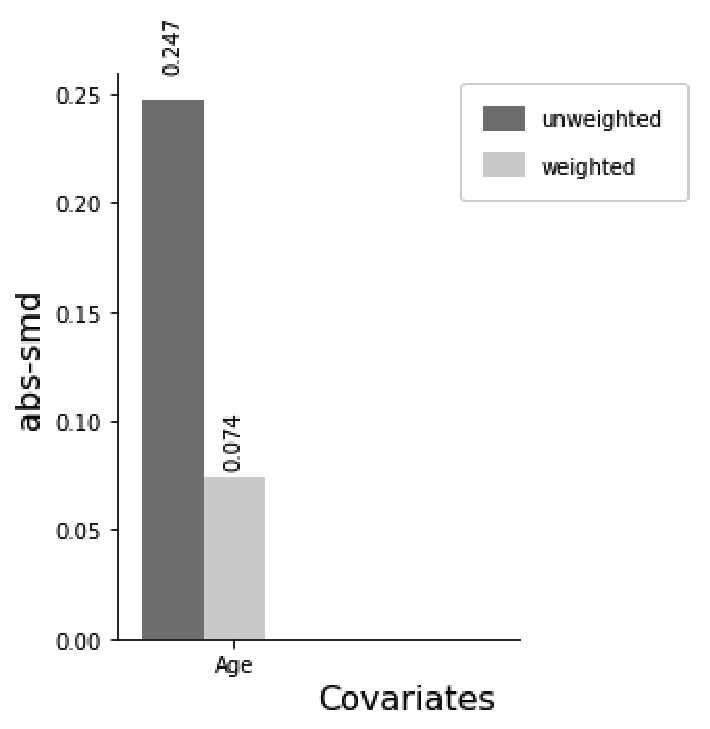} }
    \label{subfig:balance3}}
    \quad
    \subfigure [Berkeley]  {%
    {\includegraphics [scale=0.25]{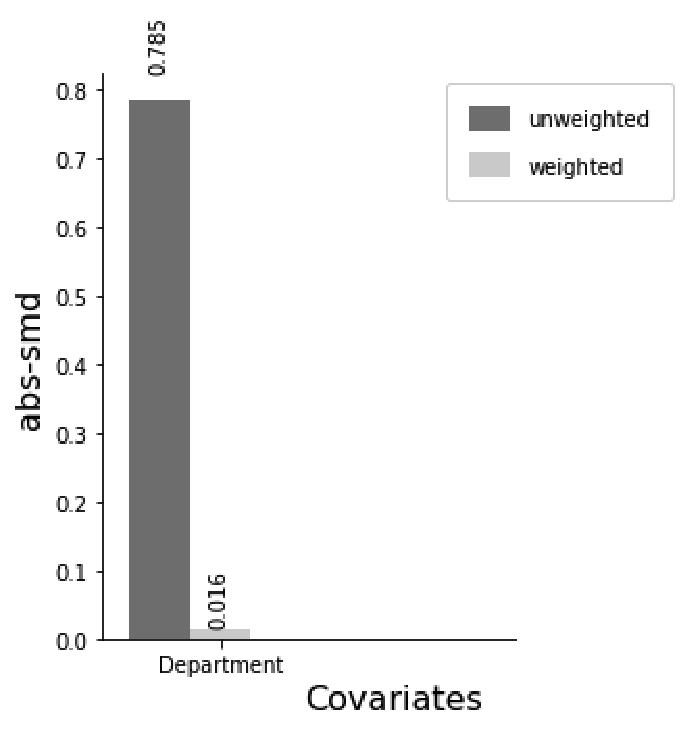} }
    \label{subfig:balance4}}
    \caption{Absolute standard mean difference of covariate values of different groups (e.g. male vs female) prior and posterior to IPW re-weighting.}
    \label{fig:balanceCovariates}
    \vspace{-3mm}
\end{figure*}

As mentioned earlier, estimation techniques aim to balance the control and the treatment groups in order to remove (or to mitigate) potential selection bias. For the particular case of re-weighting based estimation of $ATE$ ($ATE\_{IPW}$), one way to assess whether the balancing worked is to use the absolute standard mean difference (\textbf{abs-smd})~\cite{andrade2020mean} between the treatment and the control groups for each considered covariate (Column 4 in Table~\ref{tab:datasetsInfo}). That is, measuring the difference in means between groups, divided by the (pooled) standard deviation. Figure~\ref{fig:balanceCovariates} shows the \textbf{abs-smd} for each covariate in the four datasets prior (unweighted) and posterior\\ (weighted) to applying IPW re-weighting. One can observe the same pattern in all benchmark datasets. The unweighted \textbf{abs-smd} across control and treatment groups is at least one order of magnitude higher than the weighted \textbf{abs-smd}. For instance, the unweighted \textbf{abs-smd} across control and treatment groups reached $0.68$ standard-deviations for the \textit{Unemployment rate} covariate in the \textit{communities and crime}s dataset. After IPW re-weighting, the \textbf{abs-smd} is significantly reduced ($0.078$). 

\section{Suitability and applicability}
\label{sec:applicability}

\begin{figure*}[!h]
    \includegraphics[scale=0.2]{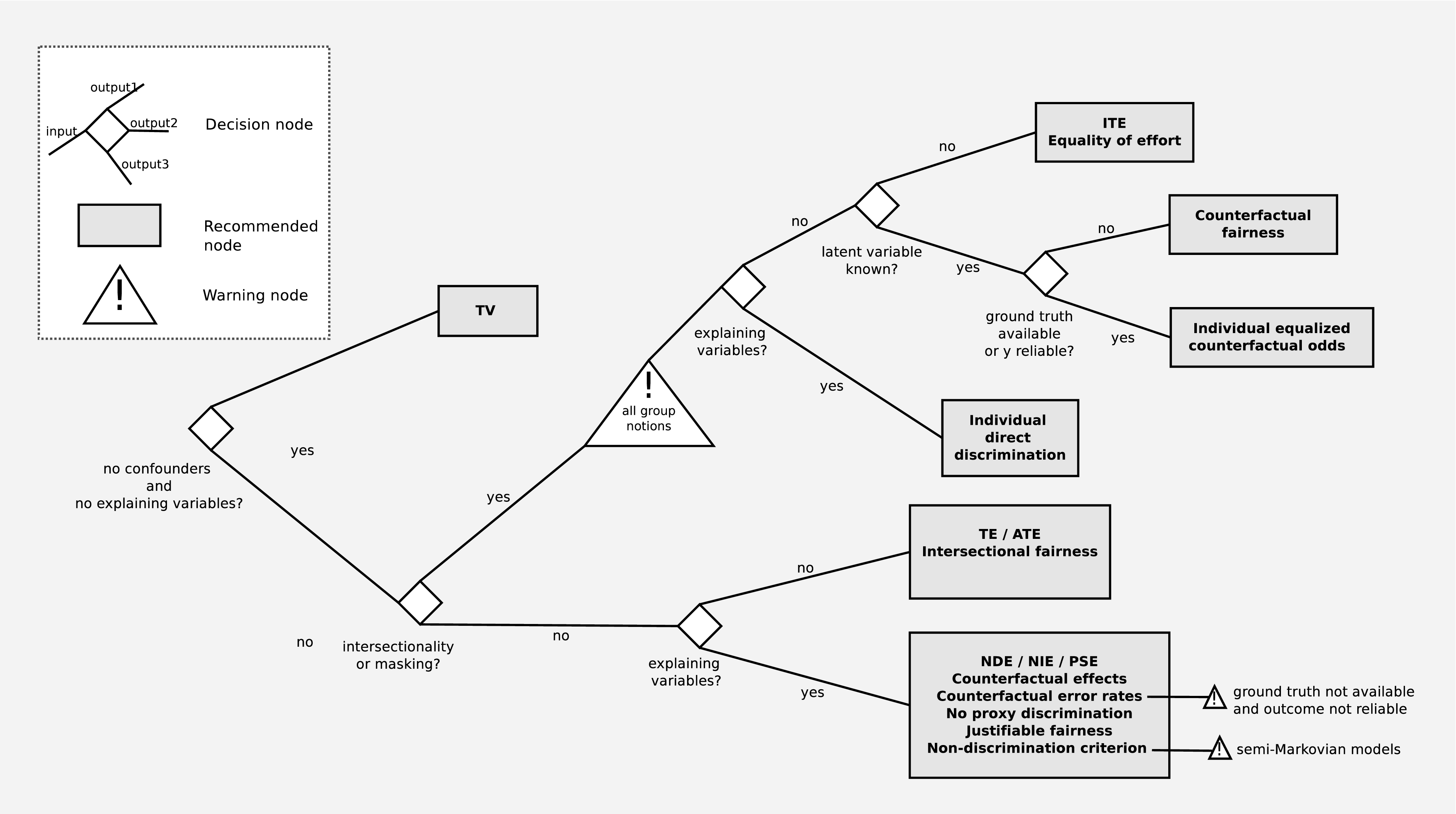} 
    \caption{Guideline for causal-based fairness notions selection}
    \label{fig:decision_diagram}
\end{figure*}

Section~\ref{sec:notions} lists 19 causal-based fairness notions. Given a real-world scenario, selecting which fairness notion to use is a challenging and error-prone task as using the wrong fairness notion may indicate unfairness in an otherwise fair scenario, or the opposite (failing to detect unfairness in an unfair scenario). On the other hand, according to Pearl's SCM framework, computing causal quantities (interventions and counterfactuals) depends on their identifiability. Hence, even if a fairness notion is appropriate in some setup, it might not be applicable because of identifiability issues. The two following subsections address the suitability and the applicability of causal-based fairness notions

\subsection{Suitability}

In this section, we try to systemize the selection process by considering the subtleties of each causal-based fairness notion and defining 6 criteria which correspond to characteristics of the real-world scenario at hand. For each criterion, we check whether it holds in the scenario at hand or not. Then, use these answers to recommend the most suitable causal-based fairness notion. The criteria are list and briefly described as follows.
\begin{itemize}
    \item \textbf{Presence of confounding}: A variable which is a common cause of two or more other variables.
    \item \textbf{Presence of explaining variable}: A variable that is correlated with the sensitive attribute such that any discrimination that is explained using that variable is considered legitimate and is acceptable.
    \item \textbf{Likelihood of intersectionality}: A specific type of bias due to the combination of sensitive attributes. An individual might not be discriminated based on race only or based on gender only, but she might be discriminated because of a combination of both.
    \item \textbf{Likelihood of masking}: A form of intentional discrimination that allows decision makers with prejudicial views to discriminate against individuals or groups while masking their intentions. 
    \item \textbf{Latent variables are known}: Latent (background) variables are not observable. However, in some scenarios, they are identified and their relationships with observable variables are known.
    \item \textbf{Ground truth or reliable outcome}: the label in the training data can or cannot be reliable. In several scenarios, the outcome is inferred by humans (job hiring, college admission, etc.) and hence can encode bias. The most reliable outcome is when the ground truth is available\footnote{An example of a scenario where ground truth is available is when predicting whether an individual has a disease. The ground truth value is observed by submitting the individual to a blood test (Assuming the blood test is flawless) for example. An example of a scenario where ground truth is not available is predicting whether a job applicant is hired. The outcome in the training data is inferred by a human decision maker which is often a subjective decision, no matter how hard she is trying to be objective.}.
\end{itemize}
The diagram in Figure~\ref{fig:decision_diagram} can be used as a guideline to select an appropriate causal-based fairness notion given a real-world scenario.

Confounding variables result in backdoor paths between the sensitive attribute ($A$) and the outcome ($Y$). For example, path\\ $A \longleftarrow C \longrightarrow Y$ in Figure~\ref{fig:fig33} is a backdoor path. Backdoor paths are not causal paths, but they contribute to the association between the $A$ and $Y$. Therefore, they are the reason why it is said that ``correlation is different than causation''. In the absence of confounding, the total causal effect ($TE$ and $ATE$) coincides with the difference in conditional probabilities $TV = \pr(y|a_1) - \pr(y|a_0)$ which correspond to statistical parity. On the other hand, if there are no explaining variables in the model representation of the world, both direct and indirect causal paths are discriminatory\footnote{Indirect causal paths all go through proxy variables.}. Consequently, assessing unfairness/bias due to the sensitive attribute does not require considering separately the different causal paths (direct, indirect, and path-specific). In such case (absence of confounding and explaining variables), causal inference is not needed to appropriately assess fairness. 

Any unintentional type of bias can also be "orchestrated" intentionally by decision makers with prejudicial views. To appropriately assess the bias in presence of such masking attempts, it is recommended to avoid group-based notions as they can more easily be gamed by prejudicial decision makers. Intersectionality is similar to masking as both lead to a discrimination which is difficult to detect at the group-level and hence require more fine-grained measures. Therefore individual causal-based fairness notions are recommended in presence of one of those criteria. For individual notions, in presence of explaining variables, it is recommended to use individual direct discrimination (Section~\ref{sec:idd}) as it is the only individual notion listed in Section~\ref{sec:notions} that distinguishes direct from indirect discrimination. Counterfactual fairness (Section~\ref{sec:counterfactual}) and individual equalized counterfactual odds (Section~\ref{sec:IndECOD}) are recommended to be used in case the latent variables are known. If the ground-truth is not available or the outcome $Y$ is not reliable, individual equalized counterfactual odds is not recommended. 

For the group causal-based fairness notions, if there are no explaining variables, there is no need to consider the different causal paths and hence $TE$, $ATE$, or interventional fairness (Section~\ref{sec:justif}) can be safely used. In presence of explaining variables, the remaining causal-based fairness notions are appropriate to use with two exceptions. First, non-discrimination criterion is misleading if the causal model is semi-Markovian because the variables $A$ can remain dependent even after conditioning on all observable variables because of the hidden counfounders. Second, as counterfactual error rates (Section~\ref{sec:cer}) are expressed in terms of the true outcome $Y$, they are not recommended in case the ground-truth is not available and the true outcome is not reliable. 

Finally, note that $ETT$, $ATT$, and $ATC$ are not generally used in fairness scenarios because, typically, the bias can be observed in both directions: when considering a disadvantaged group/individual as advantaged or the opposite. $ETT$ is relevant when studying the effect of a treatment medicine on patients. For example, if a patient agrees to take the medicine and it turns out to be painful, she may be wondering about the chances of recovery if she did not take the treatment or if she took it with a lower dose. The opposite direction (the effect of treating an individual in the control group) is not relevant in this case.

In his book, \textit{The Book of Why}~\cite{pearl2018book}, Pearl describes a causation ladder with three rungs: statistical observations (seeing), intervention (doing), and counterfactual (imagining). In this section, all causal-based fairness notions defined in Pearl's SCM framework (all notions in Section~\ref{sec:notions} except $ATE$, $ATT$, $ATC$, $ITE$, and equality of effort) are placed in the causation ladder which will help us address the problem of their applicability in real-scenarios. The causation ladder is structured in such a way that a quantity at a certain rung can be identified in terms of quantities at the rung just below it. As a consequence, the higher the rung, the more challenging the problem of identifiability, and hence the less applicable a fairness notion defined at that rung. 

The diagram in Figure~\ref{fig:diagram} shows the causation ladder and indicates at which rung every causal-based fairness notion stands. $TV$ which is the only non-causal fairness notion covered in this paper is at rung 1. It is always applicable provided that a set of observations (dataset) is available. No unresolved and non-discrimination criteria are placed midway between rungs 1 and 2 as they are applicable provided that the causal graph is available along the dataset. Non-discrimination criterion, however, requires the Markov property to be applicable because causal dependence through unobservable paths cannot be blocked. It also has an exponential complexity since it considers all combination of values of the parent variables of the outcome $Y$. A relaxation is described by the authors~\cite{zhang2017achieving} but the notion remains computationally intractable. 

\subsection{Applicability}

\begin{figure*}[!h]
	\centering
	{\includegraphics[scale=0.3]{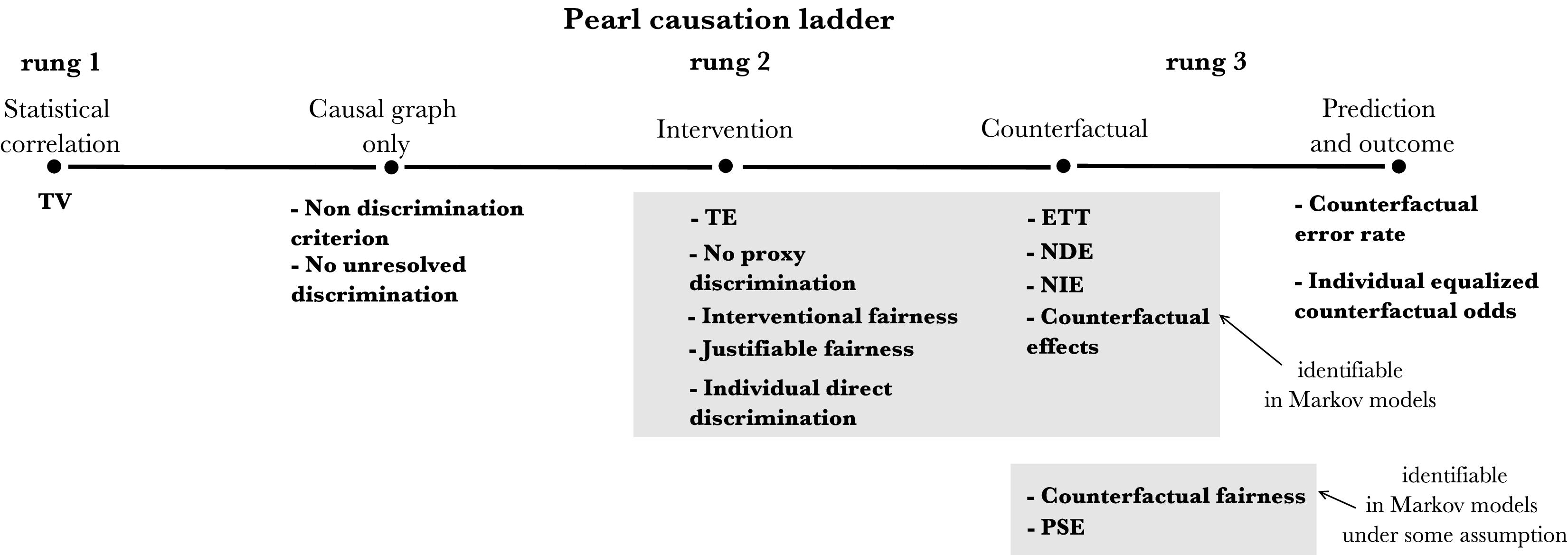}}
	\caption{Classification of causal-based fairness notions according to Pearl causation ladder~\cite{pearl2018book}}.
	\label{fig:diagram}
\end{figure*}
Fairness notions at rung 2 ($TE$, No-proxy discrimination, interventional and justifiable fairness, and individual direct discrimination) are applicable in any scenario where either experiments (RCT) are possible or hypothetical interventions are identifiable. As mentioned in Section~\ref{identInterv}, in Markovian models any intervention probability is identifiable from observational data. Hence, these fairness notions are always applicable in Markovian models. 
In semi-Markovian models, the applicability of these rung 2 notions depends on the identifiability of the intervention terms used in their respective definitions. For instance, for individual direct discrimination, the term in question is $CE(q_k, q'_k)$ in Eq.~\ref{eq:iddd}. 

The bulk of causal-based fairness notions are defined in terms of counterfactual quantities and hence are placed in rung 3 of the causation ladder. In Figure~\ref{fig:diagram}, the counterfactual notions are ranked from top to bottom according to their degree of applicability. For instance, counterfactual effects are placed on top of counterfactual fairness to indicate that the former is applicable in more scenarios than the latter. In Markovian models, the top 5 notions ($ETT$, $NDE$, $NIE$, and counterfactual effects) are always identifiable and hence applicable. That is, specific formula are already available to compute each counterfactual term used in their definitions. 

In Markovian models, the identifiability of counterfactual fairness 
depends on the identifiability of the term $\pr (y_{a_1}|\mathbf{X}=\mathbf{x},A=a_0)$ which is only identifiable if $\mathbf{X}$ does not contain any variable which is at the same time descendant of $A$ and ancestor of $Y$, that is, $\mathbf{X}\cap\mathbf{B} = \emptyset$ where $\mathbf{B} = An(Y) \cap De(A)$\cite{wu19}. $PSE$ is applicable provided that the model is Markovian and the recanting witness criterion is not satisfied. 
In semi-Markovian models, unless all model parameters are known (including $P(\mathbf{u})$)\footnote{In that case, it is possible to use the three steps abduction, action, and prediction~\cite{pearl2009causality}.}, the identifiability of rung 3 fairness notions depends on the criteria discussed in Section~\ref{sec:ident_ctf}, which rarely hold in practice. 

	Finally, counterfactual error rate and individual equalized counterfactual odds are special cases of rung 3 fairness notions as they are the only notions that condition on the true outcome $Y$ to assess the fairness of the prediction  $\hat{Y}$ (Eq.~\ref{eq:cder},\ref{eq:cier}, \ref{eq:cser}, and~\ref{eq:IndECOD}). Such conditioning has an important implication on identifiability since $Y$ is a collider, and conditioning on a collider creates a dependence between the previous variables~\cite{pearl2009causality}. This leads to unobservable confounding between the causes of $Y$. Hence, even if the causal model is Markovian, applying both notions turns it into a semi-Markovian model. Zhang and Bareinboim~\cite{zhang2018equality} define an identifiability criterion for counterfactual error rate in Markovian models called explanation criterion.

\section{Conclusion}
\label{sec:conclusion}

Notions of fairness that are inconsistent with the causal relationships in the data can lead to misleading conclusions about bias and discrimination of the outcomes. In particular, using causal reasoning to tackle the problem of fairness in machine learning has at least three advantages. First, it appropriately measure discrimination in presence of statistical anomalies (e.g., Simpson's paradox). Second, it provides natural interpretation of causal relationships between variables in support of discrimination claims. This is particularly important in the disparate treatment legal framework. Third, it makes it possible to break down the dependence between the sensitive attribute and the outcome into different paths (direct, indirect, etc.) which allows to assess fairness more accurately in presence of acceptable and unacceptable discrimination. 

Most of the causal-based notions of fairness examined in this paper rely on the availability of the causal graph. The issue of generating causal graphs consistent with the observed data is a known problem in the causal inference literature. Studying it for the specific context of machine learning fairness is a relevant direction for future work.

\section{Acknowledgements}
This work was supported by the European Research Council (ERC) project HYPATIA under the European Union’s Horizon 2020 research and innovation programme. Grant agreement n. 835294.

\bibliography{Causality_Survey}

\begin{thebibliography}{66}
\providecommand{\natexlab}[1]{#1}
\providecommand{\url}[1]{\texttt{#1}}
\expandafter\ifx\csname urlstyle\endcsname\relax
  \providecommand{\doi}[1]{doi: #1}\else
  \providecommand{\doi}{doi: \begingroup \urlstyle{rm}\Url}\fi

\bibitem[Andrade(2020)]{andrade2020mean}
Chittaranjan Andrade.
\newblock Mean difference, standardized mean difference (smd), and their use in
  meta-analysis: As simple as it gets.
\newblock \emph{The Journal of clinical psychiatry}, 81\penalty0 (5):\penalty0
  0--0, 2020.

\bibitem[Angwin et~al.(2016)Angwin, Larson, Mattu, and
  Kirchner]{angwin2016machine}
Julia Angwin, Jeff Larson, Surya Mattu, and Lauren Kirchner.
\newblock Machine bias. propublica.
\newblock \emph{See https://www. propublica.
  org/article/machine-bias-risk-assessments-in-criminal-sentencing}, 2016.

\bibitem[Athey and Imbens(2016)]{treemethods16imbens}
Susan Athey and Guido Imbens.
\newblock Recursive partitioning for heterogeneous causal effects.
\newblock \emph{Proceedings of the National Academy of Sciences}, 113\penalty0
  (27):\penalty0 7353--7360, 2016.

\bibitem[Avin et~al.(2005)Avin, Shpitser, and Pearl]{avin2005identifiability}
Chen Avin, Ilya Shpitser, and Judea Pearl.
\newblock Identifiability of path-specific effects.
\newblock In \emph{Proceedings of the 19th international joint conference on
  Artificial intelligence}, pages 357--363, 2005.

\bibitem[Barocas and Selbst(2016)]{barocas2016big}
Solon Barocas and Andrew~D Selbst.
\newblock Big data's disparate impact.
\newblock \emph{Calif. L. Rev.}, 104:\penalty0 671, 2016.

\bibitem[Ben-David et~al.(2007)Ben-David, Blitzer, Crammer, Pereira,
  et~al.]{representation2007}
Shai Ben-David, John Blitzer, Koby Crammer, Fernando Pereira, et~al.
\newblock Analysis of representations for domain adaptation.
\newblock \emph{Advances in neural information processing systems},
  19:\penalty0 137, 2007.

\bibitem[Bendick(2007)]{bendick2007situation}
Marc Bendick.
\newblock Situation testing for employment discrimination in the united states
  of america.
\newblock \emph{Horizons strat{\'e}giques}, \penalty0 (3):\penalty0 17--39,
  2007.

\bibitem[Bickel et~al.(1975)Bickel, Hammel, and O'Connell]{berkeley75}
Peter~J Bickel, Eugene~A Hammel, and J~William O'Connell.
\newblock Sex bias in graduate admissions: Data from berkeley.
\newblock \emph{Science}, 187\penalty0 (4175):\penalty0 398--404, 1975.

\bibitem[Chiappa(2019)]{chiappa2019path}
Silvia Chiappa.
\newblock Path-specific counterfactual fairness.
\newblock In \emph{Proceedings of the AAAI Conference on Artificial
  Intelligence}, volume~33, pages 7801--7808. PKP Publishing Services Network,
  2019.

\bibitem[Darlington(1971)]{darlington1971}
Richard~B Darlington.
\newblock Another look at “cultural fairness”.
\newblock \emph{Journal of educational measurement}, 8\penalty0 (2):\penalty0
  71--82, 1971.

\bibitem[Fisher(1992)]{fisher92}
Ronald~Aylmer Fisher.
\newblock Statistical methods for research workers.
\newblock In \emph{Breakthroughs in statistics}, pages 66--70. Springer, 1992.

\bibitem[Freedman et~al.(1998)Freedman, Pisani, and
  Purves]{freedman1998statistics}
D~Freedman, R~Pisani, and R~Purves.
\newblock Statistics, 3rd edn., pp. a-107, 1998.

\bibitem[Funk et~al.(2011)Funk, Westreich, Wiesen, St{\"u}rmer, Brookhart, and
  Davidian]{doublyRobust2011}
Michele~Jonsson Funk, Daniel Westreich, Chris Wiesen, Til St{\"u}rmer, M~Alan
  Brookhart, and Marie Davidian.
\newblock Doubly robust estimation of causal effects.
\newblock \emph{American journal of epidemiology}, 173\penalty0 (7):\penalty0
  761--767, 2011.

\bibitem[Galles and Pearl(1995)]{galles1995testing}
David Galles and Judea Pearl.
\newblock Testing identifiability of causal effects.
\newblock In \emph{Proceedings of the Eleventh conference on Uncertainty in
  artificial intelligence}, pages 185--195. ACM, 1995.

\bibitem[Glymour et~al.(2019)Glymour, Zhang, and Spirtes]{glymour2019review}
Clark Glymour, Kun Zhang, and Peter Spirtes.
\newblock Review of causal discovery methods based on graphical models.
\newblock \emph{Frontiers in genetics}, 10:\penalty0 524, 2019.

\bibitem[Guo et~al.(2020)Guo, Cheng, Li, Hahn, and Liu]{guo2020survey}
Ruocheng Guo, Lu~Cheng, Jundong Li, P~Richard Hahn, and Huan Liu.
\newblock A survey of learning causality with data: Problems and methods.
\newblock \emph{ACM Computing Surveys (CSUR)}, 53\penalty0 (4):\penalty0 1--37,
  2020.

\bibitem[Hardt et~al.(2016)Hardt, Price, and Srebro]{hardt2016equality}
Moritz Hardt, Eric Price, and Nati Srebro.
\newblock Equality of opportunity in supervised learning.
\newblock In \emph{Advances in neural information processing systems}, pages
  3315--3323, Spain, 2016.

\bibitem[Huan et~al.(2020)Huan, Wu, Zhang, and Wu]{huan2020fairness}
Wen Huan, Yongkai Wu, Lu~Zhang, and Xintao Wu.
\newblock Fairness through equality of effort.
\newblock In \emph{Companion Proceedings of the Web Conference 2020}, pages
  743--751, USA, 2020. ACM.

\bibitem[Huang and Valtorta(2006)]{huang06}
Yimin Huang and Marco Valtorta.
\newblock Identifiability in causal bayesian networks: A sound and complete
  algorithm.
\newblock In \emph{Proceedings of the national conference on artificial
  intelligence}, volume~21, page 1149, London, 2006. Menlo Park, CA; Cambridge,
  MA; London; AAAI Press; MIT Press; 1999, AAAI Press.

\bibitem[Imbens and Rubin(2015)]{rubin2015book}
Guido~W Imbens and Donald~B Rubin.
\newblock \emph{Causal inference in statistics, social, and biomedical
  sciences}.
\newblock Cambridge University Press, 2015.

\bibitem[Khademi et~al.(2019)Khademi, Lee, Foley, and
  Honavar]{khademi2019fairness}
Aria Khademi, Sanghack Lee, David Foley, and Vasant Honavar.
\newblock Fairness in algorithmic decision making: An excursion through the
  lens of causality.
\newblock In \emph{The World Wide Web Conference}, pages 2907--2914, USA, 2019.
  ACM.

\bibitem[Kilbertus et~al.(2017)Kilbertus, Carulla, Parascandolo, Hardt,
  Janzing, and Sch{\"o}lkopf]{kilbertus2017avoiding}
Niki Kilbertus, Mateo~Rojas Carulla, Giambattista Parascandolo, Moritz Hardt,
  Dominik Janzing, and Bernhard Sch{\"o}lkopf.
\newblock Avoiding discrimination through causal reasoning.
\newblock In \emph{Advances in Neural Information Processing Systems}, pages
  656--666, 2017.

\bibitem[K{\"u}nzel et~al.(2019)K{\"u}nzel, Sekhon, Bickel, and
  Yu]{metalearners2019}
S{\"o}ren~R K{\"u}nzel, Jasjeet~S Sekhon, Peter~J Bickel, and Bin Yu.
\newblock Metalearners for estimating heterogeneous treatment effects using
  machine learning.
\newblock \emph{Proceedings of the national academy of sciences}, 116\penalty0
  (10):\penalty0 4156--4165, 2019.

\bibitem[Kusner et~al.(2017)Kusner, Loftus, Russell, and
  Silva]{kusner2017counterfactual}
Matt~J Kusner, Joshua Loftus, Chris Russell, and Ricardo Silva.
\newblock Counterfactual fairness.
\newblock In \emph{Advances in neural information processing systems}, pages
  4066--4076, USA, 2017.

\bibitem[Lee et~al.(2011)Lee, Lessler, and Stuart]{trimming2011}
Brian~K Lee, Justin Lessler, and Elizabeth~A Stuart.
\newblock Weight trimming and propensity score weighting.
\newblock \emph{PloS one}, 6\penalty0 (3):\penalty0 e18174, 2011.

\bibitem[Loftus et~al.(2018)Loftus, Russell, Kusner, and Silva]{loftus18}
Joshua~R Loftus, Chris Russell, Matt~J Kusner, and Ricardo Silva.
\newblock Causal reasoning for algorithmic fairness.
\newblock \emph{arXiv preprint arXiv:1805.05859}, 2018.

\bibitem[Malinsky et~al.(2019)Malinsky, Shpitser, and Richardson]{malinsky19}
Daniel Malinsky, Ilya Shpitser, and Thomas Richardson.
\newblock A potential outcomes calculus for identifying conditional
  path-specific effects.
\newblock \emph{Proceedings of machine learning research}, 89:\penalty0 3080,
  2019.

\bibitem[Morgan and Winship(2015)]{morgan2015book}
Stephen~L Morgan and Christopher Winship.
\newblock \emph{Counterfactuals and causal inference}.
\newblock Cambridge University Press, 2015.

\bibitem[Nabi and Shpitser(2018)]{shpister18}
Razieh Nabi and Ilya Shpitser.
\newblock Fair inference on outcomes.
\newblock In \emph{Proceedings of the... AAAI Conference on Artificial
  Intelligence. AAAI Conference on Artificial Intelligence}, volume 2018, page
  1931. NIH Public Access, 2018.

\bibitem[O'Neill(2016)]{weapons16}
Catherine O'Neill.
\newblock Weapons of math destruction.
\newblock \emph{How Big Data Increases Inequality and Threatens Democracy},
  2016.

\bibitem[Pearl(2001)]{pearl01direct}
Judea Pearl.
\newblock Direct and indirect effects.
\newblock In \emph{Proceedings of the Seventeenth conference on Uncertainty in
  artificial intelligence}, pages 411--420, 2001.

\bibitem[Pearl(2009)]{pearl2009causality}
Judea Pearl.
\newblock \emph{Causality}.
\newblock Cambridge university press, 2009.

\bibitem[Pearl(2012)]{pearlBlog12}
Judea Pearl.
\newblock Judea pearl on potential outcomes, 2012.
\newblock URL
  \url{http://causality.cs.ucla.edu/blog/index.php/2012/12/03/judea-pearl-on-potentialoutcomes/}.

\bibitem[Pearl and Mackenzie(2018)]{pearl2018book}
Judea Pearl and Dana Mackenzie.
\newblock \emph{The book of why: the new science of cause and effect}.
\newblock Basic Books, 2018.

\bibitem[Pfohl et~al.(2019)Pfohl, Duan, Ding, and
  Shah]{pfohl2019counterfactual}
Stephen~R Pfohl, Tony Duan, Daisy~Yi Ding, and Nigam~H Shah.
\newblock Counterfactual reasoning for fair clinical risk prediction.
\newblock In \emph{Machine Learning for Healthcare Conference}, pages 325--358,
  2019.

\bibitem[Quick(2015)]{impactBias}
Kimberly Quick.
\newblock The unfair effects of impact on teachers with the toughest jobs.
\newblock \emph{The Century Foundation}, 2015.

\bibitem[Redmond and Baveja(2002)]{redmond2002data}
Michael Redmond and Alok Baveja.
\newblock A data-driven software tool for enabling cooperative information
  sharing among police departments.
\newblock \emph{European Journal of Operational Research}, 141\penalty0
  (3):\penalty0 660--678, 2002.

\bibitem[Rhee(2019)]{impact}
Michelle Rhee.
\newblock Impact: The dcps evaluation and feedback system for school-based
  personnel, 2019.

\bibitem[Romei and Ruggieri(2011)]{romei2011multidisciplinary}
Andrea Romei and Salvatore Ruggieri.
\newblock A multidisciplinary survey on discrimination analysis, 2011.

\bibitem[Rosenbaum and Rubin(1983)]{rubinPropensity83}
Paul~R Rosenbaum and Donald~B Rubin.
\newblock The central role of the propensity score in observational studies for
  causal effects.
\newblock \emph{Biometrika}, 70\penalty0 (1):\penalty0 41--55, 1983.

\bibitem[Salimi et~al.(2019)Salimi, Rodriguez, Howe, and
  Suciu]{salimi2019interventional}
Babak Salimi, Luke Rodriguez, Bill Howe, and Dan Suciu.
\newblock Interventional fairness: Causal database repair for algorithmic
  fairness.
\newblock In \emph{Proceedings of the 2019 International Conference on
  Management of Data}, pages 793--810, 2019.

\bibitem[Shimoni et~al.(2019)Shimoni, Karavani, Ravid, Bak, Ng, Alford, Meade,
  and Goldschmidt]{shimoni2019evaluation}
Yishai Shimoni, Ehud Karavani, Sivan Ravid, Peter Bak, Tan~Hung Ng,
  Sharon~Hensley Alford, Denise Meade, and Yaara Goldschmidt.
\newblock An evaluation toolkit to guide model selection and cohort definition
  in causal inference.
\newblock \emph{arXiv preprint arXiv:1906.00442}, 2019.

\bibitem[Shpitser(2013)]{shpitser13}
Ilya Shpitser.
\newblock Counterfactual graphical models for longitudinal mediation analysis
  with unobserved confounding.
\newblock \emph{Cognitive science}, 37\penalty0 (6):\penalty0 1011--1035, 2013.

\bibitem[Shpitser and Pearl(2006)]{shpitser2006identification}
Ilya Shpitser and Judea Pearl.
\newblock Identification of conditional interventional distributions.
\newblock In \emph{Proceedings of the Twenty-Second Conference on Uncertainty
  in Artificial Intelligence}, pages 437--444, 2006.

\bibitem[Shpitser and Pearl(2007)]{shpitser07-counterfactuals}
Ilya Shpitser and Judea Pearl.
\newblock What counterfactuals can be tested.
\newblock In \emph{23rd Conference on Uncertainty in Artificial Intelligence,
  UAI 2007}, pages 352--359, 2007.

\bibitem[Shpitser and Pearl(2008)]{shpitser08}
Ilya Shpitser and Judea Pearl.
\newblock Complete identification methods for the causal hierarchy.
\newblock \emph{Journal of Machine Learning Research}, 9\penalty0
  (Sep):\penalty0 1941--1979, 2008.

\bibitem[Simpson(1951)]{simpson1951interpretation}
Edward~H Simpson.
\newblock The interpretation of interaction in contingency tables.
\newblock \emph{Journal of the Royal Statistical Society: Series B
  (Methodological)}, 13\penalty0 (2):\penalty0 238--241, 1951.

\bibitem[Stuart(2010)]{matching2010stuart}
Elizabeth~A Stuart.
\newblock Matching methods for causal inference: A review and a look forward.
\newblock \emph{Statistical science: a review journal of the Institute of
  Mathematical Statistics}, 25\penalty0 (1):\penalty0 1, 2010.

\bibitem[Team et~al.(2016)]{team2016rstan}
Stan~Developent Team et~al.
\newblock Rstan: the r interface to stan.
\newblock \emph{R package version}, 2\penalty0 (1):\penalty0 522, 2016.

\bibitem[Tian(2004)]{tian04}
Jin Tian.
\newblock Identifying linear causal effects.
\newblock In \emph{AAAI}, pages 104--111, 2004.

\bibitem[Tian and Pearl(2002)]{tian02}
Jin Tian and Judea Pearl.
\newblock A general identification condition for causal effects.
\newblock In \emph{Aaai/iaai}, pages 567--573, 2002.

\bibitem[Tian and Shpitser(2003)]{tian03}
Jin Tian and Ilya Shpitser.
\newblock On the identification of causal effects.
\newblock 2003.

\bibitem[Tikka and Karvanen(2017{\natexlab{a}})]{tikka2017enhancing}
Santtu Tikka and Juha Karvanen.
\newblock Enhancing identification of causal effects by pruning.
\newblock \emph{The Journal of Machine Learning Research}, 18\penalty0
  (1):\penalty0 7072--7094, 2017{\natexlab{a}}.

\bibitem[Tikka and Karvanen(2017{\natexlab{b}})]{tikka2017simplifying}
Santtu Tikka and Juha Karvanen.
\newblock Simplifying probabilistic expressions in causal inference.
\newblock \emph{The Journal of Machine Learning Research}, 18\penalty0
  (1):\penalty0 1203--1232, 2017{\natexlab{b}}.

\bibitem[Verma and Rubin(2018)]{verma2018fairness}
Sahil Verma and Julia Rubin.
\newblock Fairness definitions explained.
\newblock In \emph{2018 ieee/acm international workshop on software fairness
  (fairware)}, pages 1--7. IEEE, 2018.

\bibitem[Wightman(1998)]{wightman1998lsac}
Linda~F Wightman.
\newblock Lsac national longitudinal bar passage study. lsac research report
  series.
\newblock 1998.

\bibitem[Wu et~al.(2019{\natexlab{a}})Wu, Zhang, and Wu]{wu19}
Yongkai Wu, Lu~Zhang, and Xintao Wu.
\newblock Counterfactual fairness: Unidentification, bound and algorithm.
\newblock In \emph{IJCAI}, pages 1438--1444, 2019{\natexlab{a}}.

\bibitem[Wu et~al.(2019{\natexlab{b}})Wu, Zhang, Wu, and Tong]{wu2019pc}
Yongkai Wu, Lu~Zhang, Xintao Wu, and Hanghang Tong.
\newblock Pc-fairness: A unified framework for measuring causality-based
  fairness.
\newblock In \emph{Advances in Neural Information Processing Systems}, pages
  3404--3414, 2019{\natexlab{b}}.

\bibitem[Yao et~al.(2020)Yao, Chu, Li, Li, Gao, and Zhang]{yao2020survey}
Liuyi Yao, Zhixuan Chu, Sheng Li, Yaliang Li, Jing Gao, and Aidong Zhang.
\newblock A survey on causal inference.
\newblock \emph{arXiv preprint arXiv:2002.02770}, 2020.

\bibitem[Zhang and Bareinboim(2018{\natexlab{a}})]{zhang2018equality}
Junzhe Zhang and Elias Bareinboim.
\newblock Equality of opportunity in classification: A causal approach.
\newblock In \emph{Advances in Neural Information Processing Systems}, pages
  3671--3681, 2018{\natexlab{a}}.

\bibitem[Zhang and Bareinboim(2018{\natexlab{b}})]{zhang2018fairness}
Junzhe Zhang and Elias Bareinboim.
\newblock Fairness in decision-making---the causal explanation formula.
\newblock In \emph{Proceedings of the AAAI Conference on Artificial
  Intelligence}, 2018{\natexlab{b}}.

\bibitem[Zhang and Wu(2017)]{zhang2017anti}
Lu~Zhang and Xintao Wu.
\newblock Anti-discrimination learning: a causal modeling-based framework.
\newblock \emph{International Journal of Data Science and Analytics},
  4\penalty0 (1):\penalty0 1--16, 2017.

\bibitem[Zhang et~al.(2016)Zhang, Wu, and Wu]{zhang2016situation}
Lu~Zhang, Yongkai Wu, and Xintao Wu.
\newblock Situation testing-based discrimination discovery: A causal inference
  approach.
\newblock In \emph{IJCAI}, volume~16, pages 2718--2724, 2016.

\bibitem[Zhang et~al.(2017{\natexlab{a}})Zhang, Wu, and Wu]{zhang2017achieving}
Lu~Zhang, Yongkai Wu, and Xintao Wu.
\newblock Achieving non-discrimination in data release.
\newblock In \emph{Proceedings of the 23rd ACM SIGKDD International Conference
  on Knowledge Discovery and Data Mining}, pages 1335--1344,
  2017{\natexlab{a}}.

\bibitem[Zhang et~al.(2017{\natexlab{b}})Zhang, Wu, and Wu]{zhang2017causal}
Lu~Zhang, Yongkai Wu, and Xintao Wu.
\newblock A causal framework for discovering and removing direct and indirect
  discrimination.
\newblock In \emph{Proceedings of the 26th International Joint Conference on
  Artificial Intelligence}, pages 3929--3935, 2017{\natexlab{b}}.

\bibitem[Zhou and Yamamoto(2020)]{zhou2020tracing}
Xiang Zhou and Teppei Yamamoto.
\newblock Tracing causal paths from experimental and observational data.
\newblock 2020.

\end{thebibliography}
\bibliographystyle{plainnat}

\end{document}